\documentclass{article}

\usepackage{arxiv}

\usepackage[utf8]{inputenc} 
\usepackage[T1]{fontenc}    
\usepackage{hyperref}       
\usepackage{url}            
\usepackage{booktabs}       
\usepackage{amsfonts}       
\usepackage{nicefrac}       
\usepackage{microtype}      
\usepackage{lipsum}		
\usepackage{doi}

\usepackage{times}
\usepackage{soul}
\usepackage[small]{caption}
\usepackage{amsthm}
\usepackage{algorithm}
\usepackage{algorithmic}
\usepackage[switch]{lineno}
\usepackage{array}

\usepackage{multirow}

\usepackage{amsmath,amssymb,amsfonts}
\usepackage{graphicx}
\usepackage{subcaption}
\usepackage{natbib}
\usepackage{fancyhdr}
\usepackage{hyperref}
\usepackage{marvosym}
\usepackage{enumitem}
\usepackage{siunitx}
\usepackage{tabularray}

\usepackage{makecell}
\usepackage{color}
\usepackage{xcolor}
\usepackage{tcolorbox}
\usepackage[normalem]{ulem}
\definecolor{myred}{rgb}{0.988, 0.267, 0.294}
\definecolor{myblue}{rgb}{0.086, 0.671, 0.980}

\newcommand{\boxref}[1]{Prompt Template~\ref{#1}}
\usepackage{float}
\newfloat{promptbox}{htbp}{lox}
\floatname{promptbox}{Box}

\theoremstyle{definition} 
\newtheorem{definition}{Definition}

\title{Improving LLM Reasoning with Homophily-aware Structural and Semantic Text-Attributed Graph Compression}

\date{} 					

\author{
Zijun Di \\
Shanghai Jiao Tong University \\
Shanghai, China \\
\texttt{dzj75du@sjtu.edu.cn}
\And
Bin Lu\thanks{Corresponding authors. Emails: \texttt{robinlu1209@sjtu.edu.cn} and \texttt{yiluofu@sjtu.edu.cn}.} \\
Shanghai Jiao Tong University \\
Shanghai, China \\
\texttt{robinlu1209@sjtu.edu.cn}
\And
Huquan Kang \\
Shanghai Jiao Tong University \\
Shanghai, China \\
\texttt{kinghiqian@sjtu.edu.cn}
\And
Luoyi Fu\footnotemark[1] \\
Shanghai Jiao Tong University \\
Shanghai, China \\
\texttt{yiluofu@sjtu.edu.cn}
\And
Jiaxin Ding \\
Shanghai Jiao Tong University \\
Shanghai, China \\
\texttt{jiaxinding@sjtu.edu.cn}
\And
Xiaoying Gan \\
Shanghai Jiao Tong University \\
Shanghai, China \\
\texttt{ganxiaoying@sjtu.edu.cn}
\And
Lei Zhou \\
Shanghai Jiao Tong University \\
Shanghai, China \\
\texttt{zhoulei1588@sjtu.edu.cn}
\And
Xinbing Wang \\
Shanghai Jiao Tong University \\
Shanghai, China \\
\texttt{xwang8@sjtu.edu.cn}
}

\renewcommand{\shorttitle}{\textit{arXiv} Template}

\hypersetup{
pdftitle={A template for the arxiv style},
pdfsubject={q-bio.NC, q-bio.QM},
pdfauthor={David S.~Hippocampus, Elias D.~Striatum},
pdfkeywords={First keyword, Second keyword, More},
}

\begin{document}
\maketitle

\begin{abstract}
    Large language models (LLMs) have demonstrated promising capabilities in Text-Attributed Graph (TAG) understanding. Recent studies typically focus on verbalizing the graph structures via handcrafted prompts, feeding the target node and its neighborhood context into LLMs. 
    However, constrained by the context window, existing methods mainly resort to random sampling, often implemented via dropping node/edge randomly, which inevitably introduces noise and cause reasoning instability.
    We argue that graphs inherently contain rich \textit{structural} and \textit{semantic} information, and that their effective exploitation can unlock potential gains in LLMs reasoning performance. 
    To this end, we propose \uline{H}omophily-aware \uline{S}tructural and \uline{S}emantic \uline{C}ompression for LLMs ($\text{H}\text{S}_2\text{C}$), a framework centered on exploiting graph homophily. 
    Structurally, guided by the principle of Structural Entropy minimization, we perform a global hierarchical partition that decodes the graph’s essential topology. This partition identifies naturally cohesive, homophilic communities, while discarding stochastic connectivity noise. 
    Semantically, we deliver the detected structural homophily to the LLM, empowering it to perform differentiated semantic aggregation based on predefined community type. 
    This process compresses redundant background contexts into concise community-level consensus, selectively preserving semantically homophilic information aligned with the target nodes. Extensive experiments on 10 node-level benchmarks across LLMs of varying sizes and families demonstrate that, by feeding LLMs with structurally and semantically compressed inputs, $\text{H}\text{S}_2\text{C}$ simultaneously enhances the compression rate and downstream inference accuracy, validating its superiority and scalability. Notably, on OGBN-ArXiv, it achieves a 94.98\% graph scale compression while improving accuracy by 3.06\%–4.92\%. Extensions to 7 diverse graph-level benchmarks further consolidate $\text{H}\text{S}_2\text{C}$'s task generalizability. 
\end{abstract}

\section{Introduction}\label{Section1}
Large language models (LLMs) have recently emerged as powerful tools for text-attributed graph (TAG) understanding, delivering strong inference performance on downstream tasks~\citep{chaohuang_survey, jiaweihan_survey, rw_tang_sigir2024graphgpt}. Existing works primarily focus on integrating graph structures and node textual attributes into LLMs through handcrafted prompts, typically combining the target node information with its local surrounding context~\citep{rw_ye_EACL2024instructglm, rw_wu_icml2025when, rw_chen_icml2024llaga}. This approach provides considerable flexibility, enabling the LLMs' reasoning of TAGs to evolve in tandem with model advancements and hereby becoming a promising direction. 

\begin{figure}[t]
    \centering
    \begin{subfigure}[b]{0.48\linewidth} 
        \centering
        \includegraphics[width=\linewidth]{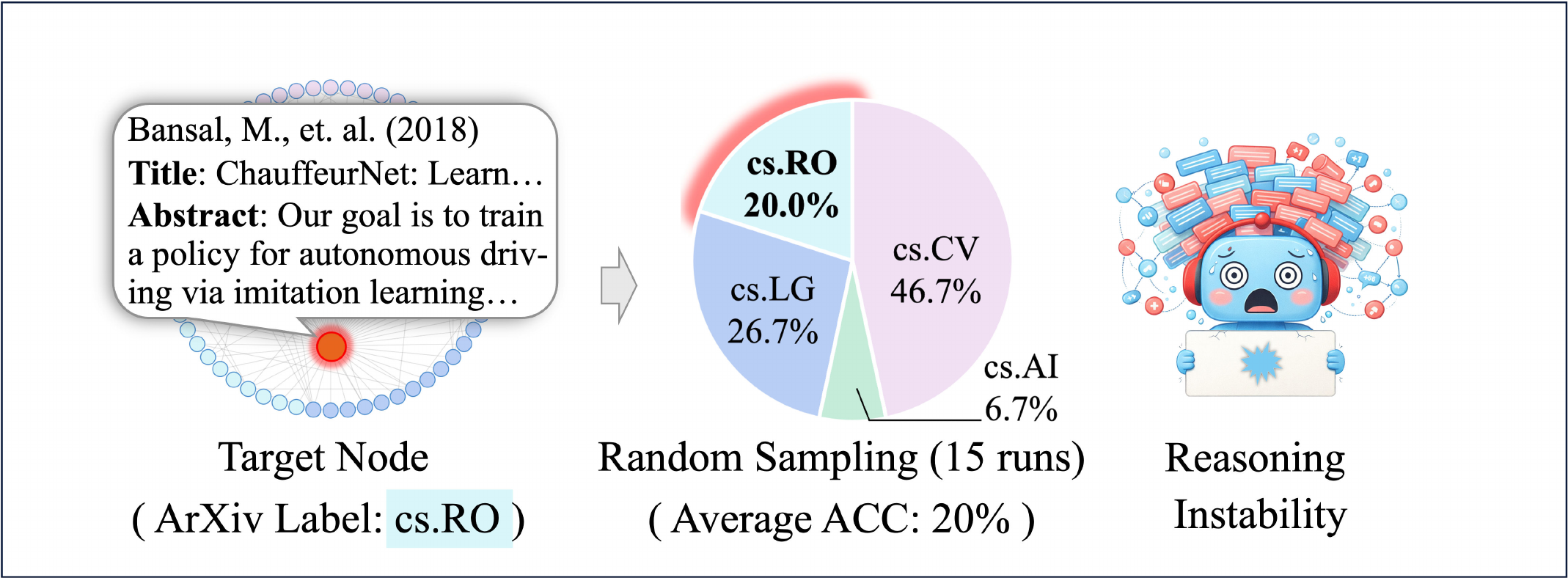}
        \caption{Reasoning instability.} 
        \label{fig:intro_fluctuation} 
    \end{subfigure}
    \hfill 
    \begin{subfigure}[b]{0.48\linewidth}
        \centering
        \includegraphics[width=\linewidth]{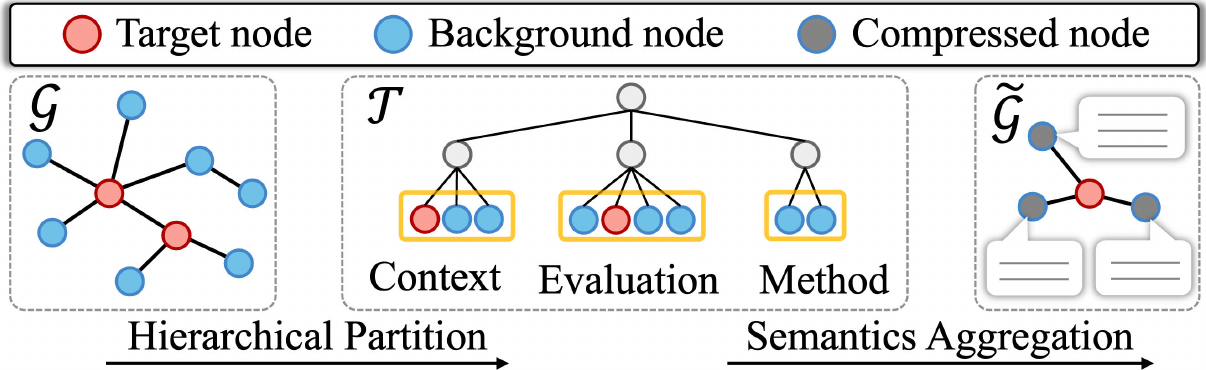}
        \caption{$\text{H}\text{S}_2\text{C}$ toy sample.}
        \label{fig:intro_framework}
    \end{subfigure}
    \vspace{-5pt}
    \captionsetup{aboveskip=5pt, belowskip=2pt} 
    \caption{\textbf{The challenge and our solution.} (a) Illustration of LLM reasoning instability induced by random sampling strategy. (b) Illustration of $\text{H}\text{S}_2\text{C}$ compressing the graph to generate compact LLM inputs.}
    
    \label{fig:intro_overall}
    \vspace{-15pt}
\end{figure}

However, constrained by context window, existing methods mainly resort to random sampling strategy~\citep{rw_wang_ijcal2024nlp_context_survey}, which inevitably introduces noise and cause reasoning instability. Specifically, stochastic sampling induces fluctuations in the neighborhood context, where the resulting drift in input semantics causes oscillations in LLM predictions. Moreover, the randomly altered local structural fragments further exacerbate variance in inference performance. As illustrated in Figure~\ref{fig:intro_fluctuation}, random sampling across 15 different seeds results in only 20\% classification accuracy for target node. Therefore, we argue for performing \textbf{graph compression} prior to feeding the data into LLMs. A straightforward way is node-level compression, which drops nodes or edges based on local importance metrics, such as degree or semantic similarity~\citep{baseline_rag_li2025large}. Nevertheless, these heuristic methods overlook the global structure dependency and semantic consistency, leading to suboptimal reasoning performance. In fact, graphs inherently encode rich \textit{structural} and \textit{semantic} information, and that their effective exploitation can unlock potential gains in LLMs reasoning performance. 

To this end, we propose \uline{H}omophily-aware \uline{S}tructural and \uline{S}emantic \uline{C}ompression for LLM ($\text{H}\text{S}_2\text{C}$), which centers on exploiting graph homophily. 
Structurally, guided by the principle of Structural Entropy minimization, we perform a global hierarchical partition to decode the graph’s essential topology. This partition identifies naturally cohesive, homophilic communities, while discarding stochastic connectivity noise. 
Semantically, we deliver the detected structural homophily to the LLM, empowering it to perform differentiated semantic aggregation based on predefined community type. This process compresses redundant background contexts into concise community-level consensus, selectively preserving semantically homophilic information aligned with the target nodes. 
We take paper classification in Figure~\ref{fig:intro_framework} as a toy example to illustrate the idea of $\text{H}\text{S}_2\text{C}$, which contains two main procedures. 
Given a citation network $\mathcal{G}$ centered with a target paper (\textcolor{myred}{the red node}), we decode $\mathcal{G}$'s essential topology through a global hierarchical partition guided by Structural Entropy (SE) minimization, which detects naturally homophilic communities while discarding connectivity noise. Each community may correspond to the target node’s comparative evaluation, or its research context or methodology.
Secondly, $\text{HS}_2\text{C}$ delivers the detected structural homophily to LLM, empowering it to perform community-variant semantics aggregation of background nodes (\textcolor{myblue}{the blue nodes}) within each communities. 
The differentiated summarization selectively preserves homophilic semantics aligned with the target node while filtering out irrelevant or noisy content.
Finally, we reconstruct a compressed $\widetilde{\mathcal{G}}$, containing sufficient homophilic-aware topology and information while minimizing overall graph storage.

To summarize, our main contributions are as follows: 
\begin{itemize}[leftmargin=*,itemsep=0pt,topsep=2pt]
    \item 
    We propose $\text{H}\text{S}_2\text{C}$, a homophily‑driven framework for community‑level compression of TAGs, jointly exploiting structural and semantic homophily to retain topological and contextual information for LLM downstream reasoning.
    \item  Rather than local node-level importance criteria, $\text{H}\text{S}_2\text{C}$ performs a hierarchical structure partition to decode graphs' essential topology and from a global perspective. This method identifies naturally homophilic structures and categorize communities into four distinct types, which promotes homophilic semantics discovery.
    \item Extensive experiments on 10 real-world node-level TAG benchmarks show that $\text{HS}_2\text{C}$ improves both compression efficiency and prediction accuracy, which consistently outperforms all baselines, validating its performance superiority and scalability. Typically, on OGBN‑ArXiv, it achieves a 94.98\% graph scale compression while improving accuracy by 3.06\%–4.92\%. Furthermore, extentions on 7 diverse graph-level benchmarks consolidate $\text{HS}_2\text{C}$'s generalizability.
\end{itemize}

\section{Related Work}\label{Section2}
\subsection{TAG Understanding with LLM}
LLM-based graph understanding methods generally fall into three main categories:
\textbf{1. GNNs as prefix}. This line of work integrates graph neural networks (GNNs) as structural encoders, whose outputs are prepended as prefix tokens to LLM inputs. A representative method, GraphGPT~\cite{rw_tang_sigir2024graphgpt}, employs self-supervised instruction tuning to align GNN representations with LLMs in the semantic space. GraphGPT transforms local graph structures into token sequences, which are then passed to LLMs for inference. Due to the inherent input length limitations of LLMs, full-graph context cannot be used. GraphGPT addresses this by sampling h-hop neighborhoods around the target node to construct compact subgraphs, which are encoded and passed to LLMs to support structure-aware reasoning. 
\textbf{2. Tuning-Required LLM Prediction}. This category focuses on fine-tuning LLMs using specially designed prompts that represent graph content. LLaGA~\cite{rw_chen_icml2024llaga} introduces a structure-aware input mechanism by constructing fixed-shape multi-hop trees for each node via random neighbor sampling. Each node’s textual attributes are encoded using pretrained models (e.g., SBERT), projected into the LLM embedding space, and concatenated with task instructions for end-to-end prediction. InstructGLM~\cite{rw_ye_EACL2024instructglm} builds localized ego-graph descriptions using templated natural language prompts, enabling instruction tuning for node classification and link prediction. To reduce computational overhead and address input length constraints, InstructGLM employs a GraphSAGE-style sampling strategy that retains representative structural semantics while ensuring prompt compactness.
\textbf{3. Tuning-free LLM prediction}. This paradigm adopts in-context learning to perform graph tasks without any fine-tuning. 
As LLMs continue to evolve rapidly, with increasing parameter scales and growing specialization across vertical domains, methods that decouple graph reasoning from specific model parameters offer greater adaptability and long-term applicability.  
Graph-LLM~\cite{Chen_GraphLLM2024KDD} explores the direct application of LLMs to graph learning by proposing two pipelines, LLMs-as-Enhancers and LLMs-as-Predictors, for node classification, demonstrating that LLMs can perform graph reasoning through textual prompts. Its findings offer new insights into the standalone potential of LLMs for graph–text integration.
Talk Like a Graph~\cite{fatemi_talk2024ICLR} investigates how to convert graph-structured data into natural language, and evaluates how different encoding strategies influence LLMs’ ability to perform graph reasoning without fine-tuning.
LLMNodeBed~\cite{rw_wu_icml2025when} designs structured input sequences by performing fixed-size random sampling over 1-hop neighborhoods of each target node. It then concatenates the textual descriptions of the target node and its sampled neighbors into a unified prompt fed directly into the LLM. This approach enables effective classification through the LLM’s inherent reasoning capabilities, without relying on GNNs or additional fine-tuning. 

However, due to the inherent context limitations and computational cost of LLMs, existing methods predominantly rely on heuristic strategies such as fixed-size random sampling of neighborhood nodes. This limits access to complete structural and semantic contexts, potentially leading to the omission of critical relational patterns. Although increasing efforts have been made to integrate LLMs with graph-structured data, the problem of neighborhood selection under input constraints remains overlooked. We argue that effective graph compression, which preserves essential topological structures and semantic content while reducing input redundancy, offers a promising and cost-effective direction. As far as we know, this work is the first to systematically address this problem from a structure awareness and semantics aggregation perspective. 

\subsection{Graph Compression}
TAG compression aims to reduce the scale of graph data while preserving critical structural and semantic information, thereby enabling efficient reasoning under the input and computational constraints of LLMs. Relevant research lines in graph reduction generally fall into two categories: Graph Coarsening and Graph Condensation.
\textbf{1. Graph Coarsening}. Traditional coarsening methods, such as spectral clustering and edge contraction algorithms, reduce graph size by iteratively grouping nodes into supernodes based on topological connectivity. However, these methods primarily focus on preserving structural properties (e.g., eigenvalues or cut sizes) while often overlooking the rich textual semantics of nodes, making them suboptimal for LLM-based reasoning where semantic context is paramount. More critically, these methods are typically target-agnostic. They indiscriminately merge target nodes and background nodes into supernodes. This loss of specific target identity is fundamentally misaligned with the objectives of node-centric downstream tasks.
\textbf{2. Graph Condensation}. A another related research line is Graph Condensation, which generates a synthetic compact graph on which a GNN can generalize to the original graph tasks. Its goal is to approximate the original training behavior to enable efficient model training with minimal data. Most condensation methods are gradient‑driven, for example, GCond~\cite{related_work_gc_1} and DosCond~\cite{related_work_gc_2} employ a gradient‑matching framework to optimize the condensed graph to mimic the GNN training trajectory on the full graph. However, the heavy computational cost of gradient‑based optimization in LLM renders these methods unsuitable for our TAG compression. 
In contrast, $\text{H}\text{S}_2\text{C}$ is explicitly tailored for LLM‑based graph reasoning rather than gradient‑driven task condensation. By performing community‑level compression that preserves the graph’s essential structure and contextual semantics, thereby alleviating LLM input and computation constraints while enhancing LLM reasoning consistency.

\section{Preliminaries}
\label{sec:others}
Denote a TAG as $\mathcal{G}= (\mathcal{V}, \mathcal{E}, \mathcal{R})$, where $\mathcal{V}$ denotes the node set, $\mathcal{E}$ is the edge set, and $\mathcal{R} = \{R_1, \ldots, R_{|\mathcal{V}|}\}$ represents the collection of raw text sequences associated with each node $v_i \in \mathcal{V}$, where each $R_i = [r_1, \ldots, r_L]$ is a word-level sequence of length $L$. Each node $v_i$ is further assigned a ground-truth label $y_i \in \mathcal{Y}$.
The node set $\mathcal{V}$ is partitioned into two subsets: the target node set $\mathcal{V}_{\text{tg}} \subset \mathcal{V}$, which contains the nodes to be classification in downstream task, and the background node set $\mathcal{V}_{\text{bg}} = \mathcal{V} \setminus \mathcal{V}_{\text{tg}}$, consisting of the remaining nodes of $\mathcal{G}$.
A complete summary of all notations used in this paper is provided in Appendix~\ref{app: Notation}:

\subsection{Problem Formulation}
Given an input TAG $\mathcal{G}$, our objective is to construct a compressed graph $\widetilde{\mathcal{G}} = (\widetilde{\mathcal{V}}, \widetilde{\mathcal{E}}, \widetilde{\mathcal{R}})$ that retains essential structural and semantic information while significantly reducing the graph scale, $|\widetilde{\mathcal{V}}| \ll |\mathcal{V}|$.
To support downstream reasoning tasks, we preserve all target nodes $\mathcal{V}_\text{tg}$ in the compressed graph and apply compression only to background nodes $\mathcal{V}_\text{bg}$. 
We aim to discover the homophilic structure and semantics for compression. Specifically, homophilic structure refers to topological patterns in which nodes sharing similar connection patterns tend to be grouped within the same community, while homophilic semantics reflects the semantic consistency of nodes’ textual attributes within community. By maintaining both the structural organization and semantic coherence of original $\mathcal{G}$, the compressed $\widetilde{\mathcal{G}}$ retains essential contextual information while greatly reducing overall graph complexity, thereby enabling more accurate and efficient downstream reasoning.

\subsection{Structural Entropy (SE)}
Structural Entropy (SE)~\cite{se_li_tit2016} is a quantitative measure of the amount of structural information embedded within a complex graph system~\cite{OOD_SEIB, se_aaai_hou2025ood, se_yang2024incremental, se_nips_liu2019com_deception, se_www_zou2023segsl}. It reveals the intrinsic hierarchical organization of a graph by decoding the original graph $\mathcal{G}$ into a coding tree $\mathcal{T}$ through a hierarchical abstraction strategy, where nodes are partitioned into communities at various levels of granularity. In recent years, structural entropy has been widely applied in diverse domains such as bioinformatics~\cite{se_icml_wu2023sega, ijcai2022p497, se_icml_wu2022poling,10735397}, information security~\cite{LI2017211, 9465220}, and social network analysis~\cite{se_icml_wu2022poling, 10.1145/3660522, Cao_Peng_Yu_Yu_2024, 10.1145/3637528.3671871}. 
Specifically, given an undirected graph $\mathcal{G} = (\mathcal{V}, \mathcal{E})$, the SE of $\mathcal{G}$ is computed by performing a biased random walk to estimate node visit probabilities and summing the entropy contributions of all nodes accordingly:
\begin{equation}\label{Eq: Structural Entropy}
    \mathcal{H}^{\mathcal{T}}(\mathcal{G})=\!\!\!\!\!\sum_{\alpha\in\mathcal{T},\alpha\neq\lambda}\!\!\!\!\mathcal{H}^{\mathcal{T}}(\mathcal{G};\alpha)=-\!\!\!\!\!\sum_{\alpha\in\mathcal{T},\alpha\neq\lambda}\!\!\!\frac{g_\alpha}{\operatorname{vol}(\mathcal{G})}\log_2 \frac{\operatorname{vol}\left(\alpha\right)}{\operatorname{vol}\left(\alpha^-\right)},
\end{equation}
where $\alpha$ denotes a non-root node in the coding tree $\mathcal{T}$, whose associated vertex set is $T_\alpha \subset \mathcal{V}$. The term $g_\alpha$ represents the number of edges connecting nodes in and outside $T_\alpha$. The $\alpha^-$ refers to the immediate predecessor of $\alpha$. The volume terms $\operatorname{vol}(\alpha)$, $\operatorname{vol}(\alpha^-)$, and $\operatorname{vol}(\mathcal{G})$ represent the sum of node degrees within $\alpha$, $\alpha^-$, and $\mathcal{G}$, respectively. Minimizing $\mathcal{H}^{\mathcal{T}}(\mathcal{G})$ yields an optimal coding tree $\mathcal{T}^*$ that achieves $\mathcal{H}(\mathcal{G})=\min_{\forall \mathcal{T}}\{\mathcal{H}^\mathcal{T}(\mathcal{G})\}$, which encodes the graph in a way that minimizes the uncertainty of node codewords encountered during random walks. In this sense, the process of measuring the SE not only quantifies the information embedded in the graph but also implicitly reveals its underlying structural organization. It effectively separates regular structural patterns from randomness or noise in the topology. 

\section{Methodology}
\begin{figure*}[!t]
    \centering
    \includegraphics[width=\linewidth]{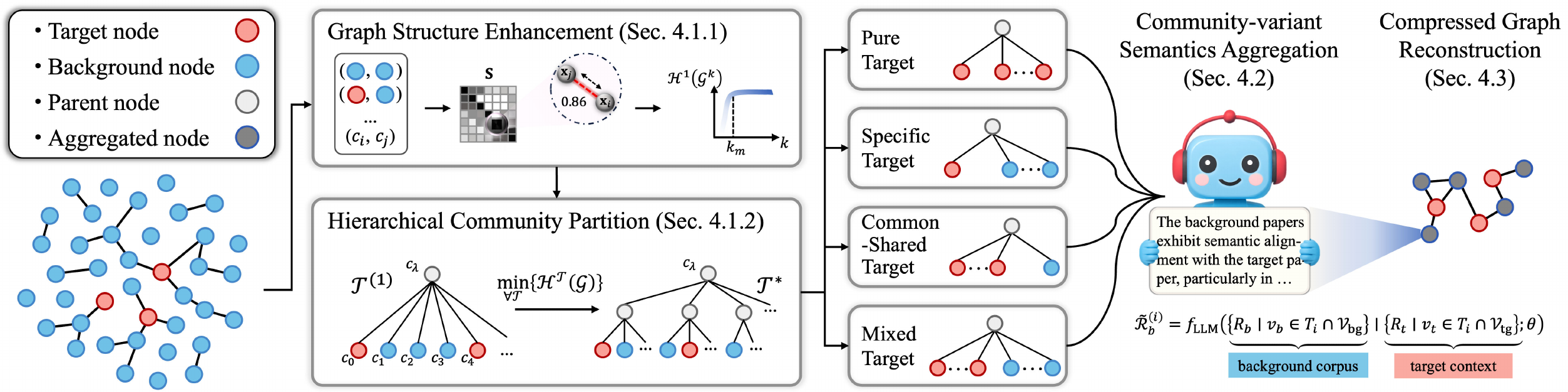}
    \caption{The overall framework of the proposed $\text{HS}_2\text{C}$, which consists of 3 modules. Firstly, we detect homophilic structures by enhancing the graph topology and minimizing SE to obtain a hierarchical community partition. 
    Secondly, we aggregate the textual attributes of background nodes within each community to generate concise, semantically aligned summaries.
    Finally, we reconstruct a compressed graph $\widetilde{\mathcal{G}}$ that preserves essential structural and semantic information for downstream inference.}
    \label{fig:Overall framework}
    \vspace{-12pt}
\end{figure*}
In this section, we elaborate the details of $\text{HS}_2\text{C}$, the overall framework is shown in Figure~\ref{fig:Overall framework}. The main idea is to perform community‑level compression of TAGs through detecting homophilic structures and aggregating homophilic semantics. In the following sections, we provide a detailed introduction to each module.

\subsection{Entropy-guided Hierarchical Structure Partition} \label{section4_1}
To identify homophilic structures for compression, we first introduce a hierarchical partitioning of the graph structure. In practice, the observed topology of a graph rarely reflects the true and complete underlying structure. Such incompleteness can arise from various factors, including missing records caused by temporal evolution, task‑specific data splits, or structural omissions during data collection. These issues make it difficult to accurately detect communities that exhibit homophilic structures. Therefore, the core idea is to \textbf{identify homophilic structures within an incomplete topology}. Inspired by Structural Entropy (SE)~\cite{se_li_tit2016}, we propose an Entropy‑guided Hierarchical Structure Partition framework that decodes the graph’s essential topology from the perspective of structural uncertainty. This framework comprises two key modules: Graph Structure Enhancement (GSE) module and Hierarchical Community Partition (HCP) module.

\subsubsection{Graph Structure Enhancement (GSE)}
We adopt a similarity based structural enhancement on top of the original topology, to recover and reinforce latent relationships and mitigate the information loss arising from the incompleteness of the original graph structure. This enhancement enriches the graph with more comprehensive structural information, 
thereby improving the accuracy and robustness of the subsequent homophilic structure detection.
Specifically, we initially treat each node $v_i$
as an independent community, defined as $c_i$, whose vertex set is $T_i=\{v_i\}$ with $T_i \subset \mathcal{V}$. In this way, the original graph $\mathcal{G}$ is viewed as a collection of $|\mathcal{V}|$ mutually independent communities, i.e., $\mathcal{G}=(\mathcal{C},\mathcal{E})$, $\mathcal{C} = \{ c_1, c_2, \ldots, c_{|\mathcal{V}|} \}$. 
According to the widely adopted homophily assumption in graph learning, objects with similar attributes tend to form connections~\cite{homophily_assump}. For example, research papers are often cited by subsequent studies in related 
domains, resulting in dense citation networks. Motivated by this principle, we compute the pairwise similarity matrix $\mathbf{S} \in \mathbb{R}^{|\mathcal{V}| \times |\mathcal{V}|}$ for any initial community pair $(c_i, c_j)$, where $c_i, c_j \in \mathcal{C}$, using the Pearson Correlation Coefficient:
\begin{equation}
    \mathbf{S}_{ij} = \frac{\sum_{m=1}^{d}\big(x_{i,m}-\mu_i\big)\big(x_{j,m}-\mu_j\big)}{\sqrt{\sum_{m=1}^{d}\big(x_{i,m}-\mu_i\big)^2}\sqrt{\sum_{m=1}^{d}\big(x_{j,m}-\mu_j\big)^2}},
\end{equation}
where $\mathbf{x}_i \in \mathbb{R}^d$ represents the feature vector of node $v_i$. 
For each node $v_i$, the $\mathbf{x}_i$ is obtained by feeding its text attribute, the raw text sequence $R_i=[r_1, r_2, \ldots]$, into a pretrained BERT model~\cite{devlin2019bert} to generate a semantic embedding: $\mathbf{x}_i=f_{\text{BERT}}({R_i};\zeta)$, where $\zeta$ denotes the parameters of the BERT model. Furthermore, $\mu_i= \frac{1}{d}\sum_{m=1}^{d}x_{i,m}$ represents the mean value of $\mathbf{x}_i$.

Building upon the similarity matrix $\mathbf{S}$, we enhance the graph structure by introducing a $k$‑nearest neighbors (KNN) strategy. 
For each community $c_i$, we define its $k$-nearest neighbor community set as $\mathcal{N}_{k}(c_i) = \text{Top-}k\big(\{c_j \mid j \neq i\}, S_{ij}\big)$, which selects the \text{top-}k most similar neighbors of $c_i$ from the similarity matrix $\mathbf{S}$. 
Based on this, the corresponding edge set is constructed as $\mathcal{E}_{k} = \{ (c_i, c_j) \mid c_j \in \mathcal{N}_{k}(c_i) \}$, yielding $\hat{\mathcal{G}}$ enhanced by an auxiliary $k$-nearest neighbor graph $\hat{\mathcal{G}}^{(k)} = (\mathcal{C}, \mathcal{E} \cup \mathcal{E}_{k})$. 
However, a critical step lies in determining the most suitable parameter $k$, as it directly affects the structural quality of the enhanced graph. 
A small $k$ may lead to an under-connected graph, insufficiently capturing contextual information or potential hierarchical relationships, whereas an excessively large $k$ may introduce noise and redundant edges. 
To systematically determine the optimal value of $k$, following Zou et al.~\cite{se_www_zou2023segsl}, we propose an entropy-guided selection strategy based on one-dimensional SE, quantifying the inherent information embedded within the graph's essential topology. 
As $k$ gradually increases, $\mathcal{H}^1(\hat{\mathcal{G}}^{(k)})$ initially rises due to the incorporation of additional structural edges, but eventually reaches a plateau. 
Based on this observation, we identify the optimal $k_m$ by tracking the marginal increase in SE, with the goal of retaining maximal structural information.
After determining the optimal value $k_m$, we derive the final enhanced graph: $\hat{\mathcal{G}}=(\mathcal{C}, \hat{\mathcal{E}})$, where $\hat{\mathcal{E}}=\mathcal{E} \cup \mathcal{E}_{k_m}$.

\subsubsection{Hierarchical Community Partition (HCP)}
The uncertainty inherent in graph topologies often conceals their essential structure. 
To uncover the latent homophilic structures within a graph, we propose a natural hierarchical partitioning method inspired by SE~\cite{se_li_tit2016}, which decomposes TAG into coding tree $\mathcal{T}$. 
SE measures the uncertainty embedded within graph topologies by performing a biased random walk throughout the entire graph and calculating the visit probability of each node, which in turn derives the codeword lengths that represent the topology. 
Correspondingly, more concise codewords can potentially reduce the uncertainty within graph structure.
By minimizing the SE, i.e., $\mathcal{H}(\hat{\mathcal{G}})=\min_{\forall \mathcal{T}}\{\mathcal{H}^\mathcal{T}(\hat{\mathcal{G}})\}, $
we can derive and decode an optimal coding tree $\mathcal{T}^\ast$, which provides an encoding of $\hat{\mathcal{G}}$ such that the uncertainty associated with the codewords of nodes visited during random walks is minimized. 
This process effectively reveals the intrinsic structural information that is embedded in the graph. 

Given a graph $\hat{\mathcal{G}} = (\mathcal{C}, \hat{\mathcal{E}})$, we initially construct a coding tree $\mathcal{T}^{(1)}$ of height 1,  in which all communities $\mathcal{C}$ in $\hat{\mathcal{G}}$ are treated as leaf communities directly connected to the root community $c_\lambda$, whose vertex set is denoted as $T_\lambda \subseteq \mathcal{V}$. At this stage, all communities $\mathcal{C}$ in the original graph constitute the set of 1‑level communities $\mathcal{C}^{(1)}$ in $\mathcal{T}^{(1)}$. 
To construct a coding tree $\mathcal{T}$ with minimal SE under the above objective, based on SEP~\cite{se_icml_wu2022poling}, we adopt two key operators, $\mathbf{MERGE}$ and $\mathbf{DROP}$. A detailed illustration of the entire construction procedure, the accompanying algorithm pseudocode, and the analysis of the time complexity are given in Appendix~\ref{app: Supplementary Algorithm Details}.

Through greedy iterative MERGE and DROP operations guided by SE minimization, we obtain an optimal coding tree $\mathcal{T}^\ast$. This tree provides a hierarchical partition of the graph $\hat{\mathcal{G}}$, minimizes structural uncertainty, and effectively captures the underlying essential structure. The nodes in the original graph, served as the leaf communities, are naturally partitioned into community clusters of the hierarchical coding tree. We take the finest‑granularity partition immediately above the original leaf communities $\mathcal{C}^{(1)}$ in the coding tree as the set of homophilic structures $\mathcal{C}_\text{homo}$,
where $\mathcal{C}_\text{homo}=\{c_1,c_2, \ldots, c_m \ | \ c_i \notin \mathcal{C}^{(1)}, \ 1\leq i \leq m, \ m < |\mathcal{V}|, \ \ T_i \subset \mathcal{V}, \ |T_i| \geq 1\}$.
Consequently, while measuring the information embedded in $\hat{\mathcal{G}}$, the procedure simultaneously detects and explicitly delineates the homophilic structures inherent in the graph.

\subsection{Community-variant Semantics Aggregation} \label{section4_2}
To further capture homophilic semantics within each community $\mathcal{C}_\text{homo}=\{c_1, \ldots,c_m\}$, we define different types of homophilic structures and propose a Community‑variant Semantics Aggregation (CSA) module.

According to the composition ratio of target nodes ($\mathcal{V}_{\text{tg}}$) and background nodes ($\mathcal{V}_{\text{bg}}$), the homophilic structures are categorized into four types: Pure Target, Specific Target, Common-Shared Target, and Mixed Target.

\begin{itemize}[leftmargin=*,itemsep=3pt,topsep=5pt]
    \item \textbf{Pure Target}: All nodes in $c_i$ are target nodes, $\forall v \in T_i$, $v \in \mathcal{V}_{\text{tg}}$.
    \item \textbf{Specific Target}: Exactly one target node in community $c_i$, and all the remaining nodes are background nodes, $\exists! v_t \in T_i,\ v_t \in \mathcal{V}_{\text{tg}}, \ \text{and} \  \ \forall v \in T_i \setminus \{v_t\},\ v \in \mathcal{V}_{\text{bg}}$.
    \item \textbf{Common-Shared Target}: Multiple target nodes and exactly one background node in $c_i$, $\exists! v_b \in T_i,\ v_b \in \mathcal{V}_{\text{bg}}, \ \text{and} \ \ \forall v \in T_i \setminus \{v_b\},\ v \in \mathcal{V}_{\text{tg}}, \ \text{with} \ \ |T_i \cap \mathcal{V}_{\text{tg}}| > 1 $.
    \item \textbf{Mixed Target}: Community $c_i$ contains multiple target nodes and multiple background nodes,     $\exists v_t \in T_i,\ v_t \in \mathcal{V}_{\text{tg}}, \ \exists v_b \in T_i,\ v_b \in \mathcal{V}_{\text{bg}}, \ \text{with } \ |T_i \cap \mathcal{V}_{\text{tg}}| > 1 \ \text{ and } \ |T_i \cap \mathcal{V}_{\text{bg}}| > 1$.
\end{itemize}

For every type of homophilic structures, we perform a semantic aggregation on the background nodes within each community.
Specifically, the text attributes of background nodes are treated as messages to be aggregated, while the text attributes of target nodes serve as contextual information guiding this aggregation.
This operation can be regarded as a context‑aware aggregation over the set of background nodes, implemented through a frozen large language model $f_{\text{LLM}}(\cdot \ ; \theta)$.
Formally, the aggregation is defined as:
\begin{equation}
    \widetilde{R}_b^{(i)} = f_\text{LLM} \ 
    \Big( \ 
    \underbrace{\{\, R_{b} \bigm| v_b \in T_i \cap \mathcal{V}_{\text{bg}} \}}_{
    \text{\colorbox[rgb]{0.70,0.88,0.95}{background corpus}}}
    \Bigm|
    \underbrace{\{\, R_{t} \bigm| v_t \in T_i \cap \mathcal{V}_{\text{tg}} \}}_{
    \text{\colorbox[rgb]{1.0,0.80,0.78}{target context}}}\ ; \ \theta
    \ \Big),
    \label{Eq. LLM summary}
\end{equation}
where $\theta$ denotes the parameters of the LLM. During the aggregation process, we design distinct prompt templates for different types of communities. An illustration of the prompt template used for a Specific‑type community is presented in \boxref{Box: Specific Template}, while the remaining templates are included in the Appendix~\ref{app:Aggregation_Prompt_Template}.
Through this homophilic semantics aggregation, the LLM generates a condensed summary of the background nodes that is semantically consistent with the target nodes within each homophilic structure. After completing this aggregation for all communities, the original coding tree $\mathcal{T}^\ast$ is transformed into a semantically compressed coding tree, denoted as $\mathcal{T}^{\ast}_\text{SC}$, where each community’s background nodes are replaced by their aggregated semantic information while the target nodes remain preserved in the hierarchical structure.

\begin{tcolorbox}[colback=white,colframe=black,title=The Prompt Template for Specific Target type Community.]
\textbf{Specific Template:} Given a single target paper and a group of background papers from the same community within a citation graph. Each paper is represented by its title and abstract. The background papers, which are connected to the target paper and may share a common semantic focus or research topic with the target paper. \\[3pt]
Target paper:
\textcolor[rgb]{0.90,0.55,0.50}{\texttt{\{RAW\_TEXT\}}}\\[3pt]
\noindent Background papers:
\textcolor[rgb]{0.36,0.61,0.84}{\texttt{\{RAW\_TEXTS\}}}\\[3pt]
Task:
Summarize the content of those background papers that are semantically aligned with the target paper, focus on shared research disciplines, domains, directions, or topics. Please write the summary in a cohesive and formal academic style. \\[3pt]
Answer:
\label{Box: Specific Template}
\end{tcolorbox}

\subsection{Compressed Graph Reconstruction} \label{section4_3}
After performing the homophilic semantics aggregation, we reconstruct a compressed graph $\widetilde{\mathcal{G}}$ based on $\mathcal{T}^\ast_\text{SC}$. This reconstructed graph preserves both the homophilic structural patterns and the homophilic semantic information, while substantially reducing the overall scale of the graph.

We retain the four community types defined in Sec. \ref{section4_2}, and discard communities consisting solely of background nodes, as these nodes constitute structural noise with respect to homophily. Formally, let $T_i \subset \mathcal{V}$ denote the set of vertices contained in community $c_i$, where $c_i \in \mathcal{C}_\text{homo}$. We define the set of retained communities as 
$\mathcal{C}_\text{homo}^{\ast} = \{\, c_i \in \mathcal{C}_\text{homo} \mid T_i \cap \mathcal{V}_{\text{tg}} \neq \varnothing \,\}, \ |\mathcal{C}_\text{homo}^\ast| < |\mathcal{C}_\text{homo}|.$
For each retained community $c_i \in \mathcal{C}_\text{homo}^\ast$, all background nodes
$\{ v_b \mid v_b \in T_i \cap \mathcal{V}_{\text{bg}} \}$ are aggregated into a single condensed background node $\widetilde{v}_{b}^{(i)}$ through the LLM‑based semantic summarization. The raw text attributes of these background nodes $\{R_b \ |\ v_b\in T_i \cap\mathcal{V}_\text{bg}\}$ are aggregated into a condensed summary $\widetilde{R}_b^{(i)}$. 
Let the set of target nodes remain unchanged as $\mathcal{V}_{\text{tg}}$, and we define 
$\widetilde{\mathcal{V}}_{\text{bg}} = \{\, \widetilde{v}_{b}^{(i)} \mid c_i \in \mathcal{C}_\text{homo}^\ast, \ T_i \cap \mathcal{V}_{\text{bg}} \neq \varnothing \,\}$.
The node set of the compressed graph is then formally defined as
$\widetilde{\mathcal{V}} = \mathcal{V}_{\text{tg}} \cup \widetilde{\mathcal{V}}_{\text{bg}}$, which consists of all original target nodes and the newly aggregated background nodes. Since background nodes within each community are aggregated, it follows that $|\widetilde{\mathcal{V}}_{\text{bg}}| \ll |\mathcal{V}_{\text{bg}}|$.

Formally, for each background node $v_b$, let $g_{v_b}$ denote the number of edges connecting $v_b$ to nodes in the graph $\hat{\mathcal{G}}$.
During compression, all edges associated with the background nodes in a community $c_i$ are consolidated into the edges of the newly created aggregated node $\widetilde{v}_b^{(i)}$:
$g_{\widetilde{v}_b^{(i)}} =
\sum\limits_{v_b \in T_i \cap \mathcal{V}_{\text{bg}}} g_{v_b}.$

The edge set of the compressed graph is denoted by $\widetilde{\mathcal{E}}$.
For each original edge $(u,v) \in \hat{\mathcal{E}}$, we construct the corresponding edge in $\widetilde{\mathcal{E}}$ according to the following mapping rules:
\begin{equation}
    \begin{cases} 
        (u,v), & u,v \in \mathcal{V}_{\text{tg}}, \\[4pt] 

        (\widetilde{v}_{b}^{(i)}, \, u), & \ v \in T_i \cap \mathcal{V}_{\text{bg}}, u \in \mathcal{V}_{\text{tg}}, \\[4pt] 
        
        (u,\, \widetilde{v}_{b}^{(i)}), & u \in \mathcal{V}_{\text{tg}},\ v \in T_i \cap \mathcal{V}_{\text{bg}}, \\[4pt] 
        
        (\widetilde{v}_{b}^{(i)},\, \widetilde{v}_{b}^{(j)}), & u \in T_i \cap \mathcal{V}_{\text{bg}},\ v \in T_j \cap \mathcal{V}_{\text{bg}},\ i \neq j.
    \end{cases}
\end{equation}

All original edges connected to background nodes within the same community $c_i$ are therefore redirected to the corresponding condensed node $\widetilde{v}_{b}^{(i)}$.
This preserves the structural connectivity of target nodes and inter‑community relationships, while substantially reducing the size of the background node set.
Formally, the compressed graph is expressed as
$\widetilde{\mathcal{G}} = \bigl(\widetilde{\mathcal{V}},\, \widetilde{\mathcal{E}},\,\widetilde{\mathcal{R}}\bigr),$
where $\widetilde{\mathcal{E}}$ is derived from $\hat{\mathcal{E}}$ according to the mapping rules described above, thereby ensuring that the homophilic structures and semantics of the original graph are retained.
For raw text attributes, each condensed background node $\widetilde{v}_{b}^{(i)}$ collects the raw‑text attributes of its associated background nodes as
${R}_{\text{b}}^{(i)} = \bigl\{ R_b \mid v_b \in T_i \cap \mathcal{V}_{\text{bg}} \bigr\}$,
These raw texts are then summarized by the LLM to produce a single aggregated text $\widetilde{R}_{\text{b}}^{(i)}$ based on Eq. \ref{Eq. LLM summary}, that is then assigned to $\widetilde{v}_{b}^{(i)}$.
The complete raw‑text set for the compressed graph is therefore defined as $\widetilde{\mathcal{R}} = \bigl\{ R_t \mid v_t \in \mathcal{V}_{\text{tg}} \bigr\} \cup \bigl\{ \widetilde{R}_{\text{b}}^{(i)} \mid c_i \in \mathcal{C}_\text{homo}^\ast \bigr\}$,
where the first subset preserves the original texts of target nodes and the second subset contains the LLM‑summarized texts for the aggregated background nodes.

\subsection{LLM Understanding over Compressed Graph} \label{section4_4}
Based on the compressed reconstruction graph $\widetilde{\mathcal{G}}=(\widetilde{\mathcal{V}}, \widetilde{\mathcal{E}}, \widetilde{\mathcal{R}})$, we perform the downstream task of node classification.
Formally, for each target node $v_i \in \mathcal{V}_{\text{tg}}$ with $\mathcal{V}_{\text{tg}} \subset \widetilde{\mathcal{V}}$, we aim to predict its label $y_i$.
The classification function can be described as
$ f_{\text{class}}(v_i)=\hat{y}_i$.

For each target node $v_i$, its $1$-hop neighbor set is obtained from the reconstructed structure as $\mathcal{N}(v_i)=\{v_j|(v_i,v_j)\in\widetilde{\mathcal{E}}\}$ by Retrieval-Augmented Generation (RAG). We then design a task‑specific prompt by incorporating both the text attributes of the target node and its neighbors. An illustration of the prompt template for the OGBN‑ArXiv dataset is provided in \boxref{Box: Node Classification Template}. The resulting prompt is fed into the frozen large language model to obtain predicted label:
\begin{equation}
    \hat{y}_i = f_{\text{LLM}} \ \bigl(\ \text{Prompt}(R_{t_i}, \{\widetilde{R}_{j} \mid v_j \in \mathcal{N}(v_i)\});\ \theta \ \bigr),
\end{equation}
where $\theta$ denotes the parameters of the LLM. 
By selecting neighbors from the compressed graph that integrates both homophilic structure and aggregated semantics, the constructed prompt provides compact yet semantically coherent contextual information, thereby enhancing the effectiveness of zero‑shot node classification. 

\begin{tcolorbox}[colback=white,colframe=blue!30!black,title=The Prompt Template for node classification.]

\textbf{Classification Template:} Given a citation graph, the 0th node (target paper) has the following information: \textcolor[rgb]{0.90,0.55,0.50}{\texttt{\{TARGET\_RAW\_TEXT\}}}\\[3pt]
The target paper is connected to the following papers: \textcolor[rgb]{0.746,0.427,0.765}{\texttt{\{NEIGHBOR\_RAW\_TEXTS\}}}\\[3pt]
Question: Based on the features of the target paper and its citation network, please determine the most appropriate ArXiv CS sub-category for the target paper.\\[3pt]
Categories: \textcolor[rgb]{1.0,0.753,0.0}{\texttt{\{CATEGORY\_LIST\}}}.\\[3pt]
Please think about the categorization of the target paper in a structured manner, and only output the single most relevant category of the target paper. Do not give any reasoning or extra text for your answer.\\[3pt]
Answer:
\label{Box: Node Classification Template}

\end{tcolorbox}


\section{Experiments} \label{section5}
In this section, we first introduce the experimental settings, including the datasets, evaluation metrics, and baseline methods. Then, we evaluate $\text{HS}_2\text{C}$, with extensive experiments to answer the following research questions (\textbf{RQs}):
\begin{itemize}[leftmargin=*, itemsep=2pt, topsep=2pt, parsep=0pt, partopsep=0pt] 
    \item \textbf{RQ1:} How does $\text{HS}_2\text{C}$ perform compared to other baselines in the node-level classification task?
    \item \textbf{RQ2:} How does graph compression affect LLM performance, and why does it boost accuracy after compression?
    \item \textbf{RQ3:} How does the entropy-guided hierarchical structure partition and community-variant semantics aggregation contribute to node classification performance?
    \item \textbf{RQ4:} How does $\text{HS}_2\text{C}$ perform on large-scale graphs? 
    \item \textbf{RQ5:} Does $\text{HS}_2\text{C}$ effectively detect homophilic structures?
    \item \textbf{RQ6:} What else can $\text{HS}_2\text{C}$ do? Apart from node-level task, can $\text{HS}_2\text{C}$ extend to other graph tasks?
\end{itemize}

\subsection{Experimental settings}
\textbf{Datasets.}
To evaluate the performance of $\text{HS}_2\text{C}$, we conduct node-level classification tasks on 10 real-world datasets, OGBN-ArXiv~\cite{ogb_arxiv}, TAPE~\cite{tape}, Instagram~\cite{instagram}, Citeseer~\cite{citeseer}, Reddit~\cite{instagram}, and Product-subset~\cite{tape, taglas}. 
which span different domains, including citation networks, social networks, and co‑purchase networks. 
Detailed dataset statistics are summarized in Appendix~\ref{subsubapp: Node-level Dataset Description}. 

\vspace{2pt}
\noindent\textbf{Evaluation Metrics.} To quantitatively measure the compression effect, we introduce the Graph Compression Rate (GCR) defined as $\text{GCR}=|\widetilde{\mathcal{V}}| \ / \ |\mathcal{V}|,$
where $|\widetilde{\mathcal{V}}|$ and $|\mathcal{V}|$ denote the numbers of nodes in the compressed and original graphs, respectively. A lower GCR (or higher $1-\text{GCR}$) indicates a greater compression ratio. In addition, we use node classification accuracy (ACC) to measure the task performance. To jointly reflect both compression efficiency and classification performance, we further introduce a comprehensive metric, Graph Compression Index (GCI), defined as: $\text{GCI} = \frac{\text{ACC}}{\text{GCR}}$.
Higher GCI indicates that the method achieves better classification accuracy under graph compression. Furthermore, we normalize the GCI by selecting the standard GCN model as a reference, and compute the normalized index as $\overline{\text{GCI}} = \frac{\text{GCI}}{\text{GCI}_\text{GCN}}$, where $\text{GCI}_{\text{GCN}}$ denotes the GCI value obtained by the classical GCN model. 
Furthermore, to evaluate the compression rate, we encapsulate the compressed graph $ \widetilde{\mathcal{G}}$ into a torch\_geometric.data.Data object, denoted as \texttt{Data(X,\! EDGE\_TEXT,\! Y,\! RAW\_TEXT)}, which includes the node features $\widetilde{\mathcal{X}}$, edge connections $\widetilde{\mathcal{E}}$, target node labels $\mathcal{Y}$, and text attributes $\widetilde{\mathcal{R}}$. For traditional neighbor sample methods, the Data object is constructed by combining the target nodes $\mathcal{V}{\text{tg}}$ and their sampled neighbors $\mathcal{N}\text{sample}(\mathcal{V}_{\text{tg}})$, together with their corresponding features, edge connections, labels, and raw-text attributes.
\vspace{2pt}

\noindent\textbf{Baselines.}
We compare our proposed $\text{HS}_2\text{C}$ with 11 baseline methods on node-level task, which we generally divide into four categories, $(1)$ Graph Neural Network (GNN) Methods, including GCN~\cite{baseline_gcn_kipf2016gcn}, GAT~\cite{baseline_gat_velivckovic2017gat}, GIN~\cite{baseline_gin_xu2018gnn}, GraphSAGE~\cite{baseline_graphsage_hamilton2017graphsage}. $(2)$ Traditional Neighbor Sample Methods, including Random~\cite{Chen_GraphLLM2024KDD, rw_chen_icml2024llaga, rw_ye_EACL2024instructglm, rw_wu_icml2025when}, Degree~\cite{baseline_degree_ali2024degree}, Number, RAG~\cite{baseline_rag_li2025large}. $(3)$ Graph Skeleton Methods, including Skeleton $\alpha$, $\beta$ and $\gamma$~\cite{baseline_skeleton_cao2024graph}. The detailed information are summarized in Appendix~\ref{subapp: baselines}.

\noindent\textbf{Implementation Details.}
We run $\text{HS}_2\text{C}$ and all baselines under 5 different random seeds. We adopt 3 LLM backbones:
$\text{LLaMA}‑3.2‑3\text{B}‑\text{Instruct}$, 
$3.1‑8\text{B}‑\text{Instruct}$~\cite{grattafiori2024llama}, $\text{Vicuna}‑7\text{B}‑\text{v}1.5$~\cite{vicuna_model}.
Detailed implementation settings are provided in Appendix~\ref{app: Supplementary Experimental settings}.
For all RAG-based baselines, we use BERT-base-uncased~\cite{devlin2019bert} as the text encoder. 

\subsection{Overall Performance (RQ1)}
\begin{figure*}[h]
    \centering
    \includegraphics[width=\linewidth]{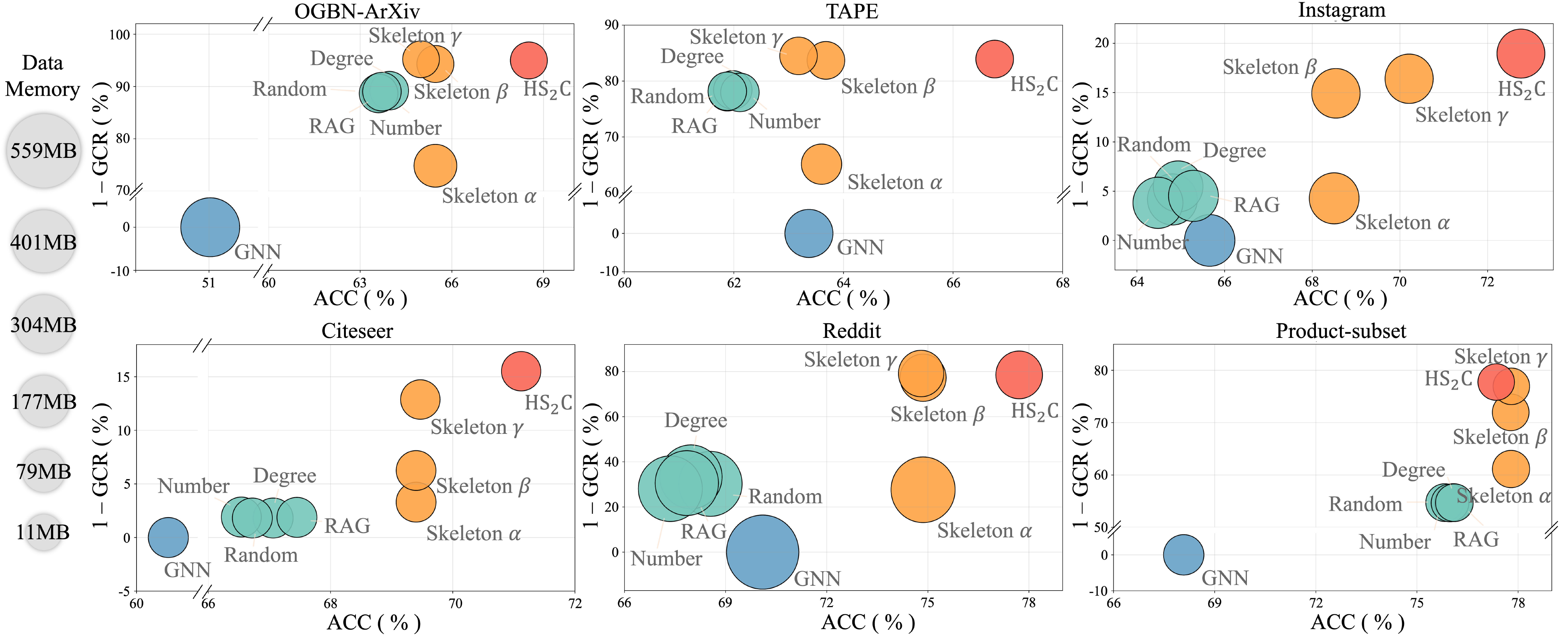}
    \caption{Comparison of Graph Compression Rate (GCR), classification accuracy (ACC), and data memory (MB).}
    \label{fig: compression all}
\end{figure*}

We record GCI and $\overline{\text{GCI}}$ in Table~\ref{tab:node_classification_gci}, and illustrate ACC and GCR in Figure~\ref{fig: compression all}, with circle size indicating memory consumption.
We have the following observations: 
\textbf{(1)} $\text{HS}_2\text{C}$ surpasses 4 representative GNNs by up to 18.66\% in $\overline{\text{GCI}}$, as it explicitly detects homophilic structures and aggregates semantically aligned information, whereas GNNs rely on full‑graph message passing with coarse text encodings that often retain redundant or noisy semantics.
\textbf{(2)} $\text{HS}_2\text{C}$ consistently outperforms traditional neighbor sampling methods. Random method, though widely used, ranks relatively low, as it completely ignores the structural and semantic relevance of neighbors to the target, which introduces excessive noise. While Degree and RAG strategies perform better by incorporating structural or semantic cues. Distinct from these methods, $\text{HS}_2\text{C}$ unifies global community‑level homophilic structure detection with semantic aggregation, achieving efficient compression and distilling the most informative context for target nodes. 
\textbf{(3)} Compared with graph skeleton methods, $\text{HS}_2\text{C}$ achieves a maximum improvement of 16.42 in $\overline{\text{GCI}}$. Skeleton $\alpha$, $\beta$, and $\gamma$ rely on locally defined structural equivalence criteria and adopt a uniform aggregation strategy. However, these methods overlook global topology, which constrains the quality of preserved structural and semantic information.
In contrast, $\text{HS}_2\text{C}$ adopts a global perspective by minimizing SE to obtain an optimal coding tree, thereby capturing the essential topology of the graph. On top of this hierarchical partition, $\text{HS}_2\text{C}$ explicitly identifies homophilic structures and categorizes communities into different types. For each type, $\text{HS}_2\text{C}$ applies tailored semantic aggregation to extract the neighbor information that is relevant to the target nodes while filtering out irrelevant semantics. 
Consequently, $\text{HS}_2\text{C}$ produces a more expressive compressed graph and significantly enhances task performance.

\begin{table*}[h]
\centering
\caption{\textbf{Graph Compression Index (GCI) Comparison}. The best results and second-best results are bolded and underlined respectively. The last column reports the mean $\overline{\text{GCI}}$, and the average ranking on each dataset. All performance improvements achieved by $\text{HS}_2\text{C}$ are statistically significant.}
\label{tab:node_classification_gci}

\vspace{-8pt}

\renewcommand{\arraystretch}{0.85}
\footnotesize 
\setlength{\tabcolsep}{1.5pt}
\begin{tabular*}{\textwidth}{@{\extracolsep{\fill}}rccccccccccccl}
\hline
\noalign{\vskip 1pt}
\multirow{2}{*}{} & \multicolumn{2}{c}{OGBN-ArXiv} & \multicolumn{2}{c}{TAPE} & \multicolumn{2}{c}{Citeseer} & \multicolumn{2}{c}{Instagram} & \multicolumn{2}{c}{Reddit} & \multicolumn{2}{c}{Product-subset} & \multirow{2}{*}{\begin{tabular}[c]{@{}c@{}}Avg. $\overline{\text{GCI}}$\\ (Rank)\end{tabular}} \\ 
\cline{2-13}
\noalign{\vskip 1pt}
& GCI & $\overline{\text{GCI}}$ & GCI & $\overline{\text{GCI}}$ & GCI & $\overline{\text{GCI}}$ & GCI & $\overline{\text{GCI}}$ & GCI & $\overline{\text{GCI}}$ & GCI & $\overline{\text{GCI}}$ & \\ 
\noalign{\vskip 1pt}
\hline
\noalign{\vskip 1pt}
GCN & 51.05   & 1.00  & 63.37  & 1.00 & 65.66 & 1.00 & 60.52 & 1.00 & 70.10  & 1.00 & 68.08  & 1.00 & 1.00 (23)  \\
GAT & 51.11   & 1.00  & 65.33  & 1.03 & 63.71 & 0.97 & 60.65 & 1.00 & 63.16  & 0.90 & 67.33  & 0.99 & 0.98 (24) \\
GIN & 47.28   & 0.93  & 58.27  & 0.92 & 65.25 & 0.99 & 52.42 & 0.87 & 69.29  & 0.99 & 67.72  & 0.99 & 0.95 (25) \\
GraphSAGE & 49.84   & 0.98  & 61.87  & 0.98 & 66.20 & 1.01 & 50.95 & 0.84 & 63.95  & 0.91 & 64.29  & 0.94 & 0.94 (26) \\ 
\noalign{\vskip 1pt}
\hline
\noalign{\vskip 1pt}
Random (3B) & 258.25 & 213.58 & 46.84 & 37.74 & 70.38 & 109.88 & 5.06 & 3.37 & 0.77 & 0.57 & 1.00 & 1.61 & 2.07 (16)  \\
Random (7B) & 187.94 & 81.13 & 42.39 & 28.98 & 66.68 & 53.80 & 3.68 & 1.28 & 0.70 & 0.44 & 0.95 & 0.79 & 1.31 (21)  \\
Random (8B) & 573.82 & 282.52 & 66.01 & 69.97 & 98.26 & 167.39 & 11.24 & 4.46 & 1.09 & 1.07 & 1.40 & 2.46 & 3.62 (9)  \\

Degree (3B) & 268.17 & 216.41 & 47.86 & 39.83 & 73.39 & 110.95 & 5.25 & 3.41 & 0.79 & 0.61 & 1.05 & 1.63 & 2.12 (14)  \\
Degree (7B) & 194.67 & 79.56 & 42.11 & 27.87 & 67.54 & 55.46 & 3.81 & 1.26 & 0.70 & 0.42 & 0.96 & 0.81 & 1.33 (19)  \\
Degree (8B) & 591.97 & 287.03 & 66.21 & 70.40 & 101.97 & 168.05 & 11.60 & 4.53 & 1.09 & 1.07 & 1.45 & 2.47 & 3.70 (7)  \\

Number (3B) & 247.56 & 211.54 & 45.62 & 38.74 & 69.71 & 109.85 & 4.85 & 3.34 & 0.75 & 0.59 & 0.99 & 1.61 & 2.02 (17)  \\
Number (7B) & 185.12 & 80.21 & 41.34 & 28.32 & 61.22 & 55.34 & 3.63 & 1.27 & 0.68 & 0.43 & 0.87 & 0.81 & 1.28 (22)  \\
Number (8B) & 561.71 & 281.95 & 65.68 & 69.37 & 93.67 & 167.53 & 11.00 & 4.45 & 1.09 & 1.06 & 1.34 & 2.46 & 3.57 (10)  \\

RAG (3B) & 261.21 & 214.58 & 46.50 & 38.73 & 72.02 & 110.14 & 5.12 & 3.39 & 0.77 & 0.59 & 1.03 & 1.62 & 2.08 (15)  \\ 
RAG (7B) & 192.06 & 80.52 & 42.23 & 28.95 & 65.12 & 55.51 & 3.76 & 1.27 & 0.70 & 0.44 & 0.93 & 0.82 & 1.32 (20)  \\ 
RAG (8B) & 580.45 & 283.60 & 66.54 & 70.65 & 97.73 & 167.99 & 11.37 & 4.48 & 1.10 & 1.08 & 1.39 & 2.47 & 3.65 (8)  \\ 
\noalign{\vskip 1pt}
\hline
\noalign{\vskip 1pt}
Skeleton ($\alpha$) (3B) & 126.68 & 127.68 & 47.21 & 33.39 & 69.51 & 125.53 & 2.48 & 2.01 & 0.78 & 0.51 & 0.99 & 1.84 & 1.44 (18)  \\
Skeleton ($\alpha$) (7B) & 96.74 & 53.11 & 43.13 & 17.10 & 62.92 & 64.86 & 1.90 & 0.84 & 0.71 & 0.26 & 0.90 & 0.95 & 0.93 (27)  \\
Skeleton ($\alpha$) (8B) & 260.08 & 182.57 & 70.86 & 72.49 & 103.43 & 200.15 & 5.09 & 2.88 & 1.17 & 1.10 & 1.48 & 2.94 & 2.44 (14)  \\

Skeleton ($\beta$) (3B) & 561.67 & 278.59 & 48.78 & 37.79 & 221.12 & 173.27 & 11.00 & 4.40 & 0.81 & 0.58 & 3.15 & 2.55 & 3.75 (6)  \\
Skeleton ($\beta$) (7B) & 446.64 & 113.46 & 44.49 & 19.36 & 200.20 & 89.88 & 8.75 & 1.79 & 0.74 & 0.29 & 2.86 & 1.32 & 2.62 (13)  \\
Skeleton ($\beta$) (8B) & 1144.21 & 389.04 & 73.11 & 81.58 & 329.02 & 277.53 & 22.41 & 6.14 & 1.21 & 1.24 & 4.69 & 4.08 & 6.63 (3)  \\

Skeleton ($\gamma$) (3B) & 663.65 & 305.48 & 53.78 & 44.16 & 235.42 & 216.74 & 13.00 & 4.82 & 0.89 & 0.67 & 3.36 & 3.18 & 4.32 (5)  \\
Skeleton ($\gamma$) (7B) & 531.05 & 119.39 & 47.55 & 32.04 & 220.96 & 109.82 & 10.40 & 1.88 & 0.79 & 0.49 & 3.15 & 1.61 & 3.05 (12)  \\
Skeleton ($\gamma$) (8B) & 1355.04 & 407.52 & 80.58 & 83.11 & 357.98 & 337.45 & 26.54 & 6.43 & 1.33 & 1.27 & 5.11 & 4.96 & 7.61 (2)  \\[1pt] 
\noalign{\vskip 1pt}
\hline
\noalign{\vskip 1pt}
$\text{H}\text{S}_2\text{C}$ (Ours) (3B) 
& 696.92 & 305.37 & 61.54 & 84.26 & 237.77 & 243.28 & 13.65 & 4.82 & 1.02 & 1.28 & 3.39 & 3.57 & 4.62 (4)  \\ 
$\text{H}\text{S}_2\text{C}$ (Ours) (7B) 
& 596.99 & 113.06 & 49.96 & 45.25 & 237.26 & 199.60 & 11.69 & 1.78 & 0.83 & 0.69 & 3.38 & 2.93 & 3.55 (11)   \\ 
$\text{H}\text{S}_2\text{C}$ (Ours) (8B) 
& \textbf{1363.98} & \textbf{415.35} & \textbf{86.11} & \textbf{88.50} & \textbf{363.23} & \textbf{347.77} & \textbf{26.72} & \textbf{6.55} & \textbf{1.42} & \textbf{1.35} & \textbf{5.18} & \textbf{5.11} & \textbf{7.72} \textbf{(1)}   \\ 
\hline
\end{tabular*}
\vspace{-15pt}
\end{table*}

\begin{table*}[h]
    \centering
    \caption{ \textbf{Node Classification Performance (\%) Comparison}. We report the mean and standard deviation for classification accuracy over 5 runs, and the best results are bolded. The last column records the average ranking on each dataset.}
    \label{subtab:node_classification_acc}
    \footnotesize 
    \renewcommand{\arraystretch}{0.9}
    \begin{tabular*}{\textwidth}{@{\extracolsep{\fill}}rccccccc}
    
    \toprule
    \toprule
    & OGBN-ArXiv & TAPE & Instagram & Citeseer & Reddit & Product-subset & Rank \\
    \midrule
    \noalign{\vskip 1pt}
    
    GCN 
    & 51.05\,$\pm$\,0.00 
    & 63.37\,$\pm$\,0.01 
    & 65.66\,$\pm$\,0.00 
    & 60.52\,$\pm$\,0.03 
    & 70.10\,$\pm$\,0.00
    & 68.08\,$\pm$\,0.00 
    & 9 \\
    \noalign{\vskip 1pt}
    GAT 
    & 51.11\,$\pm$\,0.00 
    & 65.33\,$\pm$\,0.00 
    & 63.71\,$\pm$\,0.00 
    & 60.65\,$\pm$\,0.05 
    & 63.16\,$\pm$\,0.02 
    & 67.33\,$\pm$\,0.00 
    & 10 \\
    \noalign{\vskip 1pt}
    GIN 
    & 47.28\,$\pm$\,0.01 
    & 58.27\,$\pm$\,0.08 
    & 65.25\,$\pm$\,0.00 
    & 52.42\,$\pm$\,0.08 
    & 69.29\,$\pm$\,0.00 
    & 67.72\,$\pm$\,0.00 
    & 11 \\
    \noalign{\vskip 1pt}
    GraphSAGE 
    & 49.84\,$\pm$\,0.00 
    & 61.87\,$\pm$\,0.01 
    & 66.20\,$\pm$\,0.00 
    & 50.95\,$\pm$\,0.09 
    & 63.95\,$\pm$\,0.01 
    & 64.29\,$\pm$\,0.04 
    & 12 \\

    \midrule
    \noalign{\vskip 1pt}
    Random (3B)
    & 28.65\,$\pm$\,0.27
    & 46.78\,$\pm$\,0.65 
    & 45.99\,$\pm$\,0.37 
    & 36.17\,$\pm$\,0.54 
    & 49.10\,$\pm$\,0.24
    & 49.77\,$\pm$\,0.53
    & 18 \\

    Random (7B)
    & 20.85\,$\pm$\,0.32
    & 17.77\,$\pm$\,0.11 
    & 41.62\,$\pm$\,0.28 
    & 27.77\,$\pm$\,0.44 
    & 46.52\,$\pm$\,0.95
    & 24.37\,$\pm$\,0.79
    & 23 \\

    Random (8B)
    & 63.66\,$\pm$\,0.42
    & 61.88\,$\pm$\,0.62 
    & 64.81\,$\pm$\,0.16 
    & 67.06\,$\pm$\,0.16 
    & 68.55\,$\pm$\,0.33
    & 75.82\,$\pm$\,0.14
    & 6 \\

    Degree (3B)
    & 28.97\,$\pm$\,0.00
    & 46.73\,$\pm$\,0.00 
    & 46.94\,$\pm$\,0.00 
    & 37.65\,$\pm$\,0.00 
    & 48.92\,$\pm$\,0.00
    & 50.23\,$\pm$\,0.00
    & 15 \\

    Degree (7B)
    & 21.03\,$\pm$\,0.00
    & 17.18\,$\pm$\,0.00 
    & 41.30\,$\pm$\,0.01 
    & 26.34\,$\pm$\,0.00 
    & 45.02\,$\pm$\,0.02
    & 25.11\,$\pm$\,0.00
    & 25 \\
    
    Degree (8B)
    & 63.95\,$\pm$\,0.04
    & 61.98\,$\pm$\,0.00 
    & 64.94\,$\pm$\,0.00 
    & 66.54\,$\pm$\,0.00 
    & 67.97\,$\pm$\,0.01
    & 76.08\,$\pm$\,0.05
    & 7 \\

    Number (3B)
    & 28.03\,$\pm$\,0.00
    & 46.60\,$\pm$\,0.00 
    & 44.79\,$\pm$\,0.00 
    & 37.26\,$\pm$\,0.00 
    & 50.13\,$\pm$\,0.00
    & 49.82\,$\pm$\,0.00
    & 17 \\

    Number (7B)
    & 20.96\,$\pm$\,0.00
    & 17.67\,$\pm$\,0.01 
    & 40.59\,$\pm$\,0.00 
    & 27.24\,$\pm$\,0.00 
    & 44.02\,$\pm$\,0.00
    & 25.10\,$\pm$\,0.01
    & 26 \\

    Number (8B)
    & 63.60\,$\pm$\,0.04
    & 62.11\,$\pm$\,0.02 
    & 64.48\,$\pm$\,0.00 
    & 66.72\,$\pm$\,0.01 
    & 67.36\,$\pm$\,0.09
    & 75.98\,$\pm$\,0.06
    & 8 \\

    RAG (3B)
    & 28.67\,$\pm$\,0.00
    & 46.82\,$\pm$\,0.00 
    & 45.62\,$\pm$\,0.00 
    & 36.98\,$\pm$\,0.00 
    & 50.00\,$\pm$\,0.00
    & 49.90\,$\pm$\,0.00
    & 16 \\

    RAG (7B)
    & 21.08\,$\pm$\,0.00
    & 17.57\,$\pm$\,0.00 
    & 41.43\,$\pm$\,0.00 
    & 27.64\,$\pm$\,0.00 
    & 45.21\,$\pm$\,0.00
    & 25.15\,$\pm$\,0.00
    & 24 \\

    RAG (8B)
    & 63.71\,$\pm$\,0.03
    & 61.88\,$\pm$\,0.02 
    & 65.29\,$\pm$\,0.02 
    & 67.45\,$\pm$\,0.01 
    & 67.85\,$\pm$\,0.07
    & 76.11\,$\pm$\,0.02
    & 5 \\

    \midrule
    \noalign{\vskip 1pt}

    Skeleton $\left( \alpha \right)$ (3B)
    & 31.86\,$\pm$\,0.02
    & 44.50\,$\pm$\,0.00 
    & 45.65\,$\pm$\,0.00 
    & 31.96\,$\pm$\,0.01 
    & 50.31\,$\pm$\,0.00
    & 48.79\,$\pm$\,0.00
    & 20 \\

    Skeleton $\left( \alpha \right)$ (7B)
    & 24.33\,$\pm$\,0.06
    & 18.61\,$\pm$\,0.03 
    & 41.71\,$\pm$\,0.00 
    & 16.37\,$\pm$\,0.00 
    & 45.54\,$\pm$\,0.01
    & 25.21\,$\pm$\,0.01
    & 28 \\
    
    Skeleton $\left( \alpha \right)$ (8B)
    & 65.41\,$\pm$\,0.00
    & 63.63\,$\pm$\,0.00 
    & 68.52\,$\pm$\,0.00 
    & 69.40\,$\pm$\,0.00 
    & 74.86\,$\pm$\,0.00
    & 77.79\,$\pm$\,0.00
    & 3 \\

    Skeleton $\left( \beta \right)$ (3B)
    & 32.03\,$\pm$\,0.07
    & 45.45\,$\pm$\,0.12 
    & 45.73\,$\pm$\,0.11 
    & 32.15\,$\pm$\,0.07 
    & 50.31\,$\pm$\,0.03
    & 48.56\,$\pm$\,0.02
    & 19 \\

    Skeleton $\left( \beta \right)$ (7B)
    & 25.47\,$\pm$\,0.17
    & 18.61\,$\pm$\,0.03
    & 41.71\,$\pm$\,0.13 
    & 16.47\,$\pm$\,0.02 
    & 45.55\,$\pm$\,0.04
    & 25.19\,$\pm$\,0.05
    & 27 \\
    
    Skeleton $\left( \beta \right)$ (8B)
    & 65.25\,$\pm$\,0.02
    & 63.47\,$\pm$\,0.03 
    & 68.54\,$\pm$\,0.00 
    & 69.40\,$\pm$\,0.00 
    & 74.86\,$\pm$\,0.00
    & 77.78\,$\pm$\,0.00
    & 4 \\

    Skeleton $\left( \gamma \right)$ (3B)
    & 31.83\,$\pm$\,0.07
    & 47.36\,$\pm$\,0.50 
    & 46.86\,$\pm$\,0.09 
    & 36.91\,$\pm$\,0.01 
    & 49.19\,$\pm$\,0.13
    & 49.97\,$\pm$\,0.11
    & 14 \\

    Skeleton $\left( \gamma \right)$ (7B)
    & 25.47\,$\pm$\,0.51
    & 18.51\,$\pm$\,0.07 
    & 41.43\,$\pm$\,0.03 
    & 26.79\,$\pm$\,0.33 
    & 46.17\,$\pm$\,0.10
    & 25.32\,$\pm$\,0.08
    & 22 \\
        
    Skeleton $\left( \gamma \right)$ (8B)
    & 64.99\,$\pm$\,0.00
    & 63.18\,$\pm$\,0.00 
    & 70.21\,$\pm$\,0.00 
    & 69.47\,$\pm$\,0.00 
    & 74.80\,$\pm$\,0.00
    & \textbf{77.80}\,$\pm$\,\textbf{0.00}
    & 2 \\
    
    \midrule
    \noalign{\vskip 1pt}
    
    $\text{H}\text{S}_2\text{C}$ (Ours) (3B)
    & 35.01\,$\pm$\,0.00
    & 49.08\,$\pm$\,0.04 
    & 52.00\,$\pm$\,0.02
    & 68.28\,$\pm$\,0.00
    & 50.87\,$\pm$\,0.02
    & 54.11\,$\pm$\,0.00
    & 13 \\

    $\text{H}\text{S}_2\text{C}$ (Ours) (7B)
    & 29.99\,$\pm$\,0.02
    & 18.22\,$\pm$\,0.00 
    & 42.21\,$\pm$\,0.05
    & 36.67\,$\pm$\,0.00
    & 50.76\,$\pm$\,0.02
    & 44.40\,$\pm$\,0.03
    & 21 \\
    
    $\text{H}\text{S}_2\text{C}$ (Ours) (8B)
    & \textbf{68.52}\,$\pm$\,\textbf{0.00}
    & \textbf{66.75}\,$\pm$\,\textbf{0.03} 
    & \textbf{72.76}\,$\pm$\,\textbf{0.00}
    & \textbf{71.72}\,$\pm$\,\textbf{0.37}
    & \textbf{77.71}\,$\pm$\,\textbf{0.00}
    & 77.35\,$\pm$\,{0.00}
    & \textbf{1} \\

\bottomrule
\bottomrule
\end{tabular*}
\end{table*}

\subsection{Case Study (RQ2)}\label{Scetion5_3}
\begin{figure*}[h]
    \centering
    \setlength{\abovecaptionskip}{2pt}
    \includegraphics[width=1.0\linewidth]{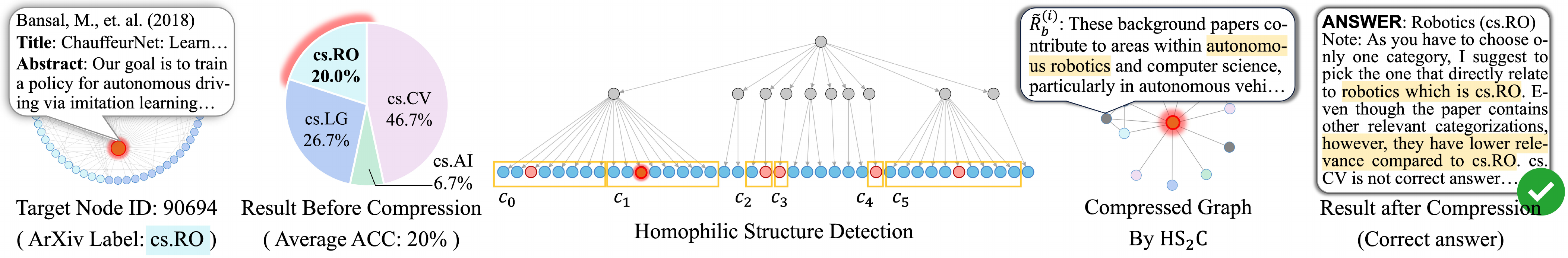}
    \captionsetup{aboveskip=0pt, belowskip=20pt}
    \caption{Node-level Case study on OGBN‑ArXiv dataset comparing the neighborhood of a target node before and after applying $\text{HS}_2\text{C}$. The left side shows the dense neighbors of target node in the original graph $\mathcal{G}$. Right: the compressed neighbors after homophilic structure detection and semantic aggregation in $\widetilde{\mathcal{G}}$.}
    \label{fig: Case_Study_1}
\end{figure*}

To further illustrate how $\text{H}\text{S}_2\text{C}$ enhances classification by refining both structural and semantic information, we present a case study on the OGBN‑ArXiv dataset, as shown in Figure~\ref{fig: Case_Study_1}. 
The left side depicts the neighborhood distribution of a target node before compression, where nodes with different colors indicate different classes. The target node (ID 90694, labeled as cs.RO) is surrounded by a dense set of neighbors, most of which belong to different categories such as cs.CV and cs.LG. Specifically, 73.33\% of its neighbors are of different classes, leading to noisy and misleading contextual information. When applying Random sampling, which is widely used in existing studies, such noisy neighbors significantly interfere with the classification process. Across 15 runs with different random seeds, the classification accuracy for this target node reaches only 20\%.
In contrast, as shown on the right side, $\text{H}\text{S}_2\text{C}$ derives a natural hierarchical partition that groups structurally similar nodes into communities.
Subsequently, homophilic semantic aggregation effectively filters out redundant neighbor information. As reflected in the output, this compressed context enables LLM to classify the target node with much higher accuracy and stronger confidence.

\subsection{Ablation Study (RQ3)} \label{sec: ablation study}
\begin{figure}[h]
    \centering
    \setlength{\abovecaptionskip}{2pt}
    \includegraphics[width=.5\linewidth]{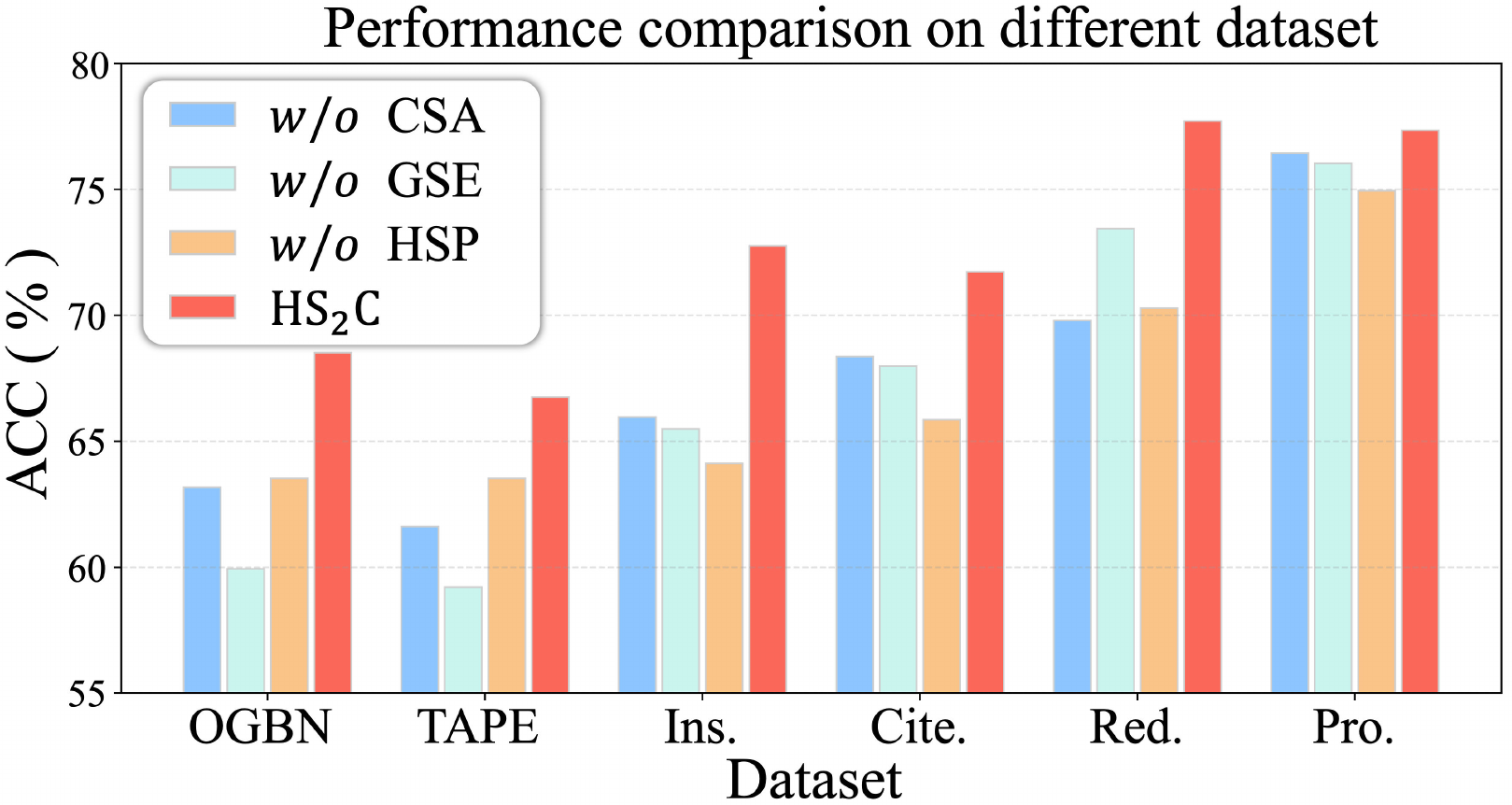}
    \captionsetup{aboveskip=2pt, belowskip=5pt}
    \caption{Ablation study results for $\text{HS}_2\text{C}$'s 3 key components.}
    \label{fig: Ablation study_1}
    \vspace{-12pt}
\end{figure}

To verify the effectiveness of each key component in $\text{HS}_2\text{C}$, we conduct an ablation study on its 3 major modules. Specifically,
we decompose $\text{HS}_2\text{C}$ into following variants: 
(1) $w/o$ GSE. Removing GSE and perform hierarchical partition directly on original $\mathcal{G}$. 
(2) $w/o$ HCP. Removing HCP, 
forming a local context for each target node with randomly sampled neighbors, 
constructing local contexts via random neighbor sampling, 
and applying the LLMNodeBed~\citep{rw_wu_icml2025when} summary template for semantic aggregation. 
(3) $w/o$ CSA. Retaining homophilic structure detection but directly concatenating the raw textual information of adjacent nodes. Each variant is run with 5 seeds, results are reported in Figure~\ref{fig: Ablation study_1}. 

We observe that removing GSE leads to a accuracy drop, confirming that structure enhancement provides richer 
connectivity cues for subsequent partitioning.
Comparing $w/o$ GSE and $w/o$ HCP further highlights that structure-aware partitioning is crucial for capturing global topological patterns, while simple neighbor sampling fails to extract meaningful structures.
Additionally, the performance gap between $w/o$ HCP and $w/o$ CSA demonstrates the importance of semantic aggregation on top of structural partitioning. 
Finally, the superior performance of the complete $\text{HS}_2\text{C}$ confirms that homophilic structure detection and homophilic semantic aggregation are mutually reinforcing. Their synergy enables accurate neighbor selection and semantic distillation, resulting in a compressed graph that preserves both structural coherence and contextual richness.

\subsection{Scalability on very Large Graph (RQ4)} \label{sec: Large-scale Graph Performance}
To assess the scalability of $\text{HS}_2\text{C}$, we further evaluate its performance on 2 real-world large-scale datasets, Book and DBLP, released by HTAG~\citep{htag_dataset}, where DBLP contains approximately 2M nodes and 30M edges. Besides the original heterogeneous graphs (Book-Hete, DBLP-Hete), we also construct corresponding homogeneous variants, Book-Homo and DBLP-Homo, by retaining only the citation/similarity links. We evaluate all methods using 3 LLM backbones and report the average GCI in Table~\ref{maintab:large_graph_performance}. The results show that $\text{HS}_2\text{C}$ consistently achieves the best performance, demonstrating its effectiveness and robustness on large-scale graphs. 

\begin{table*}[h]
    \centering
    \caption{ \textbf{Node Classification Performance (\%) Comparison on very Large Graph}. We report the mean and standard deviation for classification accuracy over 5 runs, and the best results are bolded. The last column records the average ranking on each dataset.}
    \label{maintab:large_graph_performance}
    \vspace{-8pt}
    \footnotesize 
    \setlength{\tabcolsep}{1.5pt}
    \renewcommand{\arraystretch}{1.0}
    \begin{tabular*}{\textwidth}{@{\extracolsep{\fill}}rccccccccc}
    
    \toprule
    \toprule
    \noalign{\vskip -1pt}
    & \multicolumn{2}{c}{Book-Hete} & \multicolumn{2}{c}{Book-Homo} & \multicolumn{2}{c}{DBLP-Hete} & \multicolumn{2}{c}{DBLP-Homo} & \multirow{2}{*}{Rank} \\
    \cline{2-9}
    \noalign{\vskip 2pt}
    & ACC & GCI & ACC & GCI & ACC & GCI & ACC & GCI & \\ 
    \noalign{\vskip -1pt}
    \midrule
    \noalign{\vskip -1pt}

    Random (3B)
    & 40.06\,$\pm$\,0.29
    & 998.59
    & 47.96\,$\pm$\,0.13 
    & 952.16
    & 14.78\,$\pm$\,0.40 
    & 265.78
    & 13.67\,$\pm$\,0.14 
    & 190.61
    & 13 \\

    Random (7B)
    & 31.66\,$\pm$\,0.39
    & 789.07
    & 38.31\,$\pm$\,0.22 
    & 760.55
    & 12.58\,$\pm$\,0.63 
    & 226.22
    & 11.83\,$\pm$\,0.19 
    & 164.95
    & 21 \\

    Random (8B)
    & 43.89\,$\pm$\,0.25
    & 1093.93
    & 47.86\,$\pm$\,0.15 
    & 950.17
    & 17.40\,$\pm$\,0.28 
    & 312.90
    & 20.24\,$\pm$\,0.14 
    & 282.22
    & 6 \\

    Degree (3B)
    & 40.14\,$\pm$\,0.00
    & 1054.08
    & 48.08\,$\pm$\,0.00 
    & 1148.00
    & 14.85\,$\pm$\,0.00 
    & 295.06
    & 13.51\,$\pm$\,0.00 
    & 202.61
    & 12 \\

    Degree (7B)
    & 32.28\,$\pm$\,0.00
    & 847.68
    & 38.07\,$\pm$\,0.00 
    & 908.99
    & 12.55\,$\pm$\,0.00
    & 249.36
    & 11.96\,$\pm$\,0.00 
    & 179.36
    & 20 \\
    
    Degree (8B)
    & 43.12\,$\pm$\,0.00
    & 1132.34
    & 47.19\,$\pm$\,0.00 
    & 1126.75
    & 19.05\,$\pm$\,0.00 
    & 378.52
    & 20.06\,$\pm$\,0.00 
    & 300.84
    & 5 \\

    Number (3B)
    & 39.36\,$\pm$\,0.00
    & 945.41
    & 48.13\,$\pm$\,0.00 
    & 1052.78
    & 14.68\,$\pm$\,0.00 
    & 252.63
    & 13.48\,$\pm$\,0.00 
    & 182.58
    & 14 \\

    Number (7B)
    & 31.72\,$\pm$\,0.00
    & 761.90
    & 37.03\,$\pm$\,0.00 
    & 809.98
    & 11.87\,$\pm$\,0.00 
    & 204.27
    & 12.18\,$\pm$\,0.00 
    & 164.97
    & 24 \\

    Number (8B)
    & 43.51\,$\pm$\,0.00
    & 1045.09
    & 47.47\,$\pm$\,0.00
    & 1038.35
    & 19.03\,$\pm$\,0.00 
    & 327.49
    & 20.06\,$\pm$\,0.00 
    & 271.70
    & 3 \\

    RAG (3B)
    & 40.40\,$\pm$\,0.00
    & 1021.74
    & 48.71\,$\pm$\,0.00 
    & 1108.41
    & 14.67\,$\pm$\,0.00 
    & 280.39
    & 13.66\,$\pm$\,0.00 
    & 192.03
    & 11 \\

    RAG (7B)
    & 31.93\,$\pm$\,0.00
    & 807.53
    & 37.62\,$\pm$\,0.00 
    & 856.06
    & 12.52\,$\pm$\,0.00 
    & 239.30
    & 11.63\,$\pm$\,0.00 
    & 163.49
    & 22 \\

    RAG (8B)
    & 43.23\,$\pm$\,0.00
    & 1093.31
    & 47.80\,$\pm$\,0.00 
    & 1087.71
    & 18.43\,$\pm$\,0.00 
    & 352.26
    & 20.15\,$\pm$\,0.00 
    & 283.27
    & 4 \\
    \noalign{\vskip -1pt}
    \midrule
    \noalign{\vskip -1pt}

    Skeleton $\left( \alpha \right)$ (3B)
    & 46.88\,$\pm$\,0.04
    & 1073.53
    & 48.60\,$\pm$\,0.00 
    & 1099.48
    & 11.24\,$\pm$\,0.00 
    & 81.21
    & 13.34\,$\pm$\,0.01 
    & 66.01
    & 10 \\

    Skeleton $\left( \alpha \right)$ (7B)
    & 38.73\,$\pm$\,0.04
    & 886.98
    & 39.17\,$\pm$\,0.12 
    & 886.24
    & 13.61\,$\pm$\,0.03 
    & 98.33
    & 14.72\,$\pm$\,0.01 
    & 72.82
    & 16 \\
    
    Skeleton $\left( \alpha \right)$ (8B)
    & 42.11\,$\pm$\,0.01
    & 964.30
    & 43.90\,$\pm$\,0.00 
    & 993.26
    & 16.08\,$\pm$\,0.02 
    & 116.18
    & 19.73\,$\pm$\,0.01 
    & 97.60
    & 8 \\

    Skeleton $\left( \beta \right)$ (3B)
    & 46.73\,$\pm$\,0.03
    & 4368.30
    & 48.59\,$\pm$\,0.00 
    & 1659.06
    & 12.73\,$\pm$\,0.00 
    & 387.50
    & 13.35\,$\pm$\,0.04 
    & 331.25
    & 9 \\

    Skeleton $\left( \beta \right)$ (7B)
    & 38.51\,$\pm$\,0.03
    & 3600.21
    & 38.64\,$\pm$\,0.04
    & 1319.16
    & 12.92\,$\pm$\,0.04 
    & 393.28
    & 14.78\,$\pm$\,0.00 
    & 366.95
    & 18 \\
    
    Skeleton $\left( \beta \right)$ (8B)
    & 42.09\,$\pm$\,0.01
    & 3934.79
    & 43.85\,$\pm$\,0.01 
    & 1497.22
    & 16.92\,$\pm$\,0.02 
    & 575.92
    & 19.68\,$\pm$\,0.00 
    & 488.49
    & 7 \\

    Skeleton $\left( \gamma \right)$ (3B)
    & 33.89\,$\pm$\,0.03
    & 3970.73
    & 41.99\,$\pm$\,0.00 
    & 1799.27
    & 14.79\,$\pm$\,0.03 
    & 717.94
    & 14.91\,$\pm$\,0.01 
    & 397.39
    & 17 \\

    Skeleton $\left( \gamma \right)$ (7B)
    & 29.46\,$\pm$\,0.06
    & 3452.03
    & 34.98\,$\pm$\,0.05 
    & 1498.85
    & 15.01\,$\pm$\,0.00 
    & 728.62
    & 13.77\,$\pm$\,0.01 
    & 367.09
    & 23 \\
        
    Skeleton $\left( \gamma \right)$ (8B)
    & 29.68\,$\pm$\,0.03
    & 3477.34
    & 37.38\,$\pm$\,0.02 
    & 1601.70
    & 20.06\,$\pm$\,0.04 
    & 973.75
    & 19.04\,$\pm$\,0.06 
    & 507.37
    & 15 \\
    \noalign{\vskip -1pt}
    \midrule
    \noalign{\vskip -1pt}
    
    $\text{H}\text{S}_2\text{C}$ (Ours) (3B)
    & \textbf{54.31}\,$\pm$\,\textbf{0.00}
    & \textbf{5335.03}
    & \textbf{49.22}\,$\pm$\,\textbf{0.00} 
    & \textbf{3630.79}
    & 19.68\,$\pm$\,0.00
    & 821.28
    & 17.71\,$\pm$\,0.02
    & 366.07
    & \textbf{1} \\

    $\text{H}\text{S}_2\text{C}$ (Ours) (7B)
    & 32.79\,$\pm$\,0.02
    & 3221.06
    & 36.70\,$\pm$\,0.07
    & 2794.27
    & 16.77\,$\pm$\,0.00
    & 699.42
    & 15.17\,$\pm$\,0.05
    & 313.57
    & 3 \\
    
    $\text{H}\text{S}_2\text{C}$ (Ours) (8B)
    & 45.89\,$\pm$\,0.00
    & 4507.91
    & 48.26\,$\pm$\,0.01
    & 3559.97
    & \textbf{27.91}\,$\pm$\,\textbf{0.06}
    & \textbf{1164.73}
    & \textbf{25.75}\,$\pm$\,\textbf{0.01}
    & \textbf{532.18}
    & 2 \\
    \noalign{\vskip -1pt}
\bottomrule
\bottomrule
\end{tabular*}
\end{table*}

\begin{figure*}[h]
    \centering
    \includegraphics[width=\linewidth]{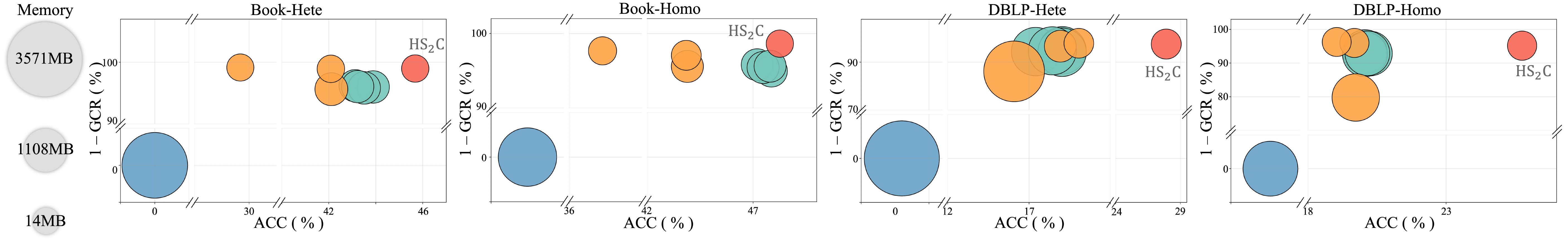}
    \caption{Comparison of Graph Compression Rate (GCR) on very large graph, classification accuracy (ACC), and data memory (MB).}
    \label{supfig: compression large graph}
\end{figure*}

\subsection{Effectiveness of Homophily Detection (RQ5)} \label{sec: Homophilic Structure Detection Performance}
To evaluate the effectiveness of homophilic structure detection performance, we define a homophily-based metric to measure how well the detected communities align with the inherent label distribution of nodes. Given a set of communities $\mathcal{C}_\text{homo}$ derived from the coding tree, each community $c_i \in \mathcal{C}_\text{homo}$ is associated with a vertex set $T_i \subset \mathcal{V}$. Let $y_k \in \mathcal{Y}$ denote the category label of node $v_k \in T_i$. We first compute the majority label $Y_i^{\max}$ within each community, $Y_i^{\max} = \arg\max_{Y \in \mathcal{Y}} \sum_{v_k \in T_i} \mathbb{I}(y_k = Y),$
where $\mathbb{I}(\cdot)$ is the indicator function. Based on this, the homophily score of each community $c_i$ is defined as: $H_i = \frac{1}{|T_i|} \sum_{v_k \in T_i} \mathbb{I}(y_k = Y_i^{\max}),$ 
which reflects the proportion of nodes in $T_i$ sharing the dominant label. 
To assess the overall performance, we compute the mean homophily score across all detected communities, $H_S = \frac{1}{|\mathcal{C}_\text{homo}|} \sum_{c_i \in \mathcal{C}_\text{homo}} H_i.$
A higher $H_S$ indicates that the community structure is more consistent with node label distributions, reflecting stronger homophily. 

We compare our proposed Hierarchical Structure Partition (HSP) module with traditional neighbor sampling baselines (Random, Degree, Number, and RAG), where the set of selected neighbors for each target node are treated as a local community. In addition, we evaluate against classical community partitioning methods such as Louvain and Leiden. As shown in Table~\ref{subtab:homo_score}, our method significantly outperforms other methods. The detected communities exhibit higher label consistency, reflecting a stronger ability to capture the intrinsic structural regularities of the graph. By adopting a global perspective and minimizing SE to construct an optimal coding tree, our method effectively preserves essential topological patterns while suppressing structural noise. These results validate the effectiveness of our method in discovering homophilc structures. 

\begin{table*}[h]
\centering
\caption{Homophily score comparison across methods.}
\label{subtab:homo_score}
\vspace{-8pt}
\footnotesize  
\renewcommand{\arraystretch}{0.9}
\begin{tabular*}{\textwidth}{@{\extracolsep{\fill}}lcccccccccc}
    \hline
    \noalign{\vskip 2pt}
    \textbf{Method} & \textbf{OGBN} & \textbf{TAPE} & \textbf{Ins.} & \textbf{Cite.} & \textbf{Red.} & \textbf{Pro.} & \textbf{B.Hete} & \textbf{B.Homo} & \textbf{D.Hete} & \textbf{D.Homo} \\
    \hline
    \noalign{\vskip 1pt}

    Original & 0.3378 & 0.1947 & 0.4925 & 0.6972 & 0.5400 & 0.7583 & 0.4082 & 0.5608 & 0.2436 & 0.6223 \\
    Random   & 0.3376 & 0.1947 & 0.4920 & 0.6972 & 0.5389 & 0.7579 & 0.4086 & 0.5610 & 0.2436 & 0.6224 \\
    Degree   & 0.3384 & 0.1947 & 0.4939 & 0.6972 & 0.5502 & 0.7581 & 0.4101 & 0.5634 & 0.2388 & 0.6186 \\
    Number   & 0.3378 & 0.1946 & 0.4926 & 0.6972 & 0.5397 & 0.7580 & 0.4034 & 0.5601 & 0.2349 & 0.6217 \\
    RAG      & 0.3386 & 0.1948 & 0.4931 & 0.6972 & 0.5401 & 0.7581 & 0.4217 & 0.5636 & 0.2898 & 0.6226 \\
    Louvain  & 0.8643 & 0.9238 & 0.9745 & 0.5982 & 0.8327 & 0.8049 & 0.3849 & 0.8628 & 0.3227 & 0.5348 \\
    Leiden   & 0.8638 & 0.9239 & \textbf{0.9760} & 0.5967 & 0.8407 & 0.8045 & 0.3949 & 0.8628 & 0.3227 & 0.5348 \\
    \hline
    \noalign{\vskip 2pt}
    Ours     & \textbf{0.9870} & \textbf{0.9425} & 0.9704 & \textbf{0.8344} & \textbf{0.9577} & \textbf{0.8964} & \textbf{0.8647} & \textbf{0.8849} & \textbf{0.9215} & \textbf{0.9573} \\
    \noalign{\vskip -1pt}
    \hline
\end{tabular*} 
\vspace{-10pt}
\end{table*}

\subsection{Task Generalizability Performance (RQ6)} \label{sec: Graph-level Task Performance}
\begin{figure}[h]
    \centering
    \includegraphics[width=.45\linewidth]{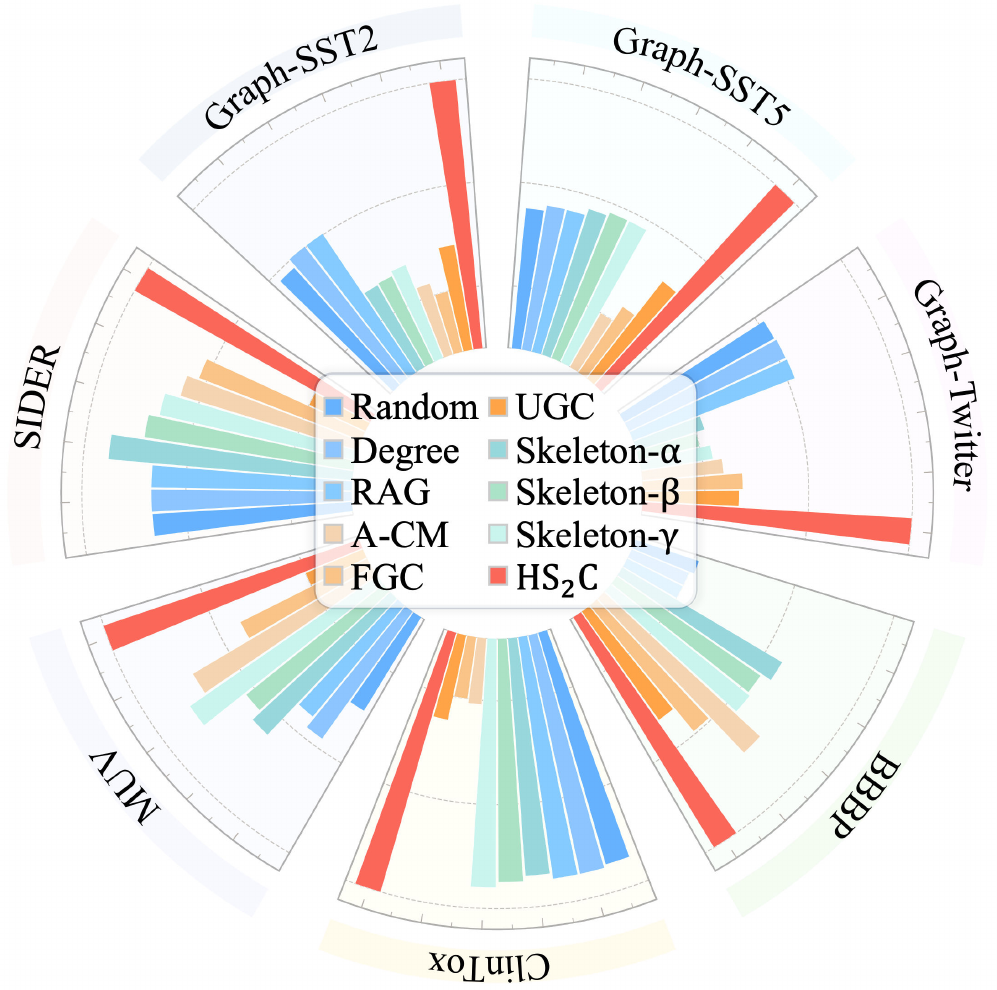}
    \captionsetup{aboveskip=2pt, belowskip=5pt}
    \caption{Graph-level task performance Comparison.}
    \vspace{-10pt}
    \label{fig: Graph_level}
\end{figure}

To investigate the task generalization of $\text{HS}_2\text{C}$, we further extend it to graph-level classification on 7 benchmarks spanning sentiment analysis and molecular property prediction. We evaluate 3 LLM backbones and compare against traditional neighbor sampling methods, Graph Skeleton methods, and Graph Coarsening methods. Figure~\ref{fig: Graph_level} partially reports the averaged performance (ACC or ROC-AUC) over the 3 backbones, where $\text{HS}_2\text{C}$ consistently achieves the best results, demonstrating its effectiveness on graph-level tasks. 

\begin{table*}[htbp]
    \centering
    \caption{\textbf{Graph-level Classification Performance (\%) Comparison}. We report the mean and standard deviation for classification ROC-AUC over 5 runs, and the best results are bolded. The last column records the average ranking on each dataset.}
    \label{subtab:graph_classification_acc}
    \vspace{-8pt}
    \resizebox{\textwidth}{!}{
    \begin{tabular}{rccccccc}
    \toprule
    & Graph-SST2 & Graph-SST5 & Graph-Twitter & BBBP & ClinTox & MUV & SIDER 
    \\

    \midrule

    Random (3B)
    & \multicolumn{1}{c}{\normalsize{55.78}\,$\pm$\,\scriptsize{0.06}} 
    & \multicolumn{1}{c}{\normalsize{19.94}\,$\pm$\,\scriptsize{0.37}} 
    & \multicolumn{1}{c}{\normalsize{37.66}\,$\pm$\,\scriptsize{2.46}}
    & \multicolumn{1}{c}{\normalsize{36.21}\,$\pm$\,\scriptsize{6.27}} 
    & \multicolumn{1}{c}{\normalsize{7.81}\,$\pm$\,\scriptsize{0.37}} 
    & \multicolumn{1}{c}{\normalsize{46.28}\,$\pm$\,\scriptsize{6.61}} 
    & \multicolumn{1}{c}{\normalsize{54.64}\,$\pm$\,\scriptsize{0.47}}  
    \\
    Random (7B)
    & \multicolumn{1}{c}{\normalsize{55.32}\,$\pm$\,\scriptsize{0.03}} 
    & \multicolumn{1}{c}{\normalsize{23.21}\,$\pm$\,\scriptsize{1.25}} 
    & \multicolumn{1}{c}{\normalsize{27.30}\,$\pm$\,\scriptsize{3.06}} 
    & \multicolumn{1}{c}{\normalsize{34.89}\,$\pm$\,\scriptsize{14.53}} 
    & \multicolumn{1}{c}{\normalsize{11.89}\,$\pm$\,\scriptsize{3.17}} 
    & \multicolumn{1}{c}{\normalsize{47.92}\,$\pm$\,\scriptsize{8.78}}
    & \multicolumn{1}{c}{\normalsize{55.16}\,$\pm$\,\scriptsize{0.52}}   
    \\
    Random (8B)
    & \multicolumn{1}{c}{\normalsize{74.09}\,$\pm$\,\scriptsize{1.96}} 
    & \multicolumn{1}{c}{\normalsize{21.73}\,$\pm$\,\scriptsize{0.58}} 
    & \multicolumn{1}{c}{\normalsize{50.99}\,$\pm$\,\scriptsize{1.73}}
    & \multicolumn{1}{c}{\normalsize{43.86}\,$\pm$\,\scriptsize{9.99}} 
    & \multicolumn{1}{c}{\normalsize{52.16}\,$\pm$\,\scriptsize{2.05}} 
    & \multicolumn{1}{c}{\normalsize{53.53}\,$\pm$\,\scriptsize{1.05}} 
    & \multicolumn{1}{c}{\normalsize{57.15}\,$\pm$\,\scriptsize{0.47}} 
    \\

    Degree (3B)
    & \multicolumn{1}{c}{\normalsize{55.79}\,$\pm$\,\scriptsize{0.06}} 
    & \multicolumn{1}{c}{\normalsize{20.15}\,$\pm$\,\scriptsize{0.18}} 
    & \multicolumn{1}{c}{\normalsize{37.70}\,$\pm$\,\scriptsize{0.01}}
    & \multicolumn{1}{c}{\normalsize{36.50}\,$\pm$\,\scriptsize{4.41}} 
    & \multicolumn{1}{c}{\normalsize{8.25}\,$\pm$\,\scriptsize{0.01}} 
    & \multicolumn{1}{c}{\normalsize{50.59}\,$\pm$\,\scriptsize{6.24}} 
    & \multicolumn{1}{c}{\normalsize{54.67}\,$\pm$\,\scriptsize{0.41}} 
    \\
    Degree (7B)
    & \multicolumn{1}{c}{\normalsize{55.31}\,$\pm$\,\scriptsize{0.01}} 
    & \multicolumn{1}{c}{\normalsize{22.94}\,$\pm$\,\scriptsize{0.22}} 
    & \multicolumn{1}{c}{\normalsize{27.38}\,$\pm$\,\scriptsize{0.00}}
    & \multicolumn{1}{c}{\normalsize{33.99}\,$\pm$\,\scriptsize{6.44}} 
    & \multicolumn{1}{c}{\normalsize{12.14}\,$\pm$\,\scriptsize{2.36}} 
    & \multicolumn{1}{c}{\normalsize{48.80}\,$\pm$\,\scriptsize{2.95}}
    & \multicolumn{1}{c}{\normalsize{55.17}\,$\pm$\,\scriptsize{0.52}}   
    \\
    Degree (8B)
    & \multicolumn{1}{c}{\normalsize{75.52}\,$\pm$\,\scriptsize{0.01}} 
    & \multicolumn{1}{c}{\normalsize{22.53}\,$\pm$\,\scriptsize{0.15}} 
    & \multicolumn{1}{c}{\normalsize{51.07}\,$\pm$\,\scriptsize{0.00}} 
    & \multicolumn{1}{c}{\normalsize{43.61}\,$\pm$\,\scriptsize{18.36}} 
    & \multicolumn{1}{c}{\normalsize{52.25}\,$\pm$\,\scriptsize{2.10}} 
    & \multicolumn{1}{c}{\normalsize{53.41}\,$\pm$\,\scriptsize{0.36}} 
    & \multicolumn{1}{c}{\normalsize{57.13}\,$\pm$\,\scriptsize{0.43}}  
    \\

    RAG (3B)
    & \multicolumn{1}{c}{\normalsize{55.79}\,$\pm$\,\scriptsize{0.06}} 
    & \multicolumn{1}{c}{\normalsize{20.16}\,$\pm$\,\scriptsize{0.16}} 
    & \multicolumn{1}{c}{\normalsize{37.80}\,$\pm$\,\scriptsize{0.01}}
    & \multicolumn{1}{c}{\normalsize{36.50}\,$\pm$\,\scriptsize{4.41}} 
    & \multicolumn{1}{c}{\normalsize{8.25}\,$\pm$\,\scriptsize{0.01}} 
    & \multicolumn{1}{c}{\normalsize{49.29}\,$\pm$\,\scriptsize{2.29}} 
    & \multicolumn{1}{c}{\normalsize{54.67}\,$\pm$\,\scriptsize{0.41}} 
    \\
    RAG (7B)
    & \multicolumn{1}{c}{\normalsize{55.31}\,$\pm$\,\scriptsize{0.01}} 
    & \multicolumn{1}{c}{\normalsize{22.90}\,$\pm$\,\scriptsize{0.17}} 
    & \multicolumn{1}{c}{\normalsize{27.39}\,$\pm$\,\scriptsize{0.00}}
    & \multicolumn{1}{c}{\normalsize{33.99}\,$\pm$\,\scriptsize{6.44}} 
    & \multicolumn{1}{c}{\normalsize{12.14}\,$\pm$\,\scriptsize{2.36}} 
    & \multicolumn{1}{c}{\normalsize{48.80}\,$\pm$\,\scriptsize{2.95}}
    & \multicolumn{1}{c}{\normalsize{55.17}\,$\pm$\,\scriptsize{0.52}} 
    \\
    RAG (8B)
    & \multicolumn{1}{c}{\normalsize{75.51}\,$\pm$\,\scriptsize{0.01}} 
    & \multicolumn{1}{c}{\normalsize{22.53}\,$\pm$\,\scriptsize{0.15}} 
    & \multicolumn{1}{c}{\normalsize{51.07}\,$\pm$\,\scriptsize{0.00}}
    & \multicolumn{1}{c}{\normalsize{43.61}\,$\pm$\,\scriptsize{18.36}} 
    & \multicolumn{1}{c}{\normalsize{52.25}\,$\pm$\,\scriptsize{2.10}} 
    & \multicolumn{1}{c}{\normalsize{53.41}\,$\pm$\,\scriptsize{0.36}} 
    & \multicolumn{1}{c}{\normalsize{57.13}\,$\pm$\,\scriptsize{0.43}} 
    \\
    \midrule
    Skeleton $\left( \alpha \right)$ (3B)
    & \multicolumn{1}{c}{\normalsize{55.55}\,$\pm$\,\scriptsize{0.11}} 
    & \multicolumn{1}{c}{\normalsize{21.75}\,$\pm$\,\scriptsize{0.19}} 
    & \multicolumn{1}{c}{\normalsize{25.36}\,$\pm$\,\scriptsize{0.26}}
    & \multicolumn{1}{c}{\normalsize{37.90}\,$\pm$\,\scriptsize{0.99}} 
    & \multicolumn{1}{c}{\normalsize{9.37}\,$\pm$\,\scriptsize{0.11}} 
    & \multicolumn{1}{c}{\normalsize{52.09}\,$\pm$\,\scriptsize{5.51}} 
    & \multicolumn{1}{c}{\normalsize{55.44}\,$\pm$\,\scriptsize{0.31}}   
    \\
    Skeleton $\left( \alpha \right)$ (7B)
    & \multicolumn{1}{c}{\normalsize{46.57}\,$\pm$\,\scriptsize{0.52}} 
    & \multicolumn{1}{c}{\normalsize{20.15}\,$\pm$\,\scriptsize{0.03}} 
    & \multicolumn{1}{c}{\normalsize{28.10}\,$\pm$\,\scriptsize{1.36}}
    & \multicolumn{1}{c}{\normalsize{47.94}\,$\pm$\,\scriptsize{1.66}} 
    & \multicolumn{1}{c}{\normalsize{10.25}\,$\pm$\,\scriptsize{0.06}} 
    & \multicolumn{1}{c}{\normalsize{50.44}\,$\pm$\,\scriptsize{0.36}}
    & \multicolumn{1}{c}{\normalsize{56.13}\,$\pm$\,\scriptsize{0.66}}  
    \\
    Skeleton $\left( \alpha \right)$ (8B)
    & \multicolumn{1}{c}{\normalsize{73.40}\,$\pm$\,\scriptsize{0.05}} 
    & \multicolumn{1}{c}{\normalsize{24.58}\,$\pm$\,\scriptsize{0.29}} 
    & \multicolumn{1}{c}{\normalsize{45.54}\,$\pm$\,\scriptsize{3.35}}
    & \multicolumn{1}{c}{\normalsize{48.89}\,$\pm$\,\scriptsize{5.54}} 
    & \multicolumn{1}{c}{\normalsize{51.49}\,$\pm$\,\scriptsize{0.05}} 
    & \multicolumn{1}{c}{\normalsize{53.75}\,$\pm$\,\scriptsize{0.25}} 
    & \multicolumn{1}{c}{\normalsize{58.21}\,$\pm$\,\scriptsize{0.98}} 
    \\

    Skeleton $\left( \beta \right)$ (3B)
    & \multicolumn{1}{c}{\normalsize{55.55}\,$\pm$\,\scriptsize{0.00}} 
    & \multicolumn{1}{c}{\normalsize{22.09}\,$\pm$\,\scriptsize{0.66}} 
    & \multicolumn{1}{c}{\normalsize{25.07}\,$\pm$\,\scriptsize{0.39}}
    & \multicolumn{1}{c}{\normalsize{40.09}\,$\pm$\,\scriptsize{2.51}} 
    & \multicolumn{1}{c}{\normalsize{9.94}\,$\pm$\,\scriptsize{0.03}} 
    & \multicolumn{1}{c}{\normalsize{50.98}\,$\pm$\,\scriptsize{1.71}} 
    & \multicolumn{1}{c}{\normalsize{54.12}\,$\pm$\,\scriptsize{1.90}} 
    \\
    Skeleton $\left( \beta \right)$ (7B)
    & \multicolumn{1}{c}{\normalsize{46.90}\,$\pm$\,\scriptsize{0.85}} 
    & \multicolumn{1}{c}{\normalsize{20.49}\,$\pm$\,\scriptsize{0.00}} 
    & \multicolumn{1}{c}{\normalsize{27.81}\,$\pm$\,\scriptsize{0.17}}
    & \multicolumn{1}{c}{\normalsize{44.75}\,$\pm$\,\scriptsize{0.60}} 
    & \multicolumn{1}{c}{\normalsize{11.55}\,$\pm$\,\scriptsize{0.42}} 
    & \multicolumn{1}{c}{\normalsize{48.06}\,$\pm$\,\scriptsize{0.04}}
    & \multicolumn{1}{c}{\normalsize{55.96}\,$\pm$\,\scriptsize{0.11}} 
    \\
    Skeleton $\left( \beta \right)$ (8B)
    & \multicolumn{1}{c}{\normalsize{73.25}\,$\pm$\,\scriptsize{0.04}} 
    & \multicolumn{1}{c}{\normalsize{24.37}\,$\pm$\,\scriptsize{0.06}} 
    & \multicolumn{1}{c}{\normalsize{44.60}\,$\pm$\,\scriptsize{0.09}}
    & \multicolumn{1}{c}{\normalsize{48.07}\,$\pm$\,\scriptsize{9.95}} 
    & \multicolumn{1}{c}{\normalsize{50.95}\,$\pm$\,\scriptsize{3.41}} 
    & \multicolumn{1}{c}{\normalsize{55.99}\,$\pm$\,\scriptsize{0.02}} 
    & \multicolumn{1}{c}{\normalsize{57.62}\,$\pm$\,\scriptsize{0.10}} 
    \\

    Skeleton $\left( \gamma \right)$ (3B)
    & \multicolumn{1}{c}{\normalsize{55.61}\,$\pm$\,\scriptsize{0.01}} 
    & \multicolumn{1}{c}{\normalsize{22.26}\,$\pm$\,\scriptsize{0.39}} 
    & \multicolumn{1}{c}{\normalsize{25.65}\,$\pm$\,\scriptsize{0.34}}
    & \multicolumn{1}{c}{\normalsize{37.34}\,$\pm$\,\scriptsize{2.08}} 
    & \multicolumn{1}{c}{\normalsize{10.47}\,$\pm$\,\scriptsize{0.36}} 
    & \multicolumn{1}{c}{\normalsize{54.85}\,$\pm$\,\scriptsize{5.18}} 
    & \multicolumn{1}{c}{\normalsize{53.16}\,$\pm$\,\scriptsize{0.84}}  
    \\
    Skeleton $\left( \gamma \right)$ (7B)
    & \multicolumn{1}{c}{\normalsize{46.28}\,$\pm$\,\scriptsize{0.68}} 
    & \multicolumn{1}{c}{\normalsize{20.32}\,$\pm$\,\scriptsize{0.00}} 
    & \multicolumn{1}{c}{\normalsize{29.25}\,$\pm$\,\scriptsize{0.14}}
    & \multicolumn{1}{c}{\normalsize{46.62}\,$\pm$\,\scriptsize{1.30}} 
    & \multicolumn{1}{c}{\normalsize{10.50}\,$\pm$\,\scriptsize{0.00}} 
    & \multicolumn{1}{c}{\normalsize{49.41}\,$\pm$\,\scriptsize{0.14}}
    & \multicolumn{1}{c}{\normalsize{57.09}\,$\pm$\,\scriptsize{1.14}} 
    \\
    Skeleton $\left( \gamma \right)$ (8B)
    & \multicolumn{1}{c}{\normalsize{74.55}\,$\pm$\,\scriptsize{0.26}} 
    & \multicolumn{1}{c}{\normalsize{24.37}\,$\pm$\,\scriptsize{0.01}} 
    & \multicolumn{1}{c}{\normalsize{44.38}\,$\pm$\,\scriptsize{0.37}}
    & \multicolumn{1}{c}{\normalsize{49.52}\,$\pm$\,\scriptsize{5.58}} 
    & \multicolumn{1}{c}{\normalsize{52.94}\,$\pm$\,\scriptsize{0.30}} 
    & \multicolumn{1}{c}{\normalsize{56.46}\,$\pm$\,\scriptsize{0.10}} 
    & \multicolumn{1}{c}{\normalsize{56.81}\,$\pm$\,\scriptsize{0.02}} 
    \\
    A-CM (3B)
    & \multicolumn{1}{c}{\normalsize{55.26}\,$\pm$\,\scriptsize{0.41}} 
    & \multicolumn{1}{c}{\normalsize{20.74}\,$\pm$\,\scriptsize{0.65}} 
    & \multicolumn{1}{c}{\normalsize{25.17}\,$\pm$\,\scriptsize{0.53}} 
    & \multicolumn{1}{c}{\normalsize{47.81}\,$\pm$\,\scriptsize{0.71}} 
    & \multicolumn{1}{c}{\normalsize{7.65}\,$\pm$\,\scriptsize{2.99}} 
    & \multicolumn{1}{c}{\normalsize{57.19}\,$\pm$\,\scriptsize{0.58}} 
    & \multicolumn{1}{c}{\normalsize{55.23}\,$\pm$\,\scriptsize{1.09}}  
    \\
    A-CM (7B)
    & \multicolumn{1}{c}{\normalsize{45.16}\,$\pm$\,\scriptsize{0.07}} 
    & \multicolumn{1}{c}{\normalsize{14.45}\,$\pm$\,\scriptsize{0.29}} 
    & \multicolumn{1}{c}{\normalsize{24.30}\,$\pm$\,\scriptsize{1.36}}
    & \multicolumn{1}{c}{\normalsize{41.47}\,$\pm$\,\scriptsize{0.23}} 
    & \multicolumn{1}{c}{\normalsize{7.50}\,$\pm$\,\scriptsize{0.14}} 
    & \multicolumn{1}{c}{\normalsize{49.35}\,$\pm$\,\scriptsize{0.40}} 
    & \multicolumn{1}{c}{\normalsize{53.00}\,$\pm$\,\scriptsize{1.10}} 
    \\
    A-CM (8B)
    & \multicolumn{1}{c}{\normalsize{72.12}\,$\pm$\,\scriptsize{1.75}} 
    & \multicolumn{1}{c}{\normalsize{19.48}\,$\pm$\,\scriptsize{0.35}} 
    & \multicolumn{1}{c}{\normalsize{51.30}\,$\pm$\,\scriptsize{0.17}}
    & \multicolumn{1}{c}{\normalsize{49.91}\,$\pm$\,\scriptsize{1.90}} 
    & \multicolumn{1}{c}{\normalsize{9.94}\,$\pm$\,\scriptsize{0.51}} 
    & \multicolumn{1}{c}{\normalsize{51.99}\,$\pm$\,\scriptsize{0.55}} 
    & \multicolumn{1}{c}{\normalsize{57.86}\,$\pm$\,\scriptsize{0.83}} 
    \\
    FGC (3B)
    & \multicolumn{1}{c}{\normalsize{55.26}\,$\pm$\,\scriptsize{0.41}} 
    & \multicolumn{1}{c}{\normalsize{22.18}\,$\pm$\,\scriptsize{0.04}} 
    & \multicolumn{1}{c}{\normalsize{27.52}\,$\pm$\,\scriptsize{0.15}}
    & \multicolumn{1}{c}{\normalsize{47.80}\,$\pm$\,\scriptsize{1.09}} 
    & \multicolumn{1}{c}{\normalsize{6.93}\,$\pm$\,\scriptsize{0.22}} 
    & \multicolumn{1}{c}{\normalsize{47.57}\,$\pm$\,\scriptsize{0.22}} 
    & \multicolumn{1}{c}{\normalsize{53.20}\,$\pm$\,\scriptsize{0.17}} 
    \\
    FGC (7B)
    & \multicolumn{1}{c}{\normalsize{47.30}\,$\pm$\,\scriptsize{0.55}} 
    & \multicolumn{1}{c}{\normalsize{15.30}\,$\pm$\,\scriptsize{0.29}} 
    & \multicolumn{1}{c}{\normalsize{24.36}\,$\pm$\,\scriptsize{0.66}}
    & \multicolumn{1}{c}{\normalsize{47.17}\,$\pm$\,\scriptsize{0.39}} 
    & \multicolumn{1}{c}{\normalsize{9.04}\,$\pm$\,\scriptsize{1.25}} 
    & \multicolumn{1}{c}{\normalsize{49.44}\,$\pm$\,\scriptsize{1.98}} 
    & \multicolumn{1}{c}{\normalsize{54.09}\,$\pm$\,\scriptsize{0.09}}   
    \\
    FGC (8B)
    & \multicolumn{1}{c}{\normalsize{68.06}\,$\pm$\,\scriptsize{0.39}} 
    & \multicolumn{1}{c}{\normalsize{19.39}\,$\pm$\,\scriptsize{0.12}} 
    & \multicolumn{1}{c}{\normalsize{52.02}\,$\pm$\,\scriptsize{0.04}}
    & \multicolumn{1}{c}{\normalsize{35.97}\,$\pm$\,\scriptsize{1.28}} 
    & \multicolumn{1}{c}{\normalsize{7.96}\,$\pm$\,\scriptsize{2.09}} 
    & \multicolumn{1}{c}{\normalsize{54.23}\,$\pm$\,\scriptsize{0.10}} 
    & \multicolumn{1}{c}{\normalsize{58.05}\,$\pm$\,\scriptsize{0.57}}  
    \\
    UGC (3B)
    & \multicolumn{1}{c}{\normalsize{55.90}\,$\pm$\,\scriptsize{0.03}} 
    & \multicolumn{1}{c}{\normalsize{21.64}\,$\pm$\,\scriptsize{0.48}} 
    & \multicolumn{1}{c}{\normalsize{31.41}\,$\pm$\,\scriptsize{1.51}} 
    & \multicolumn{1}{c}{\normalsize{35.98}\,$\pm$\,\scriptsize{3.19}} 
    & \multicolumn{1}{c}{\normalsize{9.40}\,$\pm$\,\scriptsize{2.53}} 
    & \multicolumn{1}{c}{\normalsize{52.63}\,$\pm$\,\scriptsize{0.00}} 
    & \multicolumn{1}{c}{\normalsize{54.95}\,$\pm$\,\scriptsize{2.81}} 
    \\
    UGC (7B)
    & \multicolumn{1}{c}{\normalsize{46.33}\,$\pm$\,\scriptsize{0.52}} 
    & \multicolumn{1}{c}{\normalsize{18.44}\,$\pm$\,\scriptsize{0.03}} 
    & \multicolumn{1}{c}{\normalsize{28.53}\,$\pm$\,\scriptsize{2.43}} 
    & \multicolumn{1}{c}{\normalsize{41.62}\,$\pm$\,\scriptsize{1.91}} 
    & \multicolumn{1}{c}{\normalsize{8.70}\,$\pm$\,\scriptsize{1.90}} 
    & \multicolumn{1}{c}{\normalsize{48.50}\,$\pm$\,\scriptsize{2.12}} 
    & \multicolumn{1}{c}{\normalsize{49.23}\,$\pm$\,\scriptsize{2.99}} 
    \\
    UGC (8B)
    & \multicolumn{1}{c}{\normalsize{75.14}\,$\pm$\,\scriptsize{0.41}} 
    & \multicolumn{1}{c}{\normalsize{22.51}\,$\pm$\,\scriptsize{0.37}} 
    & \multicolumn{1}{c}{\normalsize{43.30}\,$\pm$\,\scriptsize{0.70}}
    & \multicolumn{1}{c}{\normalsize{48.32}\,$\pm$\,\scriptsize{2.05}} 
    & \multicolumn{1}{c}{\normalsize{12.39}\,$\pm$\,\scriptsize{1.14}} 
    & \multicolumn{1}{c}{\normalsize{41.37}\,$\pm$\,\scriptsize{0.73}} 
    & \multicolumn{1}{c}{\normalsize{54.09}\,$\pm$\,\scriptsize{1.96}}   
    \\
    \midrule
    $\text{H}\text{S}_2\text{C}$ (Ours) (3B)
    & \multicolumn{1}{c}{\normalsize{66.67}\,$\pm$\,\scriptsize{0.02}} 
    & \multicolumn{1}{c}{\normalsize{27.52}\,$\pm$\,\scriptsize{0.21}} 
    & \multicolumn{1}{c}{\normalsize{49.12}\,$\pm$\,\scriptsize{0.08}}
    & \multicolumn{1}{c}{\normalsize{48.52}\,$\pm$\,\scriptsize{0.30}}
    & \multicolumn{1}{c}{\normalsize{10.90}\,$\pm$\,\scriptsize{0.14}}
    & \multicolumn{1}{c}{\normalsize{\textbf{57.50}}\,$\pm$\,\scriptsize{\textbf{0.14}}} 
    & \multicolumn{1}{c}{\normalsize{56.54}\,$\pm$\,\scriptsize{0.47}}    
    \\
    $\text{H}\text{S}_2\text{C}$ (Ours) (7B)
    & \multicolumn{1}{c}{\normalsize{55.79}\,$\pm$\,\scriptsize{0.01}} 
    & \multicolumn{1}{c}{\normalsize{24.03}\,$\pm$\,\scriptsize{0.00}} 
    & \multicolumn{1}{c}{\normalsize{30.34}\,$\pm$\,\scriptsize{0.10}}
    & \multicolumn{1}{c}{\normalsize{49.32}\,$\pm$\,\scriptsize{2.35}} 
    & \multicolumn{1}{c}{\normalsize{15.20}\,$\pm$\,\scriptsize{0.05}} 
    & \multicolumn{1}{c}{\normalsize{50.76}\,$\pm$\,\scriptsize{0.33}}
    & \multicolumn{1}{c}{\normalsize{56.29}\,$\pm$\,\scriptsize{0.00}} 
    \\
    $\text{H}\text{S}_2\text{C}$ (Ours) (8B)
    & \multicolumn{1}{c}{\normalsize{\textbf{79.77}}\,$\pm$\,\scriptsize{\textbf{0.01}}} 
    & \multicolumn{1}{c}{\normalsize{\textbf{30.12}}\,$\pm$\,\scriptsize{\textbf{0.20}}} 
    & \multicolumn{1}{c}{\normalsize{\textbf{53.03}}\,$\pm$\,\scriptsize{\textbf{0.00}}}
    & \multicolumn{1}{c}{\normalsize{\textbf{50.90}}\,$\pm$\,\scriptsize{\textbf{0.52}}} 
    & \multicolumn{1}{c}{\normalsize{\textbf{53.40}}\,$\pm$\,\scriptsize{\textbf{0.09}}} 
    & \multicolumn{1}{c}{\normalsize{57.00}\,$\pm$\,\scriptsize{1.60}} 
    & \multicolumn{1}{c}{\normalsize{\textbf{58.53}}\,$\pm$\,\scriptsize{\textbf{0.00}}}  
    \\
    \bottomrule
    \end{tabular}
  }
\vspace{-3pt}
\end{table*}

\section{Conclusion}\label{section6}
In this work, we propose $\text{H}\text{S}_2\text{C}$, a homophily-aware framework for community-level TAG compression, which jointly exploits structural and semantic homophily to preserve essential topology and contextual information for LLM downstream reasoning. By performing an entropy‑guided hierarchical partition from a global perspective, the framework uncovers homophilic structures and effectively facilitates semantic discovery. Extensive experiments demonstrate that $\text{H}\text{S}_2\text{C}$ achieves high compression efficiency with improved accuracy, and exhibits strong scalability and applicability, significantly enhancing the reasoning consistency of LLMs. 

In future work, we will investigate the scalability of $\text{H}\text{S}_2\text{C}$ to industrial-scale TAGs by designing more efficient compression strategies.  Additionally, we will extend our framework to the temporal graphs, where dynamically evolving structures and semantics necessitate responsive homophilic community detection and temporally-aware semantic aggregation to ensure stable and consistent LLM reasoning over time.

\newpage

\newpage

\bibliographystyle{unsrt}
\bibliography{references}

@inproceedings{rw_wang_ijcal2024nlp_context_survey,
author = "Wang, Xindi and Salmani, Mahsa and Omidi, Parsa and Ren, Xiangyu and Rezagholizadeh, Mehdi and Eshaghi, Armaghan",
title = "Beyond the limits: a survey of techniques to extend the context length in large language models",
year = "2024",
booktitle = "Proceedings of the Thirty-Third International Joint Conference on Artificial Intelligence, {IJCAI-24}",
pages= "8299--8307",
}

@inproceedings{rw_tang_sigir2024graphgpt,
author = "Tang, Jiabin and Yang, Yuhao and Wei, Wei and Shi, Lei and Su, Lixin and Cheng, Suqi and Yin, Dawei and Huang, Chao",
title = "GraphGPT: Graph Instruction Tuning for Large Language Models",
booktitle = "Proceedings of the 47th International ACM SIGIR Conference on Research and Development in Information Retrieval, {SIGIR-24}",
pages = "491--500",
year = "2024"
}

@inproceedings{rw_chen_icml2024llaga,
author = "Runjin Chen and Tong Zhao and AJAY KUMAR JAISWAL and Neil Shah and Zhangyang Wang",
title = "LLaGA: Large Language and Graph Assistant",
year = "2024",
booktitle = "Forty-first International Conference on Machine Learning, {ICML-24}",
pages = "7809--7823",
}

@inproceedings{rw_ye_EACL2024instructglm,
author = "Ye, Ruosong and Zhang, Caiqi and Wang, Runhui and Xu, Shuyuan and Zhang, Yongfeng",
title = "Language is All a Graph Needs",
year = "2024",
booktitle = "Findings of the Association for Computational Linguistics, {EACL-24}",
pages = "1955--1973"
}

@article{Chen_GraphLLM2024KDD,
author = "Chen, Zhikai and Mao, Haitao and Li, Hang and Jin, Wei and Wen, Hongzhi and Wei, Xiaochi and Wang, Shuaiqiang and Yin, Dawei and Fan, Wenqi and Liu, Hui and Tang, Jiliang",
title = "Exploring the Potential of Large Language Models (LLMs)in Learning on Graphs",
year = "2024",
issue_date = "December 2023",
volume = "25",
number = "2",
journal = "ACM SIGKDD Explorations Newsletter",
pages = "42–61",
}

@inproceedings{fatemi_talk2024ICLR,
title= "Talk like a Graph: Encoding Graphs for Large Language Models",
author= "Bahare Fatemi and Jonathan Halcrow and Bryan Perozzi",
booktitle= "The Twelfth International Conference on Learning Representations, {ICLR-24}",
year= "2024",
}

@inproceedings{rw_wu_icml2025when,
title= "When Do {LLM}s Help With Node Classification? A Comprehensive Analysis",
author= "Xixi Wu and Yifei Shen and Fangzhou Ge and Caihua Shan and Yizhu Jiao and Xiangguo Sun and Hong Cheng",
booktitle= "Forty-second International Conference on Machine Learning, {ICML-25}",
year= "2025",
}

@inproceedings{related_work_gc_1,
title= "Graph Condensation for Graph Neural Networks",
author= "Wei Jin and Lingxiao Zhao and Shichang Zhang and Yozen Liu and Jiliang Tang and Neil Shah",
booktitle= "International Conference on Learning Representations, {ICLR-22}",
year= "2022",
}

@inproceedings{related_work_gc_2,
author = "Jin, Wei and Tang, Xianfeng and Jiang, Haoming and Li, Zheng and Zhang, Danqing and Tang, Jiliang and Yin, Bing",
title = "Condensing Graphs via One-Step Gradient Matching",
year = "2022",
booktitle = "Proceedings of the 28th ACM SIGKDD Conference on Knowledge Discovery and Data Mining, {KDD-22}",
pages = "720–730",
numpages = "11",
}

@article{se_li_tit2016,
title= "Structural information and dynamical complexity of networks",
author= "Li, Angsheng and Pan, Yicheng",
journal= "IEEE Transactions on Information Theory",
volume= "62",
number= "6",
pages= "3290--3339",
year= "2016",
}

@article{OOD_SEIB,
author = "Di, Zijun and Zheng, Peng and Lu, Bin and Guan, Kai and Fu, Luoyi and Jin, Ningdi and Chen, Ye and Gan, Xiaoying and Zhou, Lei and Wang, Xinbing and Zhou, Chenghu", 
title = "Graph Out-of-Distribution Generalization Based on Structural-Entropy-Guided Information Bottleneck", 
year = "2025", 
volume = "20", 
number = "1", 
pages = "1-34", 
journal = "ACM Transactions on Knowledge Discovery from Data", 
}

@article{se_nips_liu2019com_deception,
title= "REM: From structural entropy to community structure deception",
author= "Liu, Yiwei and Liu, Jiamou and Zhang, Zijian and Zhu, Liehuang and Li, Angsheng",
journal= "Proceedings of the 33rd International Conference on Neural Information Processing Systems, {NeurIPS-19}",
pages = "12938-12948",
year= "2019",
}

@article{se_yang2024incremental,
title= "Incremental measurement of structural entropy for dynamic graphs",
author= "Yang, Runze and Peng, Hao and Liu, Chunyang and Li, Angsheng",
journal= "Artificial Intelligence",
volume= "334",
pages= "104175",
year= "2024",
}

@inproceedings{se_icml_wu2022poling,
title = "Structural entropy guided graph hierarchical pooling",
author = "Wu, Junran and Chen, Xueyuan and Xu, Ke and Li, Shangzhe",
booktitle = "Proceedings of the 39th International Conference on Machine Learning, {ICML-22}",
pages = "24017-24030",
year = "2022",
}

@inproceedings{se_icml_wu2023sega,
title= "Sega: Structural entropy guided anchor view for graph contrastive learning",
author= "Wu, Junran and Chen, Xueyuan and Shi, Bowen and Li, Shangzhe and Xu, Ke",
booktitle= "Proceedings of the 40th International Conference on Machine Learning, {ICML-23}",
pages= "37293--37312",
year= "2023",
}

@inproceedings{se_www_zou2023segsl,
title= "Se-gsl: A general and effective graph structure learning framework through structural entropy optimization",
author= "Zou, Dongcheng and Peng, Hao and Huang, Xiang and Yang, Renyu and Li, Jianxin and Wu, Jia and Liu, Chunyang and Yu, Philip S",
booktitle= "Proceedings of the ACM Web Conference 2023, {WWW-23}",
pages= "499-510",
year= "2023",
}

@inproceedings{se_aaai_hou2025ood,
title= "Structural Entropy Guided Unsupervised Graph Out-Of-Distribution Detection",
author= "Hou, Yue and Zhu, He and Liu, Ruomei and Su, Yingke and Xia, Jinxiang and Wu, Junran and Xu, Ke",
booktitle= "Proceedings of the AAAI Conference on Artificial Intelligence, {AAAI-25}",
pages= "17258-17266",
year= "2025",
}

@inproceedings{ijcai2022p497,
title     = "A Simple yet Effective Method for Graph Classification",
author    = "Wu, Junran and Li, Shangzhe and Li, Jianhao and Pan, Yicheng and Xu, Ke",
booktitle = "Proceedings of the Thirty-First International Joint Conference on
           Artificial Intelligence, {IJCAI-22}",
pages     = "3580--3586",
year      = "2022",
}

@ARTICLE{10735397,
author= "Zeng, Guangjie and Peng, Hao and Li, Angsheng and Wu, Jia and Liu, Chunyang and Yu, Philip S.",
journal= "IEEE Transactions on Knowledge and Data Engineering", 
title= "Scalable Semi-Supervised Clustering via Structural Entropy With Different Constraints", 
year= "2025",
volume= "37",
number= "1",
pages= "478-492",
}

@article{LI2017211,
title = "Resistance maximization principle for defending networks against virus attack",
author = "Angsheng Li and Xiaohui Zhang and Yicheng Pan",
journal = "Physica A: Statistical Mechanics and its Applications",
volume = "466",
pages = "211-223",
year = "2017",
}

@ARTICLE{9465220,
author= "Tian, Youliang and Zhang, Zhiying and Xiong, Jinbo and Chen, Lei and others",
journal= "IEEE Internet of Things Journal", 
title= "Achieving Graph Clustering Privacy Preservation Based on Structure Entropy in Social IoT", 
year= "2022",
volume= "9",
number= "4",
pages= "2761-2777",
}

@article{10.1145/3660522,
author = "Peng, Hao and Zhang, Jingyun and Huang, Xiang and Hao, Zhifeng and Li, Angsheng and Yu, Zhengtao and Yu, Philip S.",
title = "Unsupervised Social Bot Detection via Structural Information Theory",
year = "2024",
volume = "42",
number = "6",
pages = "1-42",
journal = "ACM Transactions on Information Systems",
}

@article{Cao_Peng_Yu_Yu_2024, 
title= "Hierarchical and Incremental Structural Entropy Minimization for Unsupervised Social Event Detection", 
journal= "Proceedings of the AAAI Conference on Artificial Intelligence, {AAAI-24}", 
author= "Cao, Yuwei and Peng, Hao and Yu, Zhengtao and Yu, Philip S.", 
year= "2024", 
pages= "8255-8264",
}

@inproceedings{10.1145/3637528.3671871,
author = "Yang, Yingguang and Wu, Qi and He, Buyun and Peng, Hao and Yang, Renyu and Hao, Zhifeng and Liao, Yong",
title = "SEBot: Structural Entropy Guided Multi-View Contrastive learning for Social Bot Detection",
year = "2024",
booktitle = "Proceedings of the 30th ACM SIGKDD Conference on Knowledge Discovery and Data Mining, {KDD-24}",
pages = "3841–3852",
}

@inproceedings{baseline_gcn_kipf2016gcn,
title= "Semi-Supervised Classification with Graph Convolutional Networks",
author= "Thomas N. Kipf and Max Welling",
booktitle= "International Conference on Learning Representations, {ICLR-17}",
year= "2017",
}

@inproceedings{baseline_gat_velivckovic2017gat,
title= "Graph Attention Networks",
author= "Petar Veličković and Guillem Cucurull and Arantxa Casanova and Adriana Romero and Pietro Liò and Yoshua Bengio",
booktitle= "International Conference on Learning Representations, {ICLR-18}",
year= "2018",
}

@inproceedings{baseline_gin_xu2018gnn,
title= "How Powerful are Graph Neural Networks?",
author= "Keyulu Xu and Weihua Hu and Jure Leskovec and Stefanie Jegelka",
booktitle= "International Conference on Learning Representations, {ICLR-19}",
year= "2019",
}

@article{baseline_graphsage_hamilton2017graphsage,
  title= "Inductive representation learning on large graphs", 
  author= "Hamilton, Will and Ying, Zhitao and Leskovec, Jure", 
  journal= "Proceedings of the 31st International Conference on Neural Information Processing Systems, {NeurIPS-17}", 
  volume= "30", 
  year= "2017"
}

@inproceedings{baseline_degree_ali2024degree,
title= "Degree-based stratification of nodes in Graph Neural Networks",
author= "Ali, Ameen and Wolf, Lior and Cevikalp, Hakan",
booktitle= "Proceedings of the 15th Asian Conference on Machine Learning",
pages= "15--27",
year= "2024",
volume = "222",
organization= "PMLR",
}

@inproceedings{baseline_degree_Zhang2023degree_importance,
author = "Zhang, Fan and Linghu, Qingyuan and Xie, Jiadong and Wang, Kai and Lin, Xuemin and Zhang, Wenjie",
title = "Quantifying Node Importance over Network Structural Stability",
year = "2023",
booktitle = "Proceedings of the 29th ACM SIGKDD Conference on Knowledge Discovery and Data Mining, {KDD-23}",
pages = "3217–3228",
}

@article{baseline_rag_li2025large,
title= "Are Large Language Models In-Context Graph Learners?",
author= "Li, Jintang and others",
journal= "arXiv preprint arXiv:2502.13562",
year= "2025",
}

@inproceedings{baseline_skeleton_cao2024graph,
title= "Graph-Skeleton:\~{} 1\% Nodes are Sufficient to Represent Billion-Scale Graph",
author= "Cao, Linfeng and Deng, Haoran and Yang, Yang and Wang, Chunping and Chen, Lei",
booktitle= "Proceedings of the ACM Web Conference 2024, {WWW-24}",
pages= "570--581",
year= "2024",
}

@inproceedings{devlin2019bert,
title= "Bert: Pre-training of deep bidirectional transformers for language understanding",
author= "Devlin, Jacob and Chang, Ming-Wei and others",
booktitle= "Proceedings of the 2019 conference of the North American chapter of the association for computational linguistics: human language technologies, volume 1 (long and short papers)",
pages= "4171--4186",
year= "2019",
}

@inproceedings{ogb_arxiv,
author = "Hu, Weihua and Fey, Matthias and Zitnik, Marinka and Dong, Yuxiao and Ren, Hongyu and Liu, Bowen and Catasta, Michele and Leskovec, Jure",
title = "Open graph benchmark: datasets for machine learning on graphs",
year = "2020",
pages = "22118-22133",
booktitle = "Proceedings of the 34th International Conference on Neural Information Processing Systems, {NeurIPS-20}",
}

@inproceedings{tape,
title= "Harnessing Explanations: {LLM}-to-{LM} Interpreter for Enhanced Text-Attributed Graph Representation Learning",
author= "Xiaoxin He and Xavier Bresson and Thomas Laurent and Adam Perold and Yann LeCun and Bryan Hooi",
booktitle= "The Twelfth International Conference on Learning Representations, {ICLR-24}",
year= "2024",
}

@inproceedings{instagram,
author = "Huang, Xuanwen and Han, Kaiqiao and Yang, Yang and Bao, Dezheng and Tao, Quanjin and Chai, Ziwei and Zhu, Qi",
title = "Can GNN be Good Adapter for LLMs?",
year = "2024",
booktitle = "Proceedings of the ACM Web Conference 2024, {WWW-24}",
pages = "893–904",
}

@inproceedings{citeseer,
author = "Yang, Zhilin and Cohen, William W. and Salakhutdinov, Ruslan",
title = "Revisiting semi-supervised learning with graph embeddings",
year = "2016",
booktitle = "Proceedings of the 33rd International Conference on International Conference on Machine Learning, {ICML-16}",
pages = "40–48",
}

@misc{taglas,
  title= "TAGLAS: An atlas of text-attributed graph datasets in the era of large graph and language models", 
  author= "Jiarui Feng and Hao Liu and Lecheng Kong and Yixin Chen and Muhan Zhang",
  year= "2024",
  eprint= "2406.14683",
  archivePrefix= "arXiv",
}

@article{grattafiori2024llama,
title= "The llama 3 herd of models",
author= "Grattafiori, Aaron and Dubey, Abhimanyu and others",
journal= "arXiv preprint arXiv:2407.21783",
year= "2024",
}

@article{homophily_assump,
title= "Valence-based homophily on Twitter: Network analysis of emotions and political talk in the 2012 presidential election",
author= "Himelboim, Itai and Sweetser, Kaye D and Tinkham, Spencer F and Cameron, Kristen and Danelo, Matthew and West, Kate",
journal= "New media \& society",
volume= "18",
number= "7",
pages= "1382--1400",
year= "2016",
}

@inproceedings{chaohuang_survey,
author = "Ren, Xubin and Tang, Jiabin and Yin, Dawei and Chawla, Nitesh and Huang, Chao",
title = "A Survey of Large Language Models for Graphs",
year = "2024",
booktitle = "Proceedings of the 30th ACM SIGKDD Conference on Knowledge Discovery and Data Mining, {KDD-24}",
pages = "6616–6626",
}

@ARTICLE{jiaweihan_survey,
author= "Jin, Bowen and Liu, Gang and Han, Chi and Jiang, Meng and Ji, Heng and Han, Jiawei",
journal= "IEEE Transactions on Knowledge and Data Engineering", 
title= "Large Language Models on Graphs: A Comprehensive Survey", 
year= "2024",
volume= "36",
number= "12",
pages= "8622-8642",
}

@inproceedings{dblp_dataset,
  title="Oag-bench: a human-curated benchmark for academic graph mining",
  author="Zhang, Fanjin and Shi, Shijie and others",
  booktitle="Proceedings of the 30th ACM SIGKDD Conference on Knowledge Discovery and Data Mining, {KDD-24}",
  pages="6214--6225",
  year="2024"
}

@inproceedings{book_dataset,
author = "Wan, Mengting and McAuley, Julian",
title = "Item recommendation on monotonic behavior chains",
year = "2018",
booktitle = "Proceedings of the 12th ACM Conference on Recommender Systems, {RecSys-18}",
pages = "86–94",
}

@inproceedings{htag_dataset,
  title="Multi-Scale Heterogeneous Text-Attributed Graph Datasets From Diverse Domains",
  author="Liu, Yunhui and Xie, Qizhuo and Shi, Jinwei and others",
  booktitle="Companion Proceedings of the ACM on Web Conference 2025, {WWW-25}",
  pages="757--760",
  year="2025"
}

@inproceedings{aconvmatch,
  title = "Graph coarsening via convolution matching for scalable graph neural network training", 
  author = "Dickens, Charles and Huang, Edward and Reganti, Aishwarya and Zhu, Jiong and Subbian, Karthik and Koutra, Danai", 
  booktitle = "Companion Proceedings of the ACM Web Conference 2024, {WWW-24}", 
  pages = "1502--1510", 
  year = "2024"
}

@InProceedings{fgc,
  title = "Featured Graph Coarsening with Similarity Guarantees", 
  author = "Kumar, Manoj and Sharma, Anurag and Saxena, Shashwat and Kumar, Sandeep", 
  booktitle = "Proceedings of the 40th International Conference on Machine Learning, {ICML-23}", 
  pages = "17953--17975", 
  year = "2023", 
}

@inproceedings{ugc,
author = "Kataria, Mohit and Kumar, Sandeep and Jayadeva", 
title = "UGC: universal graph coarsening", 
year = "2024", 
pages = "63057--63081",
booktitle = "Proceedings of the 38th International Conference on Neural Information Processing Systems, {NeurIPS-24}", 
}

@article{sst_dataset,
author = "Yuan, Hao and Yu, Haiyang and Gui, Shurui and Ji, Shuiwang", 
title = "Explainability in Graph Neural Networks: A Taxonomic Survey", 
year = "2023", 
volume = "45", 
number = "5", 
journal = "IEEE Transactions on Pattern Analysis and Machine Intelligence", 
pages = "5782–5799", 
}

@article{molecule_dataset,
  title= "MoleculeNet: a benchmark for molecular machine learning", 
  author= "Wu, Zhenqin and Ramsundar, Bharath and Feinberg, Evan N and Gomes, Joseph and Geniesse, Caleb and Pappu, Aneesh S and Leswing, Karl and Pande, Vijay", 
  journal= "Chemical science", 
  volume= "9", 
  number= "2", 
  pages= "513--530", 
  year= "2018", 
}

@inproceedings{vicuna_model,
author = "Zheng, Lianmin and Chiang, Wei-Lin and others",
title = "Judging LLM-as-a-judge with MT-bench and Chatbot Arena",
year = "2023",
booktitle = "Proceedings of the 37th International Conference on Neural Information Processing Systems, {NeurIPS-23}",
}

@inproceedings{penghao_ijcai_survey,
  title     = "A Survey of Structural Entropy: Theory, Methods, and Applications", 
  author    = "Su, Dingli and Peng, Hao and Pan, Yicheng and Li, Angsheng", 
  booktitle = "Proceedings of the Thirty-Fourth International Joint Conference on
               Artificial Intelligence, {IJCAI-25}", 
  pages     = "10660--10668", 
  year      = "2025", 
  note      = "Survey Track", 
}

\newpage

\newpage

\section{Appendix}

\subsection{Notations and Corresponding Meanings} \label{app: Notation}
In this section, we supplement the paper with a detailed list of notations and their corresponding meanings, as listed in Table~\ref{tab:notation}.

\begin{table}[h]
\centering
\caption{Notations and corresponding meanings.}
\label{tab:notation}
\resizebox{0.45\columnwidth}{!}{
\begin{tabular}{cl}
\toprule
\textbf{Notation} & \textbf{Meaning} \\
\midrule
$\mathcal{G}$ & Text-attributed graph (TAG) \\[1pt]
$\mathcal{V}$ & Node set of TAG \\[1pt]
$\mathcal{V}_{\text{tg}}$ & Target node of $\mathcal{V}$ \\[1pt]
$\mathcal{V}_{\text{bg}}$ & Background node of $\mathcal{V}$ \\[1pt]
$v_i$ & Node of $\mathcal{V}$ \\[1pt]
$\mathcal{E}$ & Edge set of TAG \\[1pt]
$\mathcal{R}$ & Raw text set of node text attribute \\[1pt]
$R_i$ & Raw text sequence of $\mathcal{R}$ \\[1pt]
$\mathcal{C}$ & Community set \\[1pt]
$c_i$ & Community of $\mathcal{C}$ \\[1pt]
$T_i$ & Vertex set of community $c_i$ \\[1pt]
$\mathbf{S}$ & Similarity matrix \\[1pt]
$\mathbf{x}_i$ & Feature vector of node $v_i$ \\[1pt]
$\mathcal{N}_k(\cdot)$ & Top-$k$ nearest neighbor \\[1pt]
$\mathcal{E}_k$ & Auxiliary edge set \\[1pt]
$\mathcal{H}^{(1)}(\cdot)$ & One-dimensional structural entropy \\[1pt]
$\hat{\mathcal{G}}$ & KNN-enhanced graph \\[1pt]
$\mathcal{T}$ & Hierarchical coding tree \\[1pt]
$\mathcal{C}_\text{homo}$ & Partitioned Homophilic community set \\[1pt]
$\widetilde{R}_b^{(i)}$ & Aggregated background text attribute of $c_i$ \\[1pt]
$\widetilde{v}_b^{(i)}$ & Aggregated background node of $c_i$ \\[1pt]
$\widetilde{\mathcal{G}}$ & Reconstructed compressed graph \\[1pt]
$\widetilde{\mathcal{V}}$ & Reconstructed node set of $\widetilde{\mathcal{G}}$ \\[1pt]
$\widetilde{\mathcal{E}}$ & Reconstructed edge set of $\widetilde{\mathcal{G}}$ \\[1pt]
$\widetilde{\mathcal{R}}$ & Reconstructed raw text set of $\widetilde{\mathcal{G}}$ \\[1pt]
\bottomrule
\end{tabular}
}
\end{table}

\section{Supplementary Algorithm Details} \label{app: Supplementary Algorithm Details}
In this section, we provide a detailed illustration of the entire coding tree construction procedure. We also include the accompanying algorithm pseudocode and present an analysis of the corresponding time complexity, as shown below.

\subsection{Structural Entropy}\label{subapp: Structural Entropy}
Structural Entropy (SE)~\citep{se_li_tit2016} is a quantitative measure of the amount of structural information embedded within a complex graph system~\citep{OOD_SEIB, penghao_ijcai_survey, se_aaai_hou2025ood, se_yang2024incremental, se_nips_liu2019com_deception, se_www_zou2023segsl}. It reveals the intrinsic hierarchical organization of a graph by decoding the original graph $\mathcal{G}$ into a coding tree $\mathcal{T}$ through a hierarchical abstraction strategy, where nodes are partitioned into communities at various levels of granularity. In recent years, structural entropy has been widely applied in diverse domains such as bioinformatics~\citep{se_icml_wu2023sega, ijcai2022p497, se_icml_wu2022poling,10735397}, information security~\citep{LI2017211, 9465220}, and social network analysis~\citep{se_icml_wu2022poling, 10.1145/3660522, Cao_Peng_Yu_Yu_2024}. 
Specifically, given an undirected graph $\mathcal{G} = (\mathcal{V}, \mathcal{E})$, the SE of $\mathcal{G}$ is computed by performing a biased random walk to estimate node visit probabilities and summing the entropy contributions of all nodes accordingly:
\begin{equation}\label{supEq: Structural Entropy}
    \mathcal{H}^{\mathcal{T}}(\mathcal{G})=\!\!\!\!\!\sum_{\alpha\in\mathcal{T},\alpha\neq\lambda}\!\!\!\!\mathcal{H}^{\mathcal{T}}(\mathcal{G};\alpha)=-\!\!\!\!\!\sum_{\alpha\in\mathcal{T},\alpha\neq\lambda}\!\!\!\frac{g_\alpha}{\operatorname{vol}(\mathcal{G})}\log_2 \frac{\operatorname{vol}\left(\alpha\right)}{\operatorname{vol}\left(\alpha^-\right)},
\end{equation}
where $\alpha$ denotes a non-root node in the coding tree $\mathcal{T}$, whose associated vertex set is $T_\alpha \subset \mathcal{V}$. The term $g_\alpha$ represents the number of edges connecting nodes in and outside $T_\alpha$. The $\alpha^-$ refers to the immediate predecessor of $\alpha$. The volume terms $\operatorname{vol}(\alpha)$, $\operatorname{vol}(\alpha^-)$, and $\operatorname{vol}(\mathcal{G})$ represent the sum of node degrees within $\alpha$, $\alpha^-$, and $\mathcal{G}$, respectively. Minimizing $\mathcal{H}^{\mathcal{T}}(\mathcal{G})$ yields an optimal coding tree $\mathcal{T}^*$ that achieves $\mathcal{H}(\mathcal{G})=\min_{\forall \mathcal{T}}\{\mathcal{H}^\mathcal{T}(\mathcal{G})\}$, which encodes the graph in a way that minimizes the uncertainty of node codewords encountered during random walks. In this sense, the process of measuring the SE not only quantifies the information embedded in the graph but also implicitly reveals its underlying structural organization. It effectively separates regular structural patterns from randomness or noise in the topology. 

\begin{figure}[htbp]
    \centering
    \includegraphics[width=0.6\linewidth]{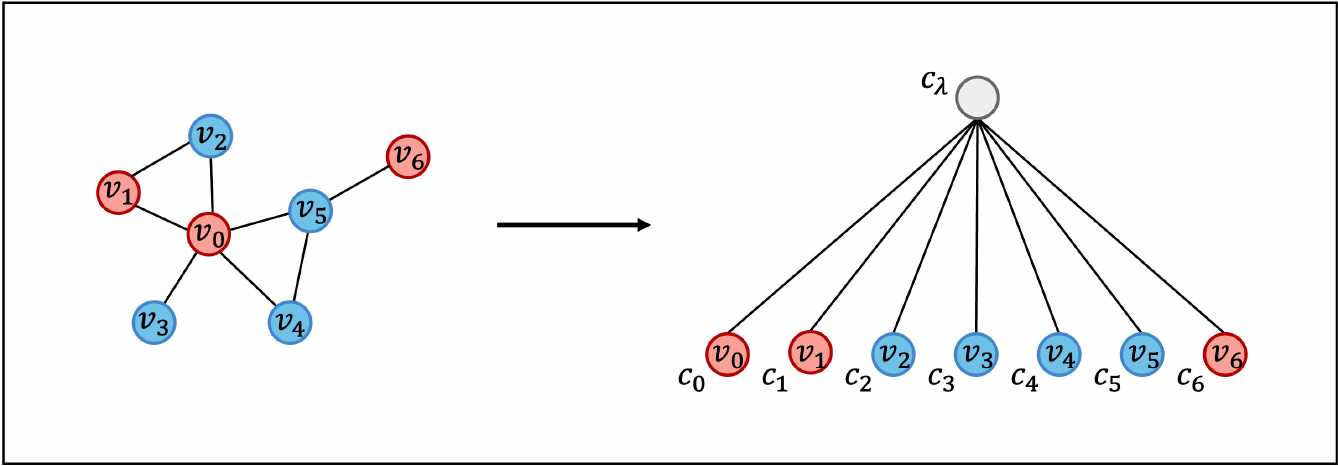}
    \caption{The initial coding tree $\mathcal{T}^{(1)}$ constructed from the graph $\mathcal{G}$. Each node $v_i \in \mathcal{V}$ is treated as an individual community $c_i$, and all such communities are organized as leaf nodes directly connected to the root community $c_\lambda$, forming the 1‑level community set $\mathcal{C}^{(1)}$.}
    \label{fig: Coding_tree_1}
\end{figure}

\subsection{The Algorithm of Coding Tree Construction}\label{subapp: The Algorithm of Coding Tree Construction}
Given an undirected graph $\mathcal{G}=(\mathcal{V}, \mathcal{E})$, we initially treat each node $v_i \in \mathcal{V}$ as a individual community $c_i$, whose vertex set is $T_i=\{v_i\}$ with $T_i \subset \mathcal{V}$. We then organize all communities in $\mathcal{C}$ as leaf communities directly connected to the root community $c_\lambda$. 
\begin{figure*}[t]
    \centering
    \includegraphics[width=\linewidth]{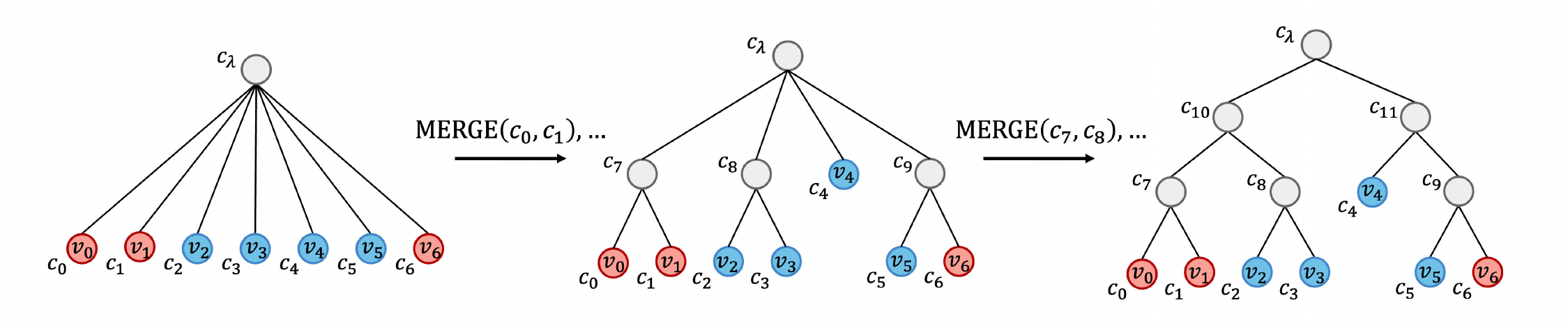}
    \vspace{-8pt}
    \caption{Illustration of the $\mathbf{MERGE}$ operation. In each iteration, two child communities of the root are greedily selected and merged to achieve the maximal reduction in SE, progressively forming a binary coding tree without height limitation.}
    \label{fig: Coding_tree_2}
    \vspace{-8pt}
\end{figure*}
\begin{figure*}[t]
    \centering
    \includegraphics[width=\linewidth]{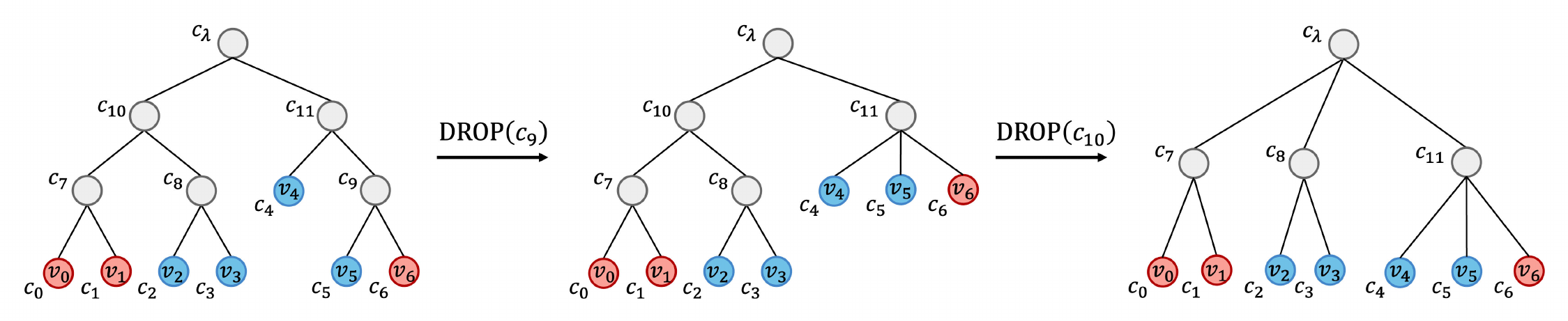}
    \vspace{-8pt}
    \caption{Illustration of the $\mathbf{DROP}$ operation. A selected intermediate community node $c_n$ is removed, and all of its child communities are directly reattached to its parent $c_n^-$, thereby compressing the coding tree to satisfy the predefined height constraint $h$.}
    \label{fig: Coding_tree_3}
    \vspace{-8pt}
\end{figure*}
At this point, the set of all these leaves corresponds to the $1$‑level community set $\mathcal{C}^{(1)}$ in the initial coding tree $\mathcal{T}^{(1)}$, as illustrated in the Figure~\ref{fig: Coding_tree_1}. To construct a coding tree T with minimal Structural Entropy (SE), we adopt two key operators, $\mathbf{MERGE}$ and $\mathbf{DROP}$. 
The MERGE operation incrementally combines candidate communities, while the DROP operation compresses the coding tree to the fixed height $h$. The formal definitions of these two operations are provided follows:

\begin{definition}[MERGE Operation]
Given two child communities $c_i$ and $c_j$ of the root community $c_\lambda$ in a coding tree $\mathcal{T}$, the operation $\mathbf{MERGE}(c_i, c_j)$ creates a new community node $c_m$ such that $c_m$ becomes the parent of $c_i$ and $c_j$, and simultaneously a child of $c_\lambda$. Formally:
\begin{align}
c_m.\mathrm{children} &= \{ c_i, c_j \}, \\
c_\lambda.\mathrm{children} &= \{ c_m \} \ \cup \ c_\lambda.\mathrm{children} .
\end{align}
\end{definition}

\begin{definition}[DROP Operation]
Given a community node $c_n$ and its parent community $c_n^{-}$ in a coding tree $\mathcal{T}$, the operation $\mathbf{DROP}(c_n)$ removes $c_n$ from $\mathcal{T}$ and reattaches all child communities of $c_n$ directly to $c_n^{-}$. Formally:
\begin{align}
c_n^{-}.\mathrm{children} &= c_n^{-}.\mathrm{children} \ \cup \ c_n.\mathrm{children}.
\end{align}
\end{definition}

To further facilitate the understanding of how SE guides the MERGE and DROP operations on communities, we define the SE of deduction from a parent community to one of its descendants.
Specifically, for a non‑leaf community node $c_\gamma$ and its descendant node $c_\alpha$ in the encoding tree $\mathcal{T}$, the SE of deduction is defined as:
\begin{equation}
    \mathcal{H}^{\mathcal{T}}(\mathcal{G}; (c_\gamma, c_\alpha])
    = \sum_{c_\beta,\, T_{\alpha} \subseteq T_{\beta} \subset T_\gamma}
    \mathcal{H}^{\mathcal{T}}(\mathcal{G}; c_\beta),
\end{equation}
where $T_{\beta}$ denotes the vertex set associated with an intermediate community $c_\beta$, which satisfies $T_{\alpha} \subseteq T_{\beta} \subset T_\gamma$.
This formulation captures the hierarchical contribution of all intermediate communities along the path from $c_\lambda$ to $c_\alpha$, thus providing a principled basis for evaluating how merging or restructuring affects the overall entropy.
It therefore serves as an essential criterion for selecting community pairs that achieve the maximal reduction in SE during the construction of the encoding tree.

The MERGE operation incrementally combines two child communities with the goal of minimizing SE, thereby producing a binary coding tree without height limitation. The procedure is illustrated in Figure~\ref{fig: Coding_tree_2}. Formally, for two child communities $c_i$ and $c_j$ of $c_\lambda$, the operation $\mathbf{MERGE}(c_i, c_j)$ generates a new parent community $c_m$. In each iteration, we greedily select the pair of communities that achieves the maximal reduction in SE:

\begin{equation}
\begin{split}
    & \mathbf{MERGE}(c_i,c_j) = \mathop{\arg\max}\limits_{c_i,\,c_j \in c_\lambda.\mathrm{children}}
     \left\{
       \mathcal{H}^{\mathcal{T}}(\mathcal{G})
       - \mathcal{H}^{\mathcal{T}^{\,\mathbf{MERGE}(c_i,c_j)}}(\mathcal{G})
     \right\}.
\end{split}
\end{equation}
When two communities $c_i$ and $c_j$ are merged into a new community node $c_m$, the structural quantities are updated according to:
\begin{gather} 
    g_{c_m}=g_{c_i}+g_{c_j}-2 \mathrm{Cut}\!\left(c_i,c_j\right), \\ \operatorname{vol}\left(c_m\right)=\operatorname{vol}\left(c_i\right)+\operatorname{vol}\left(c_j\right), 
\end{gather}
where $\mathrm{Cut}(c_i,c_j)$ represents the number of edges between $c_i$ and $c_j$. 
Subsequently, we substitute the $g_{c_m}$ and $\operatorname{vol}\left(c_j\right)$ into Eq.\ref{Eq: Structural Entropy} to derive the structural entropy associated with the merged community $c_m$, and thereby determine the SE of the coding tree after merging $c_i$ and $c_j$.
This iterative merging process continues until no further significant reduction in SE can be achieved, thereby resulting in a binary coding tree without height restrictions.

Subsequently, to adhere to the predefined height constraint $h$, we apply the DROP operation to compress the coding tree. Specifically, given a community node $c_n$ with parent $c_n^-$, $\mathbf{DROP}(c_n)$ removes $c_n$ and directly reattaches all of its children to $c_n^-$, as illustrated in the Figure~\ref{fig: Coding_tree_3}. At each DROP step, we greedily select the community node whose removal results in the smallest increase in SE:
\begin{equation}
\begin{split}
    & \mathbf{DROP}(c_n) = \mathop{\arg\min}\limits_{c_n \in \mathcal{T}, c_n \neq c_\lambda, c_n \notin \mathcal{C}^{(1)}}
    \left\{\mathcal{H}^{\mathcal{T}^{\mathbf{DROP}(c_n)}}(\mathcal{G})-\mathcal{H}^\mathcal{T}(\mathcal{G}) \right\}
\end{split}
\end{equation}
By iteratively applying this compression process, we obtain an optimal coding tree $\mathcal{T}^\ast$. The pseudocode for the coding tree construction is provided in Appendix~\ref{subapp: Pseudocode of Coding Tree Construction}, and the analysis of its algorithmic complexity is presented in Appendix~\ref{subapp: Time Complexity of Coding Tree Construction}.

In summary, through greedy iterative MERGE and subsequent DROP operations guided by SE minimization, we obtain $\mathcal{T}^{(h)}$. This tree provides a hierarchical partition of the graph $\mathcal{G}$, minimizes structural uncertainty, and effectively captures the underlying essential structure. The nodes in the original graph are naturally organized into communities $\mathcal{C}_\text{homo}=\{c_1,c_2, \ldots, c_m \ | \ c_i \notin \mathcal{C}^{(1)}, \ 1\leq i \leq m, \ m < |\mathcal{V}|, \ \ T_i \subset \mathcal{V}, \ |T_i| \geq 1\}$, which represent the finest granularity of the hierarchical community partition. Consequently, while measuring the information embedded in $\mathcal{G}$, the procedure simultaneously detects and explicitly delineates the homophilic structures inherent in the graph.

\begin{algorithm}[h]
\caption{Coding tree construction with fixed height $h$ via Structural Entropy minimization.}
\label{code:coding_tree} 
\begin{flushleft}
\textbf{Input:} an undirected graph $\mathcal{G}=(\mathcal{V}, \mathcal{E})$, a tree height $h>1$\\
\textbf{Output:} a coding tree $\mathcal{T}^{(h)}$ with height $h$
\end{flushleft}

\begin{algorithmic}[1]
\STATE Initialize a coding tree $\mathcal{T} = \mathcal{T}^{(1)}$ with a root community node $c_\lambda$. Each node in the $\mathcal{V}$ is treated as a singleton community $c_i$ whose
vertex set is denoted as $T_i=v_i$, forming the initial set of leaf communities $\mathcal{C}^{(1)}$ connected to the root, i.e. $c_i \in c_\lambda.\mathrm{children}$.
\STATE // \textit{Step 1}: Full-height binary coding tree construction
\WHILE{$|c_i.children|>2$} {
  \STATE Select $c_i$ and $c_j$ from $c_\lambda.children$, conditioned on \\
  $\mathop{\arg\max}\limits_{c_i,\,c_j \in c_\lambda.\mathrm{children}}
 \left\{
   \mathcal{H}^{\mathcal{T}}(\mathcal{G})
   - \mathcal{H}^{\mathcal{T}^{\,\mathbf{MERGE}(c_i,c_j)}}(\mathcal{G})
 \right\}$; \\
  \STATE $\text{MERGE}(c_i, c_j)$;
}
\ENDWHILE

\STATE // \textit{Step 2}: Compressing $\mathcal{T}$ to the specific height $h$;
\WHILE{$\text{Height}(\mathcal{T})>h$} {
  \STATE Select $c_i$ from $\mathcal{T}$, conditioned on \\
  $\mathop{\arg\min}\limits_{c_n \in \mathcal{T}, c_n \neq c_\lambda, c_n \notin \mathcal{C}^{(1)}}
    \left\{\mathcal{H}^{\mathcal{T}^{\mathbf{DROP}(c_n)}}(\mathcal{G})-\mathcal{H}^\mathcal{T}(\mathcal{G}) \right\}$;
  \STATE $\text{DROP}(c_i)$;
}
\ENDWHILE
\STATE return $\mathcal{T^\ast}$;
\end{algorithmic} 
\end{algorithm}

\subsection{Pseudocode of Coding Tree Construction}\label{subapp: Pseudocode of Coding Tree Construction}
In this section, we present the detailed pseudocode for the construction of the coding tree, as shown in Algorithm \ref{code:coding_tree} . The algorithm begins with an initial tree where each node is treated as a leaf community, then iteratively applies MERGE operations to minimize SE, followed by DROP operations to satisfy the predefined tree height, ultimately producing the optimal coding tree $\mathcal{T}^\ast$.

\subsection{Time Complexity of Coding Tree Construction}\label{subapp: Time Complexity of Coding Tree Construction}
Given an undirected graph $\mathcal{G} = (\mathcal{V}, \mathcal{E})$ with $|\mathcal{V}| = n$ and $|\mathcal{E}| = m$, the complexity of the Algorithm \ref{code:coding_tree} is $O \ \big(h_{\text{max}} (m \log n + n)\big),$
where $h_{\text{max}}$ denotes the height of the coding tree $\mathcal{T}$ after performing the MERGE operations without imposing tree height constraint. In practice, during SE minimization, the coding tree $\mathcal{T}$ tends to be balanced, and thus the height $h_{\text{max}}$ is approximately $O(\log n)$. 
Since real-world graphs typically satisfy $m \gg n$, the complexity almost scales linearly with $m$.

\subsection{Community-variant Semantics Aggregation Prompt Template} \label{subapp: Semantics Aggregation Prompt Template}
In this section, we comprehensively provide the details of the prompt templates utilized in our $\text{HS}_2\text{C}$ framework. Specifically, for node-level tasks, we elaborate on the homophilic semantics aggregation prompts designed for different community types (defined in Section~\ref{section4_2} and visualized in Figure~\ref{fig:Overall framework}), as shown in Box~\ref{APPBox: Specific Template}, \ref{APPBox: Common-shared Template}, \ref{APPBox: Mixed Template}. We also present the inference prompts employed for downstream node classification (refer to Section~\ref{section4_4}) in Box~\ref{APPBox: Node Classification Template}. Regarding graph-level tasks, we elaborate on the community semantics aggregation templates designed for sentiment analysis and molecular property prediction, which are illustrated in Box~\ref{APPBox: Sentiment Template}, \ref{APPBox: Molecule Template}. The corresponding prompts for graph-level classification are provided in Box~\ref{APPBox: Graph Classification Template}.

\subsubsection{Aggregation Prompt Template}\label{app:Aggregation_Prompt_Template}
\begin{promptbox}[H]
    \centering
    \begin{tcolorbox}[colback=white,colframe=black,title=The Prompt Template for Specific Target type Community.]
    \textbf{Specific Template:} Given a single target paper and a group of background papers from the same community within a citation graph. Each paper is represented by its title and abstract. The background papers, which are connected to the target paper and may share a common semantic focus or research topic with the target paper. \\[3pt]
    Target paper:
    \textcolor[rgb]{0.90,0.55,0.50}{\texttt{\{RAW\_TEXT\}}}\\[3pt]
    \noindent Background papers:
    \textcolor[rgb]{0.36,0.61,0.84}{\texttt{\{RAW\_TEXTS\}}}\\[3pt]
    Task:
    Summarize the content of those background papers that are semantically aligned with the target paper, focus on shared research disciplines, domains, directions, or topics. Please write the summary in a cohesive and formal academic style. \\[3pt]
    Answer:
    \end{tcolorbox}
    \caption{The Aggregation prompt template for Specific Target type Community used in $\text{HS}_2\text{C}$.}
    \label{APPBox: Specific Template}
\end{promptbox}

\begin{promptbox}[H]
    \centering
    \begin{tcolorbox}[colback=white,colframe=black,title=The Template for Common-shared Community.]
    \vspace{-5pt}
    \textbf{Common-shared Template:} Given a group of target papers and a single background paper from the same community within a citation graph. Each paper is represented by its title and abstract. The background paper, which is connected to the target papers and may share a common semantic focus or research topic with the target papers. \\[3pt]
    Target papers:
    \textcolor[rgb]{0.90,0.55,0.50}{\texttt{\{RAW\_TEXTS\}}}\\[3pt]
    Background paper:
    \textcolor[rgb]{0.36,0.61,0.84}{\texttt{\{RAW\_TEXT\}}}\\[3pt]
    Task:
    Summarize the content of the background paper that is semantically aligned with the target papers, focus on shared research disciplines, domains, directions, or topics. Please write the summary in a cohesive and formal academic style. \\[3pt]
    Answer:
    \vspace{-5pt}
    \end{tcolorbox}
    \caption{The Aggregation prompt template for Common-shared Target type Community used in $\text{HS}_2\text{C}$.}
    \label{APPBox: Common-shared Template}
    \vspace{-2pt}
\end{promptbox}

\begin{promptbox}[H]
    \centering
    \begin{tcolorbox}[colback=white,colframe=black,title=The Template for Mixed Target type Community.]
    \vspace{-5pt}
    \textbf{Mixed Target Template:} Given a group of target papers and a group of background papers from the same community within a citation graph. Each paper is represented by its title and abstract. The background papers, which are connected to the target papers and may share a common semantic focus or research topic with the target paper. \\[3pt]
    Target papers:
    \textcolor[rgb]{0.90,0.55,0.50}{\texttt{\{RAW\_TEXTS\}}}\\[3pt]
    Background papers:
    \textcolor[rgb]{0.36,0.61,0.84}{\texttt{\{RAW\_TEXTS\}}}\\[3pt]
    Task:
    Summarize the content of those background papers that are semantically aligned with the target papers, focus on shared research disciplines, domains, directions, or topics. Please write the summary in a cohesive and formal academic style. \\[3pt]
    Answer:
    \vspace{-5pt}
    \end{tcolorbox}
    \caption{The Aggregation template for Mixed type Community.}
    \label{APPBox: Mixed Template}
    \vspace{-5pt}
\end{promptbox}

\begin{promptbox}[H]
    \centering
    \begin{tcolorbox}[colback=white,colframe=black,title=The Prompt Template for Sentiment Dataset.]
    \vspace{-5pt}
    \textbf{Graph-SST2 Template:} \\[3pt]
    Analyze the following words from a sentence:  \\[3pt]
    \textcolor[rgb]{0.90,0.55,0.50}{\texttt{\{RAW\_TEXTS\}}}\\[3pt]
    Task: Summarize the semantic meaning and sentiment tendency of these words in a concise phrase.\\[3pt]
    Focus on: \\[1pt]
    1. The collective semantic meaning. \\[1pt]
    2. Overall sentiment tendency (Positive/Negative). \\[3pt]
    Answer:
    \vspace{-5pt}
    \end{tcolorbox}
    \caption{The Aggregation template for Sentiment Analysis Data}
    \label{APPBox: Sentiment Template}
    \vspace{-5pt}
\end{promptbox}

\begin{promptbox}[H]
    \centering
    \begin{tcolorbox}[colback=white,colframe=black,title=The Prompt Template for Molecule Dataset.]
    \vspace{-5pt}
    \textbf{Molecule Template:} \\[3pt]
    Analyze the following atoms in a molecular substructure: \textcolor[rgb]{0.90,0.55,0.50}{\texttt{\{RAW\_TEXTS\}}}\\[3pt]
    Task: Describe this substructure in a concise phrase.\\[3pt]
    Focus on: \\[1pt]
    1. The functional group or structural motif. \\[1pt]
    2. Any notable chemical properties. \\[3pt]
    Answer:
    \vspace{-5pt}
    \end{tcolorbox}
    \caption{The Aggregation prompt template for Molecule Property Prediction Datasets.}
    \label{APPBox: Molecule Template}
\end{promptbox}

\subsubsection{Downstream Task Prompt Template} \label{subapp: Downstream Task Prompt Template}

\begin{promptbox}[H]
    \centering
    \begin{tcolorbox}[colback=white,colframe=blue!30!black,title=The Prompt Template for Node-level classification.]
    \vspace{-5pt}
    \textbf{Classification Template:} Given a citation graph, the 0th node (target paper) has the following information: \textcolor[rgb]{0.90,0.55,0.50}{\texttt{\{TARGET\_RAW\_TEXT\}}}\\[3pt]
    The target paper is connected to the following papers: \textcolor[rgb]{0.746,0.427,0.765}{\texttt{\{NEIGHBOR\_RAW\_TEXTS\}}}\\[3pt]
    Question: Based on the features of the target paper and its citation network, please determine the most appropriate ArXiv CS sub-category for the target paper.\\[3pt]
    Categories: \textcolor[rgb]{1.0,0.753,0.0}{\texttt{\{CATEGORY\_LIST\}}}.\\[3pt]
    Please think about the categorization of the target paper in a structured manner, and only output the single most relevant category of the target paper. Do not give any reasoning or extra text for your answer.\\[3pt]
    Answer:
    \vspace{-5pt}
    \end{tcolorbox}
    \caption{The Downstream Node-level Task prompt template.}
    \label{APPBox: Node Classification Template}
\end{promptbox}

\begin{promptbox}[H]
    \centering
    \begin{tcolorbox}[colback=white,colframe=blue!30!black,title=The Prompt Template for Graph-level classification.]
    \vspace{-5pt}
    \textbf{Graph-SST2 Template:}\\[3pt]
    Given a sentiment analysis graph, the task is to predict a binary sentiment label (Positive / Negative) for the graph. \\[3pt]
    Hierarchical Structure: \\[1pt]
    \textcolor[rgb]{0.90,0.55,0.50}{\texttt{\{CONDENSED\_RAW\_TEXTS\}}}\\[3pt]
    Label Descriptions: \\[1pt]
    Negative - the sentence expresses negative sentiment. \\[1pt]
    Positive - the sentence expresses positive sentiment. \\[3pt]
    Output EXACTLY one word from: [Positive, Negative].\\[3pt]
    Answer:
    \vspace{-5pt}
    \end{tcolorbox}
    \caption{The Downstream Graph-level Task prompt template.}
    \label{APPBox: Graph Classification Template}
\end{promptbox}

\section{Supplementary Experimental settings}\label{app: Supplementary Experimental settings}
\subsection{Datasets}\label{subapp: Dataset Description}
In this section, we introduce the detailed information of the experimental datasets. And we summarize the statistics of node-level dataset in the Table~\ref{tab:node-level Dataset Statistic}, including data splits, evaluation metrics. We summarize the statistics of graph-level dataset in the Table~\ref{tab:graph-level Dataset Statistic}.

\begin{table*}[ht]
    \centering
    \setlength{\abovecaptionskip}{0.1cm}
    \caption{\textbf{Detailed Information and Statistics of Experimental Node-level Datasets.}}
    \label{tab:node-level Dataset Statistic}
    \setlength{\abovecaptionskip}{0cm}
    \small
    \renewcommand{\arraystretch}{1.1}
    \resizebox{\linewidth}{!}{
    \begin{tabular}{r|cccccccccc}
    \toprule
    & OGBN-ArXiv & TAPE & Instagram & Citeseer & Reddit & Product-subset & Book-Hete & Book-Homo & DBLP-Hete & DBLP-Homo \\
    \midrule
    Classes\# & 40 & 40 & 2 & 6 & 2 & 43 & 8 & 8 & 9 & 9 \\
    Node\# & 169,343 & 57,471 & 11,339 & 3,186 & 33434 & 45,855 & 786,257 & 594,484 & 1,989,010 & 964,350 \\
    Edge\# & 1,166,243 & 122,835 & 155,349 & 8,450 & 301,906 & 111,638 & 9,035,291 & 7,614,902 & 29,830,033 & 16,679,526 \\
    Target\# & 7,840 & 8,621 & 9,089 & 2,566 & 6,686 & 9,171 & 5,877 & 5,877 & 35,981 & 35,981\\
    \midrule
    Metrics & ACC & ACC & ACC & ACC & ACC & ACC & ACC & ACC & ACC & ACC \\
    \bottomrule
    \end{tabular}
    }
\end{table*}

\begin{table*}[t]
    \centering
    \caption{\textbf{Detailed Information and Statistics of Experimental Graph-level Datasets.}}
    \label{tab:graph-level Dataset Statistic}
    \small
    \renewcommand{\arraystretch}{1.2}
    \resizebox{\linewidth}{!}{
    \begin{tabular}{l|ccc|ccccccc}
    \toprule
    \textbf{Category} & \multicolumn{3}{c|}{\textbf{Sentiment Analysis (Text-to-Graph)}} & \multicolumn{7}{c}{\textbf{MoleculeNet (Molecular Graphs)}} \\
    \midrule
    \textbf{Dataset} & \textbf{Graph-SST2} & \textbf{Graph-SST5} & \textbf{Graph-Twitter} & \textbf{BACE} & \textbf{BBBP} & \textbf{ClinTox} & \textbf{HIV} & \textbf{MUV} & \textbf{SIDER} & \textbf{Tox21} \\
    \midrule
    \# Graphs & 70,042 & 11,855 & 6,940 & 1,513 & 2,039 & 1,478 & 41,127 & 93,087 & 1,427 & 7,831 \\
    \# Tasks & 1 & 1 & 1 & 1 & 1 & 2 & 1 & 17 & 27 & 12 \\
    \# Classes & 2 & 5 & 3 & 2 & 2 & 2 & 2 & 2 & 2 & 2 \\
    \midrule
    Metrics & ACC & ACC & ACC & ROC-AUC & ROC-AUC & ROC-AUC & ROC-AUC & ROC-AUC & ROC-AUC & ROC-AUC \\
    \bottomrule
    \end{tabular}
    }
\end{table*}

\subsubsection{Node-level Dataset Description}\label{subsubapp: Node-level Dataset Description}

\begin{itemize}[leftmargin=*,itemsep=3pt,topsep=5pt]
    \item \textbf{OGBN-ArXiv}~\citep{ogb_arxiv} is a citation network comprising $169,343$ Arxiv CS papers and their citation relationships from 40 different academic disciplines, such as cs.AI (Artificial Intelligence) and cs.DB (Databases). Each node represents a paper, node text attribute is the title and abstract of paper, and each edge represents a citation relationship. The objective is to predict the subject area of each paper. We randomly sample $7,840$ nodes from the default test set (papers published after $2018$) as the test subset for node classification.

    \item \textbf{TAPE}~\citep{tape} is a citation graph consisting of $57,471$ computer science papers published on arXiv in $2023$ or later, connected by citation edges. Each node corresponds to a paper and includes raw-text attributes such as title and abstract. The task is to predict the arXiv CS subject area (one of $40$ classes) for each paper—similar to OGBN‑ArXiv. Following the default test split, we randomly selected $15\%$ nodes ($8,621$) as the test set for node classification.
    
    \item \textbf{Instagram}~\citep{instagram} is a social network graph constructed from follower–followee relationships on the Instagram platform. Nodes represent individual users, and directed edges capture the following connections between them. Each node is associated with raw‑text attributes, including the user’s profile biography and recent caption texts. The classification task is to determine whether a user belongs to one of two predefined categories: normal user or commercial user. For evaluation, we use the default test set, which consists of $9,089$ nodes.

    \item \textbf{Citeseer}~\citep{citeseer} is a citation network comprising $3,186$ scientific publications, where nodes represent individual papers and edges indicate citation relationships. Each node is associated with text attributes derived from the paper’s title and abstract. Every node is labeled according to one of six predefined research categories, such as Agents, Information Retrieval. Following the dataset’s default split, we evaluate on the test set containing $2,566$ nodes.

    \item \textbf{Reddit}~\citep{instagram} is a social graph derived from user interaction data on the Reddit platform, where nodes represent individual users and edges denote social connections such as co‑participation in discussion threads. Each node is accompanied by raw‑text attributes consisting of the set of subreddits in which the user has historically published or commented. Nodes are labeled to indicate whether the user is popular or normal. For evaluation, we randomly sample $20\%$ of the nodes as the test set, resulting in a test subset containing $6,686$ nodes.

    \item \textbf{Product-subset}~\citep{tape, taglas} is a co‑purchase graph derived from the Amazon product network, where each node represents a product item and an edge indicates that two products are frequently co‑purchased. We adopt the version provided by TAPE~\citep{tape}, which is a subset of the original OGB‑Products dataset~\citep{ogb_arxiv}. The original dataset contains $2,449,029$ nodes and $61,859,140$ edges, while the Product‑subset comprises $54,025$ nodes and $144,638$ edges. Each node is associated with text attributes such as the product title and description. The classification task is to predict the product category for each item. For evaluation, we use the default test split of the dataset, which includes $9,171$ nodes.

    \item \textbf{Book-Hete}~\citep{htag_dataset, book_dataset} is a large-scale heterogeneous literature network constructed from a subset of GoodReads~\citep{book_dataset}, the world’s largest social cataloging platform for books, covering book tracking, recommendations, reviews, and discussions. We adopt the processed version released by HTAG~\citep{htag_dataset}. Book-Hete consists of three types of entities: books (594,484 nodes), authors (147,863 nodes), and publishers (43,910 nodes), connected by three types of directed relations: a book is similar to another book, an author writes a book, and a book is published by a publisher. Book nodes are associated with textual attributes, including titles and descriptions, whereas author and publisher nodes do not contain textual features. Each book is annotated with one or more of eight genres (e.g., Children, Romance, Poetry), formulating a multi-label classification task. Following a chronological splitting strategy, books published from 2017 onwards are reserved for model evaluation
    
    \item \textbf{Book-Homo}~\citep{htag_dataset, book_dataset}. To facilitate a comprehensive comparative analysis, we developed Book-Homo, a homogeneous counterpart derived from the Book-Hete dataset. This version simplifies the network by retaining only book nodes (594,484) and their internal "is similar to" directed relations. Each node is represented by its title and description as textual features. The dataset maintains the same multi-label classification task across eight genres and follows the identical chronological split, testing on books published from 2017 onwards. 

    \item \textbf{DBLP-Hete}~\citep{htag_dataset, dblp_dataset} is a large-scale heterogeneous academic dataset constructed from the Digital Bibliography \& Library Project (DBLP), a comprehensive and publicly available database of computer science publications. We adopt the processed version released by HTAG~\citep{htag_dataset}, which is derived from the original data provided by AMiner~\citep{dblp_dataset}. The dataset contains three types of entities: Papers (964,350 nodes), Authors (958,961 nodes), and Fields of Study (FoS) (65,699 nodes). The network architecture incorporates three types of directed relations: paper-cites-paper, author-writes-paper, and paper-has-topic-of-FoS. Each paper is characterized by its title and abstract, while other entity types are represented by their respective textual descriptions. Furthermore, papers are categorized into nine research domains based on their publication venues, such as Artificial Intelligence and Computer Networks. For model evaluation, we employ a chronological split: papers published from 2019 onwards serve as the test set. The task is to predict the topic label of each paper given its textual content and citation relationships.

    \item \textbf{DBLP-Homo}~\citep{htag_dataset, dblp_dataset}. To facilitate a comprehensive comparative analysis, we constructed DBLP-Homo, a homogeneous version of the academic network derived from the same source. This dataset focuses exclusively on the paper-level topology, comprising 964,350 paper nodes interconnected by a single relation type: citations. Each node is enriched with the paper's title and abstract as textual features. Consistent with DBLP-Hete, each paper is labeled with one of nine thematic topics (e.g., Theoretical Computer Science, Database and Data Mining) based on its publication venue. We adopt the same time-based splitting strategy as in DBLP-Hete, where models are evaluated on papers published from 2019 onwards.

\end{itemize}

\subsubsection{Graph-level Dataset Description}\label{subsubapp: Graph-level Dataset Description}
\begin{itemize}[leftmargin=*,itemsep=3pt,topsep=5pt]
    \item \textbf{Graph-SST2}~\citep{sst_dataset} is a graph-structured sentiment classification benchmark derived from the Stanford Sentiment Treebank (SST-2). In this dataset, each sentence is transformed into a directed graph using dependency parsing, where nodes represent individual tokens and edges denote their syntactic dependencies. Each graph is assigned a binary sentiment label (positive or negative) based on the original sentence sentiment.

    \item \textbf{Graph-SST5}~\citep{sst_dataset} is a fine-grained sentiment classification benchmark that transforms the Stanford Sentiment Treebank (SST-5) into a graph-structured format. Each graph represents a sentence, where nodes correspond to individual tokens and edges capture the syntactic dependencies between them. Unlike binary sentiment analysis, Graph-SST5 assigns each graph to one of five sentiment intensities: very negative, negative, neutral, positive, or very positive.

    \item \textbf{Graph-Twitter}~\citep{sst_dataset} is a graph-level sentiment classification benchmark constructed from social media posts. In this dataset, each tweet is converted into a dependency tree-based graph, where nodes represent individual tokens and edges capture the syntactic dependencies extracted from the informal text. The task is formulated as a multi-class classification problem, where the goal is to categorize each graph into one of three sentiment labels: positive, neutral, or negative.

    \item \textbf{BBBP}~\citep{molecule_dataset} is a binary classification dataset from MoleculeNet, derived from a study on the modeling and prediction of barrier permeability. It contains over 2,000 compounds labeled according to their ability to penetrate the blood-brain barrier (BBB). As the BBB acts as a selective membrane separating circulating blood from the brain extracellular fluid, its permeability is a critical factor in the development of drugs targeting the central nervous system (CNS). The task is to predict whether a molecule is permeable (positive) or non-permeable (negative). To evaluate the model's generalization capability across structurally diverse compounds, we employ scaffold splitting for dataset partitioning.

    \item \textbf{ClinTox}~\citep{molecule_dataset} 
    is a multi-task binary classification benchmark designed to distinguish between drugs approved by the FDA and those that have failed clinical trials due to toxicity. The dataset comprises 1,491 drug compounds with known chemical structures. It involves two classification tasks: (1) predicting clinical trial toxicity (toxic vs. non-toxic) and (2) predicting FDA approval status (approved vs. not approved). The list of FDA-approved drugs is derived from the SWEETLEAD database, while drugs failing due to toxicity are compiled from the Aggregate Analysis of ClinicalTrials.gov (AACT) database. Given the structural diversity of pharmaceutical compounds, we employ scaffold splitting to evaluate the model's generalization ability.

    \item \textbf{MUV}~\citep{molecule_dataset} (Maximum Unbiased Validation) is a benchmark dataset curated from the PubChem BioAssay database, specifically designed to validate virtual screening techniques. It comprises 17 binary classification tasks, representing different biological targets, with approximately 93,000 compounds. A unique feature of MUV is its construction using refined nearest neighbor analysis, which minimizes spatial embedding bias and artificial enrichment of active compounds. This rigorous design ensures that high performance is achieved through learning genuine chemical features rather than exploiting dataset artifacts. We employ scaffold splitting to evaluate the model's robustness in predicting activity for structurally novel compounds.

    \item \textbf{SIDER}~\citep{molecule_dataset} is a database of marketed drugs and adverse drug reactions (ADR), curated to facilitate the study of drug safety and pharmacovigilance. The version used in this work, derived from the DeepChem benchmark, groups reported side effects into 27 system organ classes (SOC) according to the MedDRA classification system. The dataset contains 1,427 approved drugs with known chemical structures. The task is formulated as a multi-task binary classification problem, where the goal is to predict the presence or absence of side effects in each of the 27 organ systems for a given drug. Due to the diverse chemical nature of marketed drugs, we employ scaffold splitting to evaluate the model's generalization capability on structurally distinct compounds.
\end{itemize}

\subsection{Baselines} \label{subapp: baselines}
In this section, we introduce the detailed information of baselines in  \textbf{Section~\ref{section5} Experiments}. 
For the node-level task, we compare our proposed $\text{HS}_2\text{C}$ with 11 baseline methods, which we generally divide into three categories, 
\textbf{(1) Graph Neural Network (GNN) Methods}, including GCN~\citep{baseline_gcn_kipf2016gcn}, GAT~\citep{baseline_gat_velivckovic2017gat}, GIN~\citep{baseline_gin_xu2018gnn}, GraphSAGE~\citep{baseline_graphsage_hamilton2017graphsage}. \textbf{(2) Traditional Neighbor Sample Methods}, including Random~\citep{Chen_GraphLLM2024KDD, rw_chen_icml2024llaga, rw_ye_EACL2024instructglm, rw_wu_icml2025when}, Degree~\citep{baseline_degree_ali2024degree}, Number, RAG~\citep{baseline_rag_li2025large}. \textbf{(3) Graph Skeleton Methods}, including Skeleton $\alpha$, $\beta$ and $\gamma$~\citep{baseline_skeleton_cao2024graph}. 

Beyond node-level tasks, to evaluate the generalization ability of our proposed method, we further extend $\text{HS}_2\text{C}$ to graph-level tasks. For the graph-level classification, we compare $\text{HS}_2\text{C}$ with 9 baselines, grouped into three categories:
\textbf{(1) Traditional Neighbor Sample Methods}, including Random~\citep{Chen_GraphLLM2024KDD, rw_chen_icml2024llaga, rw_ye_EACL2024instructglm, rw_wu_icml2025when}, Degree~\citep{baseline_degree_ali2024degree}, RAG~\citep{baseline_rag_li2025large}.
\textbf{(2) Graph Skeleton Methods}, including Skeleton $\alpha$, $\beta$ and $\gamma$~\citep{baseline_skeleton_cao2024graph}.
\textbf{(3) Graph Coarsening Methods}, including A-CM~\citep{aconvmatch}, FGC~\citep{fgc}, UGC~\citep{ugc}. The detailed descriptions of these baselines are as follows:
\begin{itemize}[leftmargin=*,itemsep=3pt,topsep=5pt]
    \item[\textbf{(1)}] \textbf{Graph Neural Network (GNN) Methods}
        For all GNN methods, the set of target nodes used for testing is identical to that in our experiments, while the remaining nodes are used as the training and validation sets for model training. We utilize the default node features provided by the corresponding datasets.
        \item \textbf{GCN}~\citep{baseline_gcn_kipf2016gcn}: learns node representations by iteratively aggregating and transforming features from a node’s local neighborhood. In each layer, a node updates its feature by combining its own representation with those of its neighbors through a normalized aggregation scheme, which ensures that information is propagated while preventing scale explosion. Formally, given an input feature matrix $X$ and an adjacency matrix $A$, a single GCN layer is typically defined as
        \[
        H^{(l+1)} = \sigma\ \big(\widetilde{D}^{-\tfrac{1}{2}}\,\widetilde{A}\,\widetilde{D}^{-\tfrac{1}{2}}\,H^{(l)} W^{(l)}\big),
        \]
        where $\widetilde{A} = A + I$ is the adjacency matrix with added self‑connections, $\widetilde{D}$ is the corresponding degree matrix, $W^{(l)}$ is a learnable weight matrix, and $\sigma(\cdot)$ is a non‑linear activation function. By stacking multiple layers, GCN enables nodes to capture higher‑order neighborhood information.

        \item \textbf{GAT}~\citep{baseline_gat_velivckovic2017gat}: introduces an attention mechanism to selectively aggregate neighborhood information. GAT learns attention coefficients that indicate the relative importance of each neighbor’s features for a given target node. Specifically, for a node $i$ with a neighbor $j \in \mathcal{N}(i)$, a shared attention mechanism computes a normalized attention weight $\alpha_{ij}$ based on their transformed feature vectors:
        \begin{equation*}
            \begin{split}
                & e_{ij} = \text{LeakyReLU} \ \big(a^\top [W\mathbf{h}_i \,\|\, W\mathbf{h}_j]\big), \\[2pt]
                &\alpha_{ij} = \text{softmax}_j(e_{ij}),
            \end{split}
        \end{equation*}
        where $W$ is a learnable weight matrix, $a$ is a learnable attention vector, and $\|\,$ denotes concatenation. The attention coefficients $\alpha_{ij}$ are then used to weight the neighbor features in the aggregation:
        \[
        \mathbf{h}_i' = \sigma \ \bigg(\sum_{j \in \mathcal{N}(i)} \alpha_{ij} W \mathbf{h}_j\bigg),
        \]
        where $\sigma(\cdot)$ is a non‑linear activation function.  
        By stacking multiple layers and employing multi‑head attention to stabilize the learning process, GAT allows nodes to focus on the most relevant neighbors.
        
        \item \textbf{GIN}~\citep{baseline_gin_xu2018gnn}: In each layer, a node updates its representation by summing the features of its neighbors together with its own features, followed by a learnable transformation. Formally, for a node $v$, the update rule in layer $k$ is:
        \[
        \mathbf{h}_v^{(k)} = \text{MLP}^{(k)} \ \bigg( (1+\epsilon^{(k)}) \cdot \mathbf{h}_v^{(k-1)} \;+\; \sum_{u \in \mathcal{N}(v)} \mathbf{h}_u^{(k-1)} \bigg),
        \]
        where $\mathbf{h}_v^{(k-1)}$ denotes the feature of node $v$ from the previous layer, $\epsilon^{(k)}$ is either a learnable parameter or a fixed scalar, and $\text{MLP}^{(k)}$ is a multilayer perceptron applied after aggregation. By using a sum aggregation, GIN can distinguish different graph structures. 
        
        \item \textbf{GraphSAGE}~\citep{baseline_graphsage_hamilton2017graphsage}: At each layer, GraphSAGE first samples a fixed-size subset of neighbors for each node to control computational cost. Then, it applies a learnable aggregation function (e.g., mean aggregator, LSTM-based aggregator, or pooling aggregator) to combine the neighbor features. The aggregated neighborhood representation is concatenated with the node’s own features and passed through a learnable weight matrix followed by a non-linear activation. Formally, for node $v$ in layer $k$:
        \begin{equation*}
        \begin{split}
            \mathbf{h}_v^{(k)} = \sigma \ \bigg( W^{(k)} \cdot \big[ \, 
            & \mathbf{h}_v^{(k-1)} \, \| \\
            & \text{AGG.}^{(k)} \big(\{ \mathbf{h}_u^{(k-1)},\, u \in \mathcal{N}(v) \}\big) \big] \bigg),
        \end{split}
        \end{equation*}
        where $\text{AGG.}^{(k)}(\cdot)$ denotes a differentiable aggregator function, $W^{(k)}$ is a learnable weight matrix, $\|$ denotes concatenation, and $\sigma(\cdot)$ is a non-linear activation function. 

    \item[\textbf{(2)}] \textbf{Traditional Neighbor Sample Methods}
        We consider several traditional neighbor sampling strategies. These methods differ in the criteria used to select 1‑hop neighbors for each target node, yet all follow the same procedure of preparing neighbor text for the prompt. Specifically, each method determines a subset of neighbors based on its sampling rule, concatenates the text attributes of the selected neighbors, and then combines this concatenated neighbor text with the central node’s text to construct the input prompt. The overall sampling process in all methods is is limited by the predefined token number of the LLM context constraint.
        \item \textbf{Random}~\citep{rw_tang_sigir2024graphgpt,rw_chen_icml2024llaga,rw_ye_EACL2024instructglm,rw_wu_icml2025when}: randomly samples 1-hop neighbors for each target node, without considering any structural or attribute-based criteria, and concatenates the text attributes of the neighboring nodes. Then the concatenated neighbor text is combined with the central node’s text attribute and filled into the prompt \ref{Box: Node Classification Template}. 

        \item \textbf{Degree}~\citep{baseline_degree_ali2024degree, baseline_degree_Zhang2023degree_importance}: selects neighbors based on their node degrees. Neighbors with higher degrees are prioritized for sampling. The sampling process is limited by the predefined token number of the LLM context constraint.

        \item \textbf{Number}: samples neighbors based on the token lengths of their textual attributes, following the principle of including as many neighbors as possible. In other words, neighbors with shorter text attributes are more likely to be selected. The sampling process is limited by the predefined token number of the LLM context constraint.

        \item \textbf{RAG}~\citep{baseline_rag_li2025large}: adopts a retrieval‑augmented generation (RAG) strategy. For each target node, it retrieves semantically relevant neighbors and uses the text of the retrieved neighbors as additional input context, thereby enhancing classification with semantically consistent information.

    \item[\textbf{(3)}] \textbf{Graph Skeleton Methods}~\citep{baseline_skeleton_cao2024graph}
        In the graph skeleton framework, background nodes are further divided into two categories according to their structural roles. A bridge node refers to a background node that connects multiple target nodes or communities, thereby contributing critical structural information and enabling long‑range message propagation across the graph. In contrast, an affiliation node is a background node associated exclusively with a single target node, typically encoding fine‑grained local attributes or contextual details with limited structural diversity. For a fair comparison in our baseline experiments, we strictly follow the original Graph Skeleton procedure while aligning the semantic condensation step with that of our own method. In particular, the condensation of background nodes is performed using the same LLM, LLaMA‑3.1‑8B‑Instruct, as in our proposed approach.
    
        \item \textbf{Skeleton $\alpha$} merges background nodes based on a structural equivalence criterion. Each background node is characterized by its Multiple Structure-Set (MSS), defined by the collection of target nodes it can reach and the corresponding shortest-path distances. Background nodes with identical MSS are considered functionally equivalent in message passing and are merged into a single representative node, with their features aggregated accordingly.
        
        \item \textbf{Skeleton $\beta$} further relaxes the strict equivalence criteria imposed in Skeleton $\alpha$.  Instead of requiring background nodes to match both the set of reachable target nodes and their respective path distances, Skeleton $\beta$ groups background nodes solely based on the set of target nodes they connect to, while disregarding specific distance information.

        \item \textbf{Skeleton $\gamma$} provides the highest level of condensation by further compressing affiliation nodes. It first applies the Skeleton $\beta$ strategy to condense bridge nodes, and then directly merges the remaining affiliation nodes into their corresponding target nodes. This is achieved by updating the feature vector of each target node through the aggregation of its own features with those of its affiliated background nodes. After feature aggregation, the affiliation nodes are removed from the graph.

    \item[\textbf{(4)}] \textbf{Graph Coarsening Methods}~\citep{aconvmatch, fgc, ugc} 
    aim to reduce the scale of a graph by aggregating original nodes into supernodes while preserving critical structural or spectral properties (e.g., eigenvalues, local connectivity). These methods typically decouple the compression process from GNN training, relying on theoretical mechanisms such as spectral approximation or convolution matching to ensure representation quality. Since these algorithms are inherently target-agnostic, which indiscriminately merge target nodes and background nodes into supernodes, we employ these methods as baselines for graph-level classification.
    
        \item \textbf{A-CONVMATCH} (A-CM) is a graph coarsening method designed to preserve the inductive bias of GNNs. It formulates coarsening as learning a node-to-supernode partition (mapping) matrix $P$, optimizing the summarized graph such that node representations produced by a (simplified) graph convolution on the original graph are well-approximated by those produced on the coarsened graph and projected back via $P$. Concretely, A-CM employs a greedy pairwise matching strategy: it iteratively identifies and merges pairs of nodes (typically neighbors) that produce the most similar local convolutional responses, thereby grouping functionally equivalent nodes into supernodes. In our pipeline, we treat each supernode as a community, summarize the textual attributes of nodes within each community using an LLM, and reconstruct the coarsened graph with these summaries for downstream graph-level classification.
        
        \item \textbf{FGC} (Featured Graph Coarsening) is an optimization-based dimensionality reduction framework that distinctively learns the coarsened graph matrix $L_c$ and feature matrix $\tilde{X}$ \emph{jointly}, integrating both structural and attribute information. It defines a linear mapping matrix $P$ to map original nodes to "supernodes" via an objective that leverages Dirichlet Energy to enforce feature smoothness and homophily, while employing $l_{1,2}$-norm regularization to ensure a sparse and balanced node-to-supernode assignment. A key advantage of FGC is its theoretical guarantee of $\epsilon$-similarity ($\epsilon \in [0,1)$) between the original and coarsened graphs. 

        \item \textbf{UGC} (Universal Graph Coarsening) is a scalable graph coarsening method based on randomized hashing, designed to universally handle both homophilic and heterophilic graph data. It characterizes each node via an \emph{augmented feature vector} $F = \{(1-\alpha) \cdot X \oplus \alpha \cdot A\}$, which dynamically balances raw node attributes $X$ and structural adjacency information $A$ using a heterophily factor $\alpha$. UGC employs Locality Sensitive Hashing (LSH) to project these high-dimensional augmented features into buckets; nodes colliding in the same hash bucket are deemed structurally and attributively similar and are merged into a single supernode. This mechanism achieves linear time complexity while effectively preserving spectral properties. 
\end{itemize}

\begin{table*}[h]
\centering
\caption{Compression Results on \textbf{OGBN-ArXiv}. Comparison of GCR, Memory, Accuracy and GCI.}
\label{subtab:ogbn_arxiv_compression}
\footnotesize
\renewcommand{\arraystretch}{1.0}
\resizebox{\linewidth}{!}{
    \begin{tabular}{c|ccc|ccc|ccc|ccc}
    \toprule
    \toprule
    \multicolumn{13}{c}{\textbf{OGBN-ArXiv}} \\
    \cmidrule(r){1-13}
    \multirow{2}{*}{$|\mathcal{V}_\text{tg}|=7840$} 
    & GCR
    & \multirow{2}{*}{$|\widetilde{\mathcal{V}}|$} 
    & Data Memory 
    & & ACC (\%) &
    & \multicolumn{3}{c|}{GCI = ACC / GCR}
    & & $\overline{\text{GCI}}$ & \\  

    & (\%) & & (MB) & 3B & 7B & 8B & 3B & 7B & 8B & 3B & 7B & 8B \\
    \hline
    \noalign{\vskip 1pt}
    GCN&100.00&169343&304.18&-&51.05&-&-&51.05&-&-&1.00&- \\[1pt]
    GAT&100.00&169343&304.18&-&51.11&-&-&51.11&-&-&1.00&- \\[1pt]
    GIN&100.00&169343&304.18&-&47.28&-&-&47.28&-&-&0.93&- \\[1pt]
    GraphSAGE&100.00&169343&304.18&-&49.84&-&-&49.84&-&-&0.98&- \\[1pt]
    
    \hline
    \noalign{\vskip 1pt}
    Random&11.09&18787&33.11&28.65&20.85&63.66&258.25&187.94&573.82&5.06&3.68&11.24 \\[1pt]
    Degree&10.80&18294&32.37&28.97&21.03&63.95&268.17&194.67&591.97&5.25&3.81&11.60 \\[1pt]
    Number&11.32&19174&33.31&28.03&20.96&63.60&247.56&185.12&561.71&4.85&3.63&11.00 \\[1pt]
    RAG&10.98&18587&32.84&28.67&21.08&63.71&261.21&192.06&580.45&5.12&3.76&11.37 \\[1pt]
    
    \hline
    \noalign{\vskip 1pt}
    Skeleton ($\alpha$)&25.15&42589&79.84&31.86&24.33&65.41&126.68&96.74&260.08&2.48&1.90&5.09 \\[1pt]
    Skeleton ($\beta$)&5.70&9657&16.43&32.03&25.47&65.25&561.67&446.64&1144.21&11.00&8.75&22.49 \\[1pt]
    Skeleton ($\gamma$)&4.80&8122&13.98&31.83&25.47&64.99&663.65&531.05&1355.04&13.00&10.40&26.54 \\[1pt]
    
    \hline
    \noalign{\vskip 2pt}
    $\text{HS}_2\text{C}$ (Ours)
    &5.02&8507&15.70&35.01&29.99&\textbf{68.52}&696.92&596.99&\textbf{1363.98}&13.65&11.69&\textbf{26.72} \\
    \bottomrule
    \bottomrule
    \end{tabular}
}
\end{table*}

\begin{table*}[h]
\centering
\caption{Compression Results on \textbf{TAPE}. Comparison of GCR, Memory, Accuracy and GCI.}
\label{subtab:tape_compression}
\footnotesize
\renewcommand{\arraystretch}{1.0}
\resizebox{\linewidth}{!}{
    \begin{tabular}{c|ccc|ccc|ccc|ccc}
    \toprule
    \toprule
    \multicolumn{13}{c}{\textbf{TAPE}} \\
    \cmidrule(r){1-13}
    \multirow{2}{*}{$|\mathcal{V}_\text{tg}|=8621$} 
    & GCR
    & \multirow{2}{*}{$|\widetilde{\mathcal{V}}|$} 
    & Data Memory 
    & & ACC (\%) &
    & \multicolumn{3}{c|}{GCI = ACC / GCR}
    & & $\overline{\text{GCI}}$ & \\  

    & (\%) & & (MB) & 3B & 7B & 8B & 3B & 7B & 8B & 3B & 7B & 8B \\
    \hline
    \noalign{\vskip 1pt}
    GCN	&100.00	&57471 &145.38 &- &63.37 &- &- &63.37 &- &- &1.00 &- \\[1pt]
    GAT	&100.00	&57471 &145.38 &- &65.33 &- &- &65.33 &- &- &1.03 &- \\[1pt]
    GIN	&100.00	&57471 &145.38 &- &58.27 &- &- &58.27 &- &- &0.92 &- \\[1pt]
    GraphSAGE &100.00 &57471 &145.38 &- &61.87 &- &- & 61.87 &- &- & 0.98 &- \\[1pt]
    
    \hline
    \noalign{\vskip 1pt}
    Random &21.90 &12588 &31.96 &46.78 &17.77 &61.88 &213.58 &81.13 &282.52 &3.37 &1.28 &4.46 \\[1pt]
    Degree &21.59 &12410 &31.55 &46.73 &17.18 &61.98 &216.41 &79.56 &287.03 &3.41 &1.26 &4.53 \\[1pt]
    Number &22.03 &12660 &32.01 &46.60 &17.67 &62.11 &211.54 &80.21 &281.95 &3.34 &1.27 &4.45 \\[1pt]
    RAG	   &21.82 &12540 &31.85 &46.82 &17.57 &61.88 &214.58 &80.52 &283.60 &3.39 &1.27 &4.48 \\[1pt]
    
    \hline
    \noalign{\vskip 1pt}
    Skeleton ($\alpha$)	
    &34.85	&20030	&51.48 &44.50 &18.61 &63.63 &127.68 &53.40 &182.48 &2.01 &0.84 &2.88 \\[1pt]
    Skeleton ($\beta$)	
    &16.31	&9376	&23.49 &45.45 &18.61 &63.47 &278.59 &114.07 &389.04 &4.40 &1.80 &6.14 \\[1pt]
    Skeleton ($\gamma$)	
    &15.50	&8910	&22.35 &47.36 &18.51 &63.18 &305.48 &119.39 &407.52 &4.82 &1.88 &6.43 \\[1pt]
    
    \hline
    \noalign{\vskip 1pt}
    $\text{HS}_2\text{C}$ (Ours) 
    &16.07 &9236 &23.36 &49.08 &18.22 &\textbf{66.75} &305.37 &113.37 &\textbf{415.35} &4.82 &1.79 &\textbf{6.55} \\
    \bottomrule
    \bottomrule
    \end{tabular}
}
\end{table*}

\begin{table*}[ht]
\centering
\caption{Compression Results on \textbf{Instagram}. Comparison of GCR, Memory, Accuracy and GCI.}
\label{subtab:instagram_compression}
\footnotesize
\renewcommand{\arraystretch}{1.0}
\resizebox{\linewidth}{!}{
    \begin{tabular}{c|ccc|ccc|ccc|ccc}
    \toprule
    \toprule
    \multicolumn{13}{c}{\textbf{Instagram}} \\
    \cmidrule(r){1-13}
    \multirow{2}{*}{$|\mathcal{V}_\text{tg}|=9089$} 
    & GCR
    & \multirow{2}{*}{$|\widetilde{\mathcal{V}}|$} 
    & Data Memory 
    & & ACC (\%) &
    & \multicolumn{3}{c|}{GCI = ACC / GCR}
    & & $\overline{\text{GCI}}$ & \\  

    & (\%) & & (MB) & 3B & 7B & 8B & 3B & 7B & 8B & 3B & 7B & 8B \\
    \hline
    \noalign{\vskip 1pt}
    GCN	&100.00 &11339 &184.09 &- &65.66 &- &- &65.66 &- &- &1.00 &- \\[1pt]
    GAT	&100.00 &11339 &184.09 &- &63.71 &- &- &63.71 &- &- &0.97 &- \\[1pt]
    GIN	&100.00 &11339 &184.09 &- &65.25 &- &- &65.25 &- &- &0.99 &- \\[1pt]
    GraphSAGE &100.00 &11339 &184.09 &- &66.20 &- &- &66.20 &- &- &1.01 &- \\[1pt]
    
    \hline
    \noalign{\vskip 1pt}
    Random & 95.84	& 10867	& 176.53 &45.99 &41.62 & 64.81 &47.99 &43.43 &67.62 &0.73 &0.66 & 1.03 \\[1pt]
    Degree & 94.52	& 10718	& 174.12 &46.94 &41.30 & 64.94 &47.86 &42.11 &66.21 &0.73 &0.64 & 1.01  \\[1pt]
    Number & 96.18	& 10906	& 177.16 &44.79 &40.59 & 64.48 &45.62 &41.34 &65.68 &0.69 &0.63 & 1.00  \\[1pt]
    RAG	   & 95.48	& 10826	& 175.87 &45.62 &41.43 & 65.29 &46.50 &42.23 &66.54 &0.71 &0.64 & 1.01  \\[1pt]
    
    \hline
    \noalign{\vskip 1pt}
    Skeleton ($\alpha$)	& 95.73	& 10855	& 175.08 &45.65 &41.71 & 68.52 &47.21 &43.13 &70.86 &0.72 &0.66 & 1.08 \\[1pt]
    Skeleton ($\beta$)	& 85.07	& 9646	& 155.41 &45.73 &41.71 & 68.54 &48.78 &44.49 &73.11 &0.74 &0.68 & 1.11 \\[1pt]
    Skeleton ($\gamma$)	& 83.59	& 9478	& 152.71 &46.86 &41.43 & 70.21 &53.78 &47.55 & 80.58 &0.82 &0.72 & 1.23 \\[1pt]
    
    \hline
    \noalign{\vskip 1pt}
    $\text{HS}_2\text{C}$ (Ours) & 81.04 & 9189 & 148.51 &52.00 &42.21 &\textbf{72.76} &61.54 &49.96 &\textbf{86.11} &0.94 &0.76 &\textbf{1.31} \\ 
    \bottomrule
    \bottomrule
    \end{tabular}
}
\end{table*}

\begin{table*}[h]
\centering
\caption{Compression Results on \textbf{Citeseer}. Comparison of GCR, Memory, Accuracy and GCI.}
\label{subtab:citeseer_compression}
\footnotesize
\renewcommand{\arraystretch}{1.0}
\resizebox{\linewidth}{!}{
    \begin{tabular}{c|ccc|ccc|ccc|ccc}
    \toprule
    \toprule
    \multicolumn{13}{c}{\textbf{Citeseer}} \\
    \cmidrule(r){1-13}
    \multirow{2}{*}{$|\mathcal{V}_\text{tg}|=2566$} 
    & GCR
    & \multirow{2}{*}{$|\widetilde{\mathcal{V}}|$} 
    & Data Memory 
    & & ACC (\%) &
    & \multicolumn{3}{c|}{GCI = ACC / GCR}
    & & $\overline{\text{GCI}}$ & \\  

    & (\%) & & (MB) & 3B & 7B & 8B & 3B & 7B & 8B & 3B & 7B & 8B \\
    \hline
    \noalign{\vskip 1pt}
    GCN&100.00&3186&48.54&- &60.52 &- &- &60.52 &- &- &1.00 &- \\[1pt]
    GAT&100.00&3186&48.54&- &60.65 &- &- &60.65 &- &- &1.00 &- \\[1pt]
    GIN&100.00&3186&48.54&- &52.42 &- &- &52.42 &- &- &0.87 &- \\[1pt]
    GraphSAGE&100.00&3186&48.54&- &50.95 &- &- &50.95 &- &- &0.84 &- \\[1pt]
    
    \hline
    \noalign{\vskip 1pt}
    Random&98.18&3128&47.65&36.17&27.77&67.06&36.84&28.28&68.30&0.61&0.47&1.13 \\[1pt]
    Degree&98.09&3125&47.61&37.65&26.34&66.54&39.83&27.87&70.40&0.66&0.46&1.16 \\[1pt]
    Number&98.18&3128&47.65&37.26&27.24&66.72&38.74&28.32&69.37&0.64&0.47&1.15  \\[1pt]
    RAG   &98.12&3126&47.62&36.98&27.64&67.45&38.73&28.95&70.65&0.64&0.48&1.17  \\[1pt]
    
    \hline
    \noalign{\vskip 1pt}
    Skeleton ($\alpha$)&96.70&3081&46.86&31.96&16.37&69.40&33.39&17.10&72.49&0.55&0.28&1.20 \\[1pt]
    Skeleton ($\beta$) &93.75&2987&45.41&32.15&16.47&69.40&37.79&19.36&81.58&0.62&0.32&1.35 \\[1pt]
    Skeleton ($\gamma$)&87.13&2776&42.24&36.91&26.79&69.47&44.16&32.04&83.11&0.73&0.53&1.37 \\[1pt]
    
    \hline
    \noalign{\vskip 1pt}
    $\text{HS}_2\text{C}$ (Ours)&84.49&2692&41.00&68.28&36.67&\textbf{71.72}&84.26&45.25&\textbf{88.50}&1.39&0.75&\textbf{1.46} \\
    \bottomrule
    \bottomrule
    \end{tabular}
}
\end{table*}

\begin{table*}[h]
\centering
\caption{Compression Results on \textbf{Reddit}. Comparison of GCR, Memory, Accuracy and GCI.}
\label{subtab:reddit_compression}
\footnotesize
\renewcommand{\arraystretch}{1.0}
\resizebox{\linewidth}{!}{
    \begin{tabular}{c|ccc|ccc|ccc|ccc}
    \toprule
    \toprule
    \multicolumn{13}{c}{\textbf{Reddit}} \\
    \cmidrule(r){1-13}
    \multirow{2}{*}{$|\mathcal{V}_\text{tg}|=6866$} 
    & GCR
    & \multirow{2}{*}{$|\widetilde{\mathcal{V}}|$} 
    & Data Memory 
    & & ACC (\%) &
    & \multicolumn{3}{c|}{GCI = ACC / GCR}
    & & $\overline{\text{GCI}}$ & \\  

    & (\%) & & (MB) & 3B & 7B & 8B & 3B & 7B & 8B & 3B & 7B & 8B \\
    \hline
    \noalign{\vskip 1pt}
    GCN	&100.00	&33434	&559.89 &- &70.10 &- &- &70.10 &- &- &1.00 &- \\[1pt]
    GAT	&100.00	&33434	&559.89 &- &63.16 &- &- &63.16 &- &- &0.90 &- \\[1pt]
    GIN	&100.00	&33434	&559.89 &- &69.29 &- &- &69.29 &- &- &0.99 &- \\[1pt]
    GraphSAGE &100.00	&33434	&559.89 &- &63.95 &- &- &63.95 &- &- &0.91 &- \\[1pt]
    
    \hline
    \noalign{\vskip 1pt}
    Random	&69.76&23325&390.28&49.10&46.52&68.55&70.38&66.68&98.26&1.00&0.95&1.40 \\[1pt]
    Degree	&66.66&22286&372.67&48.92&45.02&67.97&73.39&67.54&101.97&1.05&0.96&1.45 \\[1pt]
    Number	&71.91&24042&401.67&50.13&44.02&67.36&69.71&61.22&93.67&0.99&0.87&1.34 \\[1pt]
    RAG	    &69.43&23213&388.26&50.00&45.21&67.85&72.02&65.12&97.73&1.03&0.93&1.39 \\[1pt]
    
    \hline
    \noalign{\vskip 1pt}
    Skeleton ($\alpha$)	&72.38&24199&403.51&50.31&45.54&74.86&69.51&62.92&103.43&0.99&0.90&1.48 \\[1pt]
    Skeleton ($\beta$)	&22.75&7607&126.55&50.31&45.55&74.86&221.12&200.20&329.02&3.15&2.86&4.69 \\[1pt]
    Skeleton ($\gamma$)	&20.89&6986&116.19&49.19&46.17&74.80&235.42&220.96&357.98&3.36&3.15&5.11 \\[1pt]
    
    \hline
    \noalign{\vskip 1pt}
    $\text{HS}_2\text{C}$ (Ours) &21.39&7153&132.32&50.87&50.76&\textbf{77.71}&237.77&237.26&\textbf{363.23}&3.39&3.38&\textbf{5.18} \\
    \bottomrule
    \bottomrule
    \end{tabular}
}
\end{table*}

\begin{table*}[h]
\centering
\caption{Compression Results on \textbf{Product-subset}. Comparison of GCR, Memory, Accuracy and GCI.}
\label{subtab:products_compression}
\footnotesize
\renewcommand{\arraystretch}{1.0}
\resizebox{\linewidth}{!}{
    \begin{tabular}{c|ccc|ccc|ccc|ccc}
    \toprule
    \toprule
    \multicolumn{13}{c}{\textbf{Product-subset}} \\
    \cmidrule(r){1-13}
    \multirow{2}{*}{$|\mathcal{V}_\text{tg}|=9171$} 
    & GCR
    & \multirow{2}{*}{$|\widetilde{\mathcal{V}}|$} 
    & Data Memory 
    & & ACC (\%) &
    & \multicolumn{3}{c|}{GCI = ACC / GCR}
    & & $\overline{\text{GCI}}$ & \\  

    & (\%) & & (MB) & 3B & 7B & 8B & 3B & 7B & 8B & 3B & 7B & 8B \\
    \hline
    \noalign{\vskip 1pt}
    GCN	& 100.00 & 45855 & 52.48 &- &68.08 &- &- &68.08 &- &- &1.00 &-  \\[1pt]
    GAT	& 100.00 & 45855 & 52.48 &- &67.33 &- &- &67.33 &- &- &0.99 &- \\[1pt]
    GIN	& 100.00 & 45855 & 52.48 &- &67.72 &- &- &67.72 &- &- &0.99 &- \\[1pt]
    GraphSAGE & 100.00 & 45855	& 52.48 &- &64.29 &- &- &64.29 &- &- &0.94 &- \\[1pt]
    
    \hline
    \noalign{\vskip 1pt}
    Random	& 45.29	& 20770	& 23.15 &49.77&24.37&75.82&109.88&53.80&167.39&1.61&0.79&2.46 \\[1pt]
    Degree	& 45.27	& 20760	& 23.13 &50.23&25.11&76.08&110.95&55.46&168.05&1.63&0.81&2.47 \\[1pt]
    Number	& 45.35	& 20797	& 23.17 &49.82&25.10&75.98&109.85&55.34&167.53&1.61&0.81&2.46 \\[1pt]
    RAG	    & 45.31	& 20775	& 23.15 &49.90&25.15&76.11&110.14&55.51&167.99&1.62&0.82&2.47 \\[1pt]
    
    \hline
    \noalign{\vskip 1pt}
    Skeleton ($\alpha$)	& 38.87	& 17822	& 19.32 &48.79&25.21&77.79&125.53&64.86&200.15&1.84&0.95&2.94 \\[1pt]
    Skeleton ($\beta$)	& 28.03	& 12851	& 13.70 &48.56&25.19&77.78&173.27&89.88&277.53&2.55&1.32&4.08 \\[1pt]
    Skeleton ($\gamma$)	& 23.06	& 10572	& 11.56 &49.97&25.32&\textbf{77.80}&216.74&109.82&337.45&3.18&1.61&4.96 \\[1pt]
    
    \hline
    \noalign{\vskip 1pt}
    $\text{HS}_2\text{C}$ (Ours) & 22.24 & 10199 & 11.65 &54.11&44.40&77.35&243.28&199.60&\textbf{347.77}&3.57&2.93&\textbf{5.11} \\
    \bottomrule
    \bottomrule
    \end{tabular}
}
\end{table*}

\end{document}